\documentclass{article}

\usepackage{microtype}
\usepackage{subcaption}
\usepackage{booktabs} \usepackage{multirow}
\usepackage[table]{xcolor}

\usepackage{hyperref}

\usepackage[accepted]{icml2024}

\usepackage{amsmath}
\usepackage{amssymb}
\usepackage{mathtools}
\usepackage{amsthm}
\usepackage{bm}
\usepackage{pifont}

\usepackage[capitalize,noabbrev]{cleveref}

\usepackage{listings}
\lstdefinestyle{overleaf}{
    backgroundcolor=\color[rgb]{0.95,0.95,0.92},   
    commentstyle=\color[rgb]{0,0.6,0},
    keywordstyle=\color{magenta},
    numberstyle=\tiny\color[rgb]{0.5,0.5,0.5},
    stringstyle=\color[rgb]{0.58,0,0.82},
    basicstyle=\ttfamily\footnotesize,
    breakatwhitespace=false,         
    breaklines=true,                 
    captionpos=b,                    
    keepspaces=true,                 
    numbers=left,                    
    numbersep=5pt,                  
    showspaces=false,                
    showstringspaces=false,
    showtabs=false,                  
    tabsize=2
}
\lstdefinestyle{simple}{
  backgroundcolor=\color{white},
  basicstyle=\fontsize{7.5pt}{7.5pt}\ttfamily\selectfont,
  keywordstyle=\fontsize{7.5pt}{7.5pt}\ttfamily\selectfont,
}
\theoremstyle{plain}

\theoremstyle{definition}

\theoremstyle{remark}

\usepackage[textsize=tiny]{todonotes}
\usepackage{xspace}

\definecolor{vpblue}{RGB}{33, 82, 206}
\definecolor{vporange}{RGB}{198, 64, 50}

\newcommand{\OURS}{VideoPrism\xspace}
\newcommand{\ie}{\textit{i.e.}\xspace}
\newcommand{\eg}{\textit{e.g.}\xspace}

\newcommand{\etc}{\textit{etc.}\xspace}
\newcommand{\upcolor}[1]{\textcolor{vpblue}{($\uparrow$#1)}}
\newcommand{\downcolor}[1]{\textcolor{vporange}{($\downarrow$#1)}}
\newcommand{\tabbar}[2]{\rowcolor{gray!15} \multicolumn{#1}{l}{\textit{#2}}}
\newcommand{\tabrow}{\rowcolor{gray!10 }}
\newcommand{\fadecell}[1]{\textcolor{gray}{#1}}
\newcommand{\cmark}{\ding{51}}
\newcommand{\xmark}{\ding{55}}
\newcommand{\myparagraph}[1]{\paragraph{#1}}

\icmltitlerunning{VideoPrism: A Foundational Visual Encoder for Video Understanding}

\begin{document}

\twocolumn[
\icmltitle{VideoPrism: A Foundational Visual Encoder for Video Understanding}

\icmlsetsymbol{equal}{*}
\icmlsetsymbol{core}{\textdagger}
\icmlsetsymbol{lead}{\textdaggerdbl}
\icmlsetsymbol{intern}{\textsection}

\begin{icmlauthorlist}
\icmlauthor{Long Zhao}{equal}
\icmlauthor{Nitesh B.\ Gundavarapu}{equal}
\icmlauthor{Liangzhe Yuan}{equal}
\icmlauthor{Hao Zhou}{equal}
\icmlauthor{Shen Yan}{core}
\icmlauthor{Jennifer J.\ Sun}{core}
\icmlauthor{Luke Friedman}{core}
\icmlauthor{Rui Qian}{core}
\icmlauthor{Tobias Weyand}{}
\icmlauthor{Yue Zhao}{intern}
\icmlauthor{Rachel Hornung}{}
\icmlauthor{Florian Schroff}{}
\icmlauthor{Ming-Hsuan Yang}{}
\icmlauthor{David A.\ Ross}{}
\icmlauthor{Huisheng Wang}{}
\icmlauthor{Hartwig Adam}{}
\icmlauthor{Mikhail Sirotenko}{lead}
\icmlauthor{Ting Liu}{lead}
\icmlauthor{Boqing Gong}{lead}
\end{icmlauthorlist}
\begin{center}
    \normalsize{Google}
\end{center}

\icmlcorrespondingauthor{Long Zhao}{longzh@google.com}
\icmlcorrespondingauthor{Mikhail Sirotenko}{msirotenko@google.com}
\icmlcorrespondingauthor{Ting Liu}{liuti@google.com}
\icmlcorrespondingauthor{Boqing Gong}{bgong@google.com}

\icmlkeywords{Video Foundation Model, Vision-Language Model, Self-Supervised Learning}

\vskip 0.3in
]

\printAffiliationsAndNotice{
$^*$Equal primary contribution. $^\dagger$Equal core technical contribution. $^\ddag$Equal senior contribution, project leads. $^\S$This work was done while the author was a student researcher at Google Research} 

\begin{abstract}
We introduce \OURS, a general-purpose video encoder that tackles diverse video understanding tasks with a single frozen model.
We pretrain \OURS on a heterogeneous corpus containing 36M high-quality video-caption pairs and 582M video clips with noisy parallel text (\eg, ASR transcripts). The pretraining approach improves upon masked autoencoding by global-local distillation of semantic video embeddings and a token shuffling scheme, enabling \OURS to focus primarily on the video modality while leveraging the invaluable text associated with videos. We extensively test \OURS on four broad groups of video understanding tasks, from web video question answering to CV for science, achieving state-of-the-art performance on 31 out of 33 video understanding benchmarks. Our models are released at \url{https://github.com/google-deepmind/videoprism}.
\end{abstract} 
\begin{figure*}[t]
\vskip 0.1in
\begin{center}
\centerline{\includegraphics[width=\linewidth]{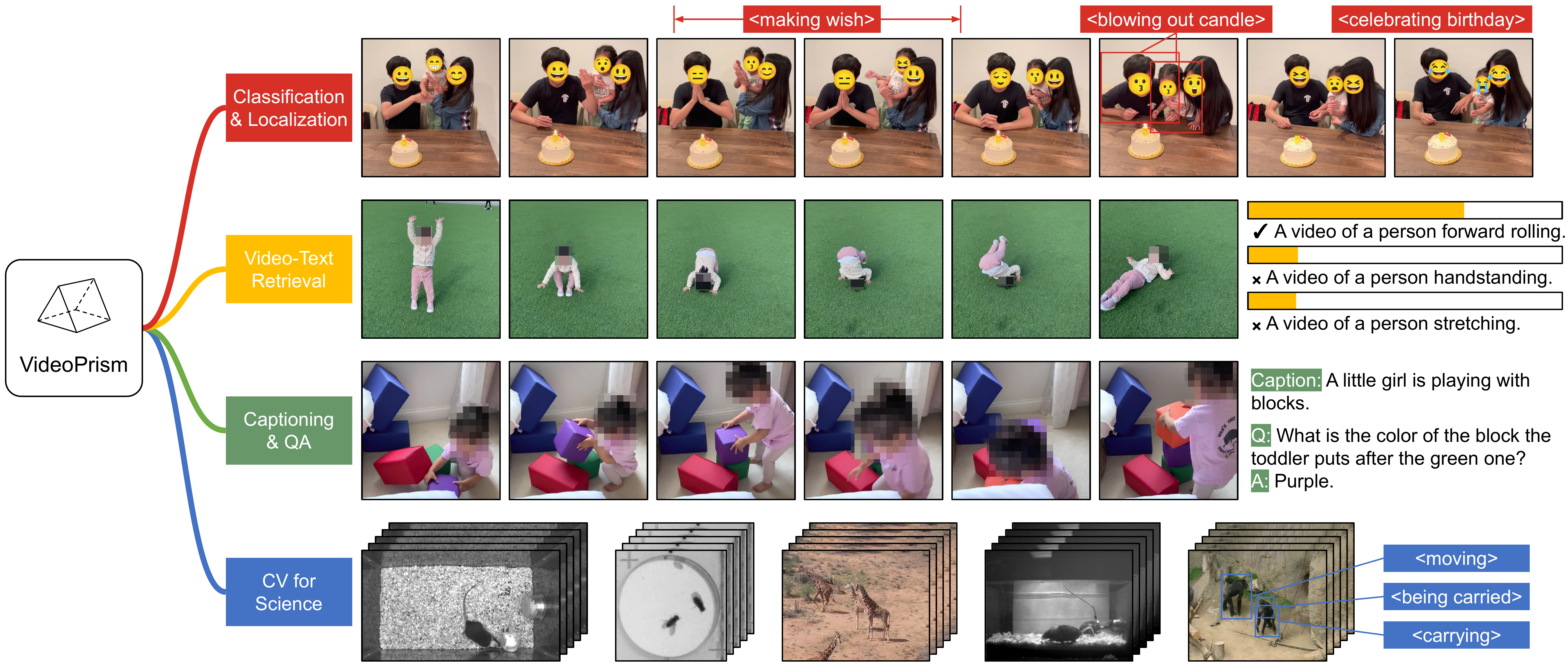}}
\captionof{figure}{\textbf{\OURS} is a general-purpose video encoder that enables state-of-the-art results over a wide spectrum of video understanding tasks by producing video representations from one \emph{single frozen} model. }
\label{fig:teaser}
\end{center}
\vskip -0.1in
\end{figure*}

\section{Introduction}
Videos are a rich and dynamic archive of real-world perceptual experience, spanning diverse domains from everyday life to scientific observations.
Video foundation models (ViFMs) hold enormous potential to unlock new insights within this vast corpus.
While prior work has made great progress towards general video understanding~\cite{xu2021videoclip,wang2022internvideo,yan2022videococa,tong2022videomae,li2023unmasked,wang2023masked}, building a truly foundational video model is still an elusive goal.  Existing models often struggle to balance appearance-heavy tasks with motion-centric reasoning, falling behind task-specialized models across many benchmarks~\cite{yuan2023videoglue}.

We introduce \OURS, a general-purpose video encoder designed to tackle  a wide spectrum of video understanding tasks, including classification, localization, retrieval, captioning, and question answering (QA) (\cref{fig:teaser}). Evaluated extensively on computer vision (CV) datasets and CV for science domains like neuroscience and ecology, \OURS achieves state-of-the-art performance with minimal adaptation, using a \textit{single frozen} model. We emphasize this frozen-encoder setting following prior work~\cite{radford2021learning,alayrac2022flamingo,tang2023video,li2023blip} and for its practical utility given the otherwise high computational and memory cost of finetuning video models.

The design philosophy behind \OURS is as follows. Pretraining data is fundamental to foundation models (FMs)~\cite{bommasani2021opportunities}, and the ideal pretraining data for ViFMs would be a representative sample of all videos in the world. Most videos from this  sample will have no (or very noisy) parallel text describing the content; however, when such text exists, it provides priceless semantic clues about the video space. Accordingly, our pretraining strategy should focus primarily on the video modality and yet take full advantage of any available video-text pairs.

On the data side, we approximate the desired  pretraining corpus by assembling 36M high-quality video-caption pairs and 582M video clips with noisy parallel text (\eg, ASR transcripts, generated captions, and retrieved text). On the modeling side, we first contrastively learn semantic video embeddings~\cite{radford2021learning,jia2021scaling} from all our video-text pairs of various qualities. Subsequently, we capitalize on the extensive video-only data by distilling the semantic embeddings globally and token-wise, improving upon masked video modeling~\cite{tong2022videomae,feichtenhofer2022masked,wang2023masked} described below.

Despite its success for natural language~\cite{devlin2019bert,brown2020language,anil2023palm}, masked data modeling remains challenging for CV as raw visual signals lack semantics. Existing works approach this challenge by borrowing {indirect} semantics (\eg, using CLIP~\cite{radford2021learning} to bootstrap models~\cite{fang2022eva,fang2023eva} or tokenizers~\cite{peng2022beit}) or {implicitly} promoting them (\eg, tokenizing visual patches~\cite{zhou2022ibot,bao2022beit,oquab2023dinov2}, combining a high masking ratio and  lightweight decoder~\cite{he2022masked}). 

We build on the above ideas with a two-stage approach tailored to our pretraining data. 
We first train a video encoder, along with a paired text encoder, over the video-text pairs using a contrastive objective~\cite{gutmann2010noise,radford2021learning}.
Next, we continue training the encoder over all video-only data by masked video modeling with two improvements:
(1) the model is required to predict both the video-level global embedding and token-wise embeddings from the first stage based on unmasked input video patches; (2) random shuffle is applied to the encoder's output tokens before they are passed to the decoder to avoid learning shortcuts. Notably, our pretraining utilizes two supervisory signals: a video's text description and its contextual self-supervision, enabling \OURS to excel on both appearance- and motion-focused tasks. Indeed, previous works have shown that video captions mainly reveal appearance cues~\cite{wang2023paxion}, and contextual self-supervision facilitates learning  motion~\cite{tong2022videomae}.

\begin{figure}[t]
\vskip -0.1in
\begin{center}
\includegraphics[width=\linewidth]{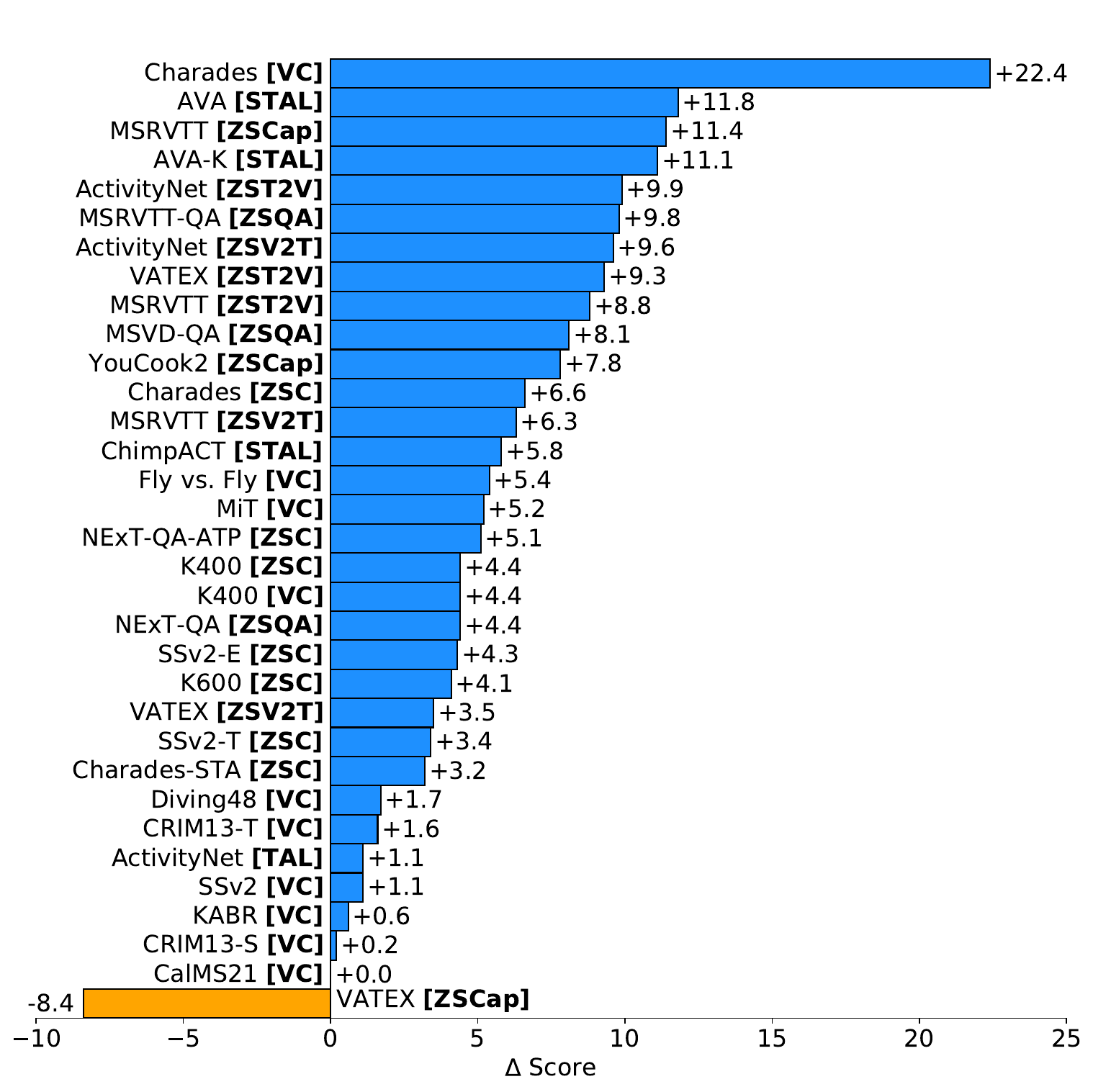} \\
\vskip -0.1in
\caption{\textbf{\OURS \textit{vs.}\ the previous best-performing FMs.} Please find the details of this figure in \cref{app:evaluation_data}.}
\label{fig:flvid_all_stats}
\end{center}
\vskip -0.1in
\end{figure}

\begin{table*}[t]
\caption{\textbf{Composition of our pretraining corpus.} We report the numbers of videos and clips we were able to access during pretraining.}
\label{tbl:pretraining_data}
\begin{center}
\resizebox{0.94\textwidth}{!}{\begin{tabular}{@{}l|cllcrr@{}}
    \toprule
    Pretraining datasets & Public & Domain & Caption source & Caption quality & \# of videos & \# of clips \\
    \midrule
Anonymous-Corpus \#1 & \xmark & Web video & Manual labelled & High & 36.1M & 36.1M  \\
    \midrule
    WTS-70M~\cite{stroud2020learning} & \cmark & YouTube video & Metadata & Low & 55.1M & 55.1M \\
    YT-Temporal-180M~\cite{zellers2021merlot} & \cmark & YouTube video & ASR & Low & 2.3M & 87.8M \\
    VideoCC~\cite{nagrani2022learning} & \xmark & YouTube video & Image captions for mining & Low & 133.5M & 191.1M \\
    InternVid~\cite{wang2023internvid} & \cmark & YouTube video & Generated by VLM/LLM & Medium & 2.8M & 7.0M \\
    Anonymous-Corpus \#2 & \xmark & YouTube video & ASR & Low & 44.6M & 170.3M  \\
    Anonymous-Corpus \#3 & \xmark & YouTube video & Generated by VLM/LLM & Medium & 36.7M & 71.5M \\
\bottomrule
\end{tabular}
}
\end{center}
\vskip -0.1in
\end{table*} 

\myparagraph{Contributions.} 
\OURS is a state-of-the-art, general-purpose video encoder. 
We advocate for a scalable strategy for collecting pretraining videos, combining manually captioned videos with those containing noisy textual descriptions.
We design a unique two-stage pretraining approach tailored to this hybrid data, leveraging video-language contrastive learning to harvest semantics, followed by improved masked video modeling with global-local distillation and token shuffling.
Finally, we present a comprehensive evaluation on four broad categories of understanding tasks across 33 diverse benchmarks, including videos from the web, scripted performances, and scientific experiments. 
Results demonstrate that \OURS significantly outperforms existing ViFMs on 31 benchmarks (\cref{fig:flvid_all_stats}). Importantly, no single baseline model consistently achieves second-best performance, indicating \OURS's robust generalizability.

 \section{Approach}

\subsection{Pretraining data}
\label{sec:app:data}
Our pretraining data consists of 36M clips (sampled from 36M videos) with high-quality manually labelled \emph{captions} and 582M clips (from 275M videos) with noisy parallel \emph{text}, as summarized in \cref{tbl:pretraining_data}. 
The 36M high-quality video-caption pairs in Anonymous-Corpus \#1 are the largest of its kind for ViFMs, to our knowledge, but they are still an order of magnitude smaller than the image-language data used to fuel image FMs~\cite{radford2021learning,yu2022coca}. Hence, we also collect large-scale video-text data whose noisy text is generated through ASR, metadata, and large multimodal models~\cite{wang2023internvid,zhao2024distilling}, \etc This subset of videos corresponds to the rows from WTS-70M to Anonymous-Corpus \#3 in \cref{tbl:pretraining_data}, and we provide more details in \cref{app:data}. 

Importantly, unlike previous works~\cite{tong2022videomae,wang2022internvideo,li2023unmasked,wang2023videomae}, we do not incorporate any training sets from the evaluation benchmarks, \eg, Kinetics~\cite{kay2017kinetics}, for either pretraining or post-pretraining. This conscious choice avoids overly tuning our model towards certain evaluation benchmarks. Moreover, we carefully de-duplicate the pretraining corpus against the videos in all the 33 evaluation benchmarks used in this paper to ensure that there is no data leakage.

\subsection{Model architecture}
\label{sec:app:arch}
The \OURS model architecture stems from the standard Vision Transformer (ViT)~\cite{dosovitskiy2021image}, with a factorized design in space and time following ViViT~\cite{arnab2021vivit}. However, we remove the global average pooling layer of ViViT immediately after the spatial encoder so that the spatiotemporal dimensions remain in the output token sequence, facilitating the downstream tasks that require fine-grained features (\eg, spatiotemporal action localization). We experiment with two model configurations: \emph{\OURS-g} and \emph{\OURS-B}. \OURS-g is the ViT-giant network~\cite{zhai2022scaling} with 1B parameters in the spatial encoder, and \OURS-B is a smaller variant with the ViT-Base network~\cite{dosovitskiy2021image}. \cref{app:architecture} describes the two network architectures in detail.

\begin{figure*}[t]
\vskip 0.1in
\begin{center}
\centerline{\includegraphics[width=.98\linewidth]{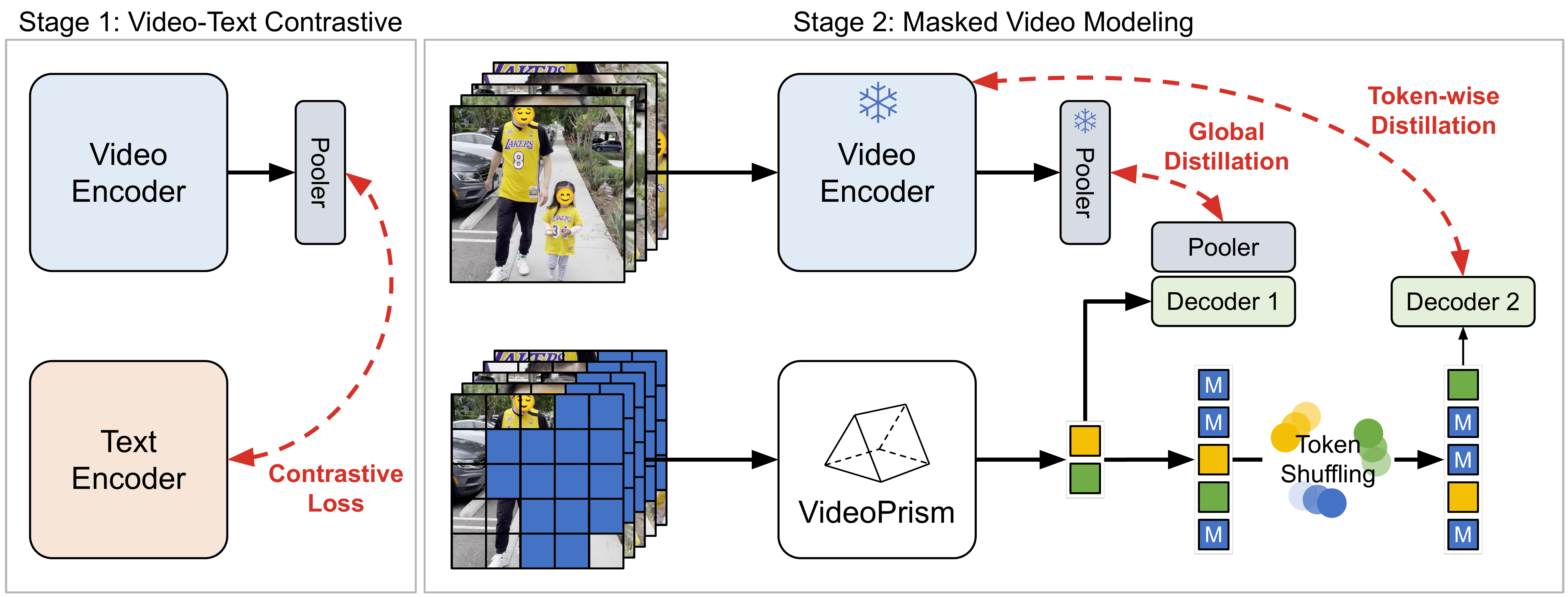}}
\caption{\textbf{Illustration of our two-stage pretraining.} Stage 1 trains video and text encoders with contrastive loss on video-text pairs, supplying semantic video embeddings to the next stage. Stage 2 continues to train the video encoder, now called \OURS, using improved masked autoencoding on video-only clips. The frozen Stage-1 encoder uses unmasked 3D video patches to produce a global semantic embedding of the whole video and token-wise embeddings. Decoder 2 processes shuffled tokens with positional embedding, while Decoder 1 has no positional embedding.}
\label{fig:pipeline}
\end{center}
\vskip -0.1in
\end{figure*}

\subsection{Training algorithm}
Our goal is to leverage both video-text pairs and the video-only data curated in \cref{sec:app:data}  to train \OURS scalably, so as to make \OURS a foundational video encoder capable of capturing both appearance and motion semantics from videos. We highlight the video-only modality rather than solely relying on video-text because the text in our large-scale pretraining corpus is very noisy for a majority of the videos. As shown in \cref{fig:pipeline}, the training pipeline of \OURS consists of two stages: \emph{video-text contrastive training} and \emph{masked video modeling}.

\subsubsection{Stage 1: Video-text contrastive training}
\label{sec:app:train:teacher}
In the first stage, we conduct contrastive learning to align a video encoder with a text encoder using all the video-text pairs. Following prior arts~\cite{radford2021learning,jia2021scaling,cheng2023vindlu}, we minimize a symmetric cross-entropy loss over the similarity scores of all video-text pairs in a mini-batch, initialize the spatial encoding modules using the image model of CoCa~\cite{yu2022coca}, and include WebLI~\cite{chen2023pali} (about 1B images with alt-text) to the pretraining. The video encoder's features are aggregated through a multi-head attention pooler (MAP)~\cite{lee2019set} before the loss computation. This stage allows the video encoder to learn rich visual semantics from language supervision, and the resulting model supplies semantic video embeddings for the second-stage training.

\subsubsection{Stage 2: Masked video modeling}\label{sec:app:train:student}

Training solely on vision-text data as in Stage 1 presents challenges: text descriptions can be noisy, and they often capture appearance more than motion~\cite{hendricks2021probing,momeni2023verbs}.
To address this, our second-stage training focuses on learning both appearance and motion information from video-only data.
Building upon the success of masked autoencoding for motion understanding~\cite{wang2022internvideo,wang2023masked}, we adapt this approach for the second stage, while ensuring that the model retains the semantic knowledge acquired in the first stage.

In this stage, we continue to train the video encoder on video-only data using improved masked video modeling. These improvements include (1) a novel token shuffling scheme to prevent decoding shortcuts and (2) global and token-wise distillation losses to effectively leverage the knowledge acquired in the first stage. 
As illustrated in Figure~\ref{fig:pipeline}, the second-stage (student) model learns to predict the first-stage (teacher) model's embeddings of \emph{all} tokens based on a masked video.
The encoder-decoder Transformers are decoupled following \citet{he2022masked}'s design.

\myparagraph{Token shuffling.} As we effectively initialize the second-stage model from the first stage, one issue is that the model may create a shortcut for the decoder to copy and paste the unmasked tokens while predicting only the masked ones, making it an easier task to solve than predicting all tokens. To address this issue, we \emph{randomly shuffle} the token sequence output by the encoder before feeding it to the decoder, and the decoder adds positional embeddings to this sequence after the shuffling. Note that this shuffling operation avoids the copy-and-paste shortcut of unmasked tokens that the decoder can potentially explore. One can also view it akin to Jigsaw puzzles~\cite{noroozi2016unsupervised} which the decoder tries to solve for the unmasked tokens while it predicts the masked ones. 

\myparagraph{Global-local distillation.} Unlike the masked distillation for images~\cite{fang2022eva,fang2023eva}, we find that our second-stage model underperforms the first-stage teacher on appearance-heavy tasks when only the masked modeling loss is utilized, probably attributing to {catastrophic forgetting}~\cite{mccloskey1989catastrophic} in the two-stage pretraining. To mitigate this issue, we add an additional loss to let the second-stage model distill the global embedding of the full intact video from the first-stage teacher using the visible tokens.
Hence, the second-stage training loss combines the  token-wise masked video modeling and global distillation. Due to space limit, we refer readers to \cref{app:implement} for the detailed implementation and training configurations. \begin{table*}[t]
\caption{
\textbf{Evaluating FMs on the VideoGLUE benchmark~\cite{yuan2023videoglue} with frozen backbones.} Only weights in the task heads are trained using the downstream tasks' training sets.
On all video classification (VC) tasks except Charades, we report top-1 accuracy. On Charades, temporal localization (TAL), and spatiotemporal localization (STAL) tasks, we use mean average precision (mAP) as the evaluation metric. -B, -L, -g indicate that the underlying models are respectively the base, large, and giant ViT~\cite{dosovitskiy2021image}.
}
\label{tbl:videoglue_fz_main}
\begin{center}
\begin{scriptsize}
\begin{tabular}{@{}l|cc|cc|c|c|cc@{}} 
\toprule
\multirow{2}{*}{Methods} & \multicolumn{2}{c|}{VC (A)} & \multicolumn{2}{c|}{VC (M)} & VC (ML) & TAL & \multicolumn{2}{c}{STAL} \\
& \textbf{K400} & \textbf{MiT} & \textbf{SSv2} & \textbf{D48} & \textbf{Charades} & \textbf{ActivityNet} & \textbf{AVA} & \textbf{AVA-K} \\
\midrule
\tabbar{9}{Base-scale models} \\
CLIP-B~\cite{radford2021learning} & 75.2 & 32.6 & 41.0 & 44.1 & 11.2 & 32.7 & 21.1 & 25.9 \\
VATT-B~\cite{akbari2021vatt} & 75.1 & 32.1 & 57.8 & 49.7 & 33.3 & 35.3 & 20.3 & 22.2 \\
InternVideo-B~\cite{wang2022internvideo} & 69.3 & 26.3 & 58.2 & 55.6 & 13.0 & 33.3 & 13.4 & 15.7 \\
UMT-B~\cite{li2023unmasked} & 77.1 & 34.0 & 47.7 & 47.8 & 30.1 & 35.8 & 20.7 & 21.1 \\
\textbf{\OURS-B} & \textbf{84.2} \upcolor{7.1} & \textbf{40.8} \upcolor{6.8} & \textbf{63.6} \upcolor{5.4} & \textbf{67.4} \upcolor{12.} & \textbf{40.4} \upcolor{7.1} & \textbf{36.6} \upcolor{0.8} & \textbf{30.6} \upcolor{9.5} & \textbf{31.8} \upcolor{5.9} \\
\midrule
\tabbar{9}{Large-scale models} \\
VideoMAE-v2-g~\cite{wang2023videomae} & 82.1 & 35.0 & 56.1 & 60.5 & 22.4 & 35.3 & 21.5 & 23.3 \\
InternVideo-L~\cite{wang2022internvideo} & 78.6 & 33.7 & 67.4 & 69.6 & 20.9 & 35.9 & 20.8 & 21.3 \\
UMT-L~\cite{li2023unmasked} & 82.8 & 40.3 & 54.5 & 49.0 & 39.9 & 36.7 & 24.4 & 26.2 \\
\textbf{\OURS-g} & \textbf{87.2} \upcolor{4.4} & \textbf{45.5} \upcolor{5.2} & \textbf{68.5} \upcolor{1.1} & \textbf{71.3} \upcolor{1.7} & \textbf{62.3} \upcolor{22.} & \textbf{37.8} \upcolor{1.1} & \textbf{36.2} \upcolor{12.} & \textbf{37.3} \upcolor{11.} \\
\bottomrule
\end{tabular}
\end{scriptsize}
\end{center}
\vskip -0.1in
\end{table*}

\section{Experiments}\label{sec:exp}
We evaluate \OURS on a wide spectrum of video-centric understanding tasks to demonstrate its capability and generalizability. We group the tasks into four categories: (1) general video-only understanding, including classification and spatiotemporal localization (\cref{sec:exp:videoglue}), (2) zero-shot video-text retrieval (\cref{sec:exp:vt_retrieval}), (3) zero-shot video captioning and QA (\cref{sec:exp:cap_vqa}), and (4) CV for science (\cref{sec:exp:ai4s}). For all experiments in the main paper, we freeze \OURS as a video encoder and only train task-specific components for the tasks in groups (1), (2), and (4) and some adaptation layers connecting \OURS to an LLM for (3). In the appendices, we report more results of end-to-end and adapter finetuning. Note that our evaluation strategy, freezing the visual encoder, aligns with prior works~\cite{he2022masked,singh2022flava,yuan2023videoglue} and is almost a go-to choice for building VideoLLMs~\cite{tang2023video}. It is especially needed for videos because finetuning a ViFM is  prohibitively expensive, while a frozen ViFM allows one to amortize the cost of video encoding across multiple tasks. All results in the main text are produced using the same frozen VideoPrism-B/g checkpoint, corresponding to the base/giant model.

\subsection{Classification and spatiotemporal localization}
\label{sec:exp:videoglue}

We compare \OURS with state-of-the-art FMs on a video-only understanding benchmark: VideoGLUE~\cite{yuan2023videoglue}.
By design, VideoGLUE evaluates FMs through four adaptation methods over eight hallmark datasets, representing appearance-focused action recognition (VC (A)), motion-rich action recognition (VC (M)), multi-label video classification (VC (ML)), temporal action localization (TAL), and spatiotemporal action localization (STAL). 
This benchmark introduces a VideoGLUE score (VGS), considering the tradeoff between adaptation costs and performance, to provide a holistic view of FMs' capabilities on the video-only understanding tasks.
We present the frozen-backbone evaluation results in the main paper and leave the rest to \cref{app:videoglue}. We employ an MAP head~\cite{yuan2023videoglue} in action recognition (MAP probing) and spatiotemporal localization and use G-TAD~\cite{xu2019gtad} for temporal localization (see \cref{app:videoglue:tasks} for details).

\myparagraph{Datasets.}
The eight datasets in VideoGLUE are as follows. For apperance-focused action recognition, Kinetics-400 (K400)~\cite{kay2017kinetics} and Moments-in-Time (MiT)~\cite{monfort2019moments} are sourced from web videos. Something-Something v2 (SSv2)~\cite{goyal2017something} and Diving48 (D48)~\cite{li2018resound} are fine-grained motion-rich action recognition datasets. Besides, Charades~\cite{sigurdsson2016hollywood} provides a multi-label classification problem using scripted indoor videos. The temporal localization task entails one dataset, ActivityNet v1.3~\cite{caba2015activitynet}, and the spatiotemporal localization contains Atomic Visual Actions (AVA)~\cite{gu2018ava} and AVA-Kinetics (AVA-K)~\cite{li2020ava}.

\myparagraph{Main results.}
\cref{tbl:videoglue_fz_main} shows the frozen-backbone results on VideoGLUE.
\OURS outperforms the baselines on all datasets by a large margin. Besides, increasing VideoPrism's underlying model size from ViT-B to ViT-g significantly improves the performance. Notably, no baselines can perform second best on all benchmarks, indicating the previous methods might be developed towards certain aspects of video understanding, while \OURS consistently improves on this wide range of tasks. This result implies that \OURS packed various video signals into one encoder: semantics at multiple granularities, appearance \textit{vs.}\ motion cues, spatiotemporal information, and robustness to diverse video sources (\eg, web videos \textit{vs.}\ scripted performance). 

In~\cref{app:videoglue:benchmark}, following the VideoGLUE setup, we conduct experiments on other adaptation methods, including end-to-end and parameter-efficient finetuning, and multi-layer attention pooling. Different adaption methods trade off computational cost with performance, accounting for real-world application considerations, and the VGS aggregates them into a scalar value. \OURS achieves VGS $51.25$, outperforming all baseline FMs in \cref{tbl:videoglue_score} and scoring $13.6\%$ higher than the second best model (UMT).

\subsection{Zero-shot video-text retrieval and classification}
\label{sec:exp:vt_retrieval}

\begin{table*}[t]
\caption{\textbf{Results of zero-shot video-text retrieval}. We report the Recall@1 (R@1) and R@5 for all the benchmarks. Note that we follow the 1K-A split of MSRVTT produced by~\citet{bain2021frozen} which contains $1,000$ videos for testing. Please refer to~\cref{app:lit:msrvtt} for the results on the full split of MSRVTT proposed by~\citet{xu2016msr}.}
\label{tbl:zs_retrieval}
\begin{center}
\begin{scriptsize}
\begin{tabular}{@{}l|cccc|cccc|cccc@{}}
    \toprule
    \multirow{3}{*}{Methods} & \multicolumn{4}{c|}{\textbf{MSRVTT (1K-A)}} & \multicolumn{4}{c|}{\textbf{VATEX}} & \multicolumn{4}{c}{\textbf{ActivityNet}} \\
    & \multicolumn{2}{c}{Text $\rightarrow$ Video} & \multicolumn{2}{c|}{Video $\rightarrow$ Text} & \multicolumn{2}{c}{Text $\rightarrow$ Video} & \multicolumn{2}{c|}{Video $\rightarrow$ Text} & \multicolumn{2}{c}{Text $\rightarrow$ Video} & \multicolumn{2}{c}{Video $\rightarrow$ Text} \\
    & R@1 & R@5 & R@1 & R@5 & R@1 & R@5 & R@1 & R@5 & R@1 & R@5 & R@1 & R@5 \\
    \midrule
    CLIP-L~\cite{radford2021learning} & 35.0 & - & 32.3 & - & 45.2 & - & 59.2 & - & 25.2 & - & 20.7 & - \\
    Singularity-B~\cite{lei2022revealing} & 34.0 & 56.7 & - & - & - & - & - & - & 30.6 & 55.6 & - & - \\
    VideoCoCa-g~\cite{yan2022videococa} & 43.9 & 69.9 & 45.4 & 68.6 & 53.2 & 83.3 & 73.6 & 93.2 & 34.5 & 63.2 & 33.0 & 61.6 \\
    InternVideo-L~\cite{wang2022internvideo} & 40.7 & - & 39.6 & - & 49.5 & - & 69.5 & - & 30.7 & - & 31.4 & - \\
    UMT-L~\cite{li2023unmasked} & 42.6 & 64.4 & 38.6 & 59.8 & - & - & - & - & 42.8 & 69.6 & 40.7 & 67.6 \\
\midrule
    \multirow{2}{*}{\textbf{\OURS-B}} & 51.4 & 74.4 & 50.2 & 73.2 & 57.7 & 88.5 & 76.2 & 93.7 & 49.6 & 76.7 & 47.9 & 75.0 \\
    & \upcolor{7.5} & \upcolor{4.5} & \upcolor{4.8} & \upcolor{4.6} & \upcolor{4.5} & \upcolor{5.2} & \upcolor{2.6} & \upcolor{0.5} & \upcolor{6.8} & \upcolor{7.1} & \upcolor{7.2} & \upcolor{7.4} \\
    \hline
    \multirow{2}{*}{\textbf{\OURS-g}} & \textbf{52.7} & \textbf{77.2} & \textbf{51.7} & \textbf{75.2} & \textbf{62.5} & \textbf{91.0} & \textbf{77.1} & \textbf{95.6} & \textbf{52.7} & \textbf{79.4} & \textbf{50.3} & \textbf{77.1} \\
    & \upcolor{8.8} & \upcolor{7.3} & \upcolor{6.3} & \upcolor{6.6} & \upcolor{9.3} & \upcolor{7.7} & \upcolor{3.5} & \upcolor{2.4} & \upcolor{9.9} & \upcolor{9.8} & \upcolor{9.6} & \upcolor{9.5} \\
    \bottomrule
\end{tabular}
\end{scriptsize}
\end{center}
\vskip -0.1in
\end{table*}

\begin{table*}[t]
\caption{\textbf{Comparison to state-of-the-art results on zero-shot video classification}. Results are reported in Top-1/5 accuracy (\%) on Kinetics-400 and Something-Something v2, multi-choice (MC) retrieval accuracy (\%) on NExT-QA (ATP-Hard) and Charades-STA, and mean average precision (mAP) on Charades. In line with~\citet{ni2022expanding}, we follow the single-view evaluation protocol for simplicity. Models pretrained with extra modalities (\eg, audio) in addition to vision and language are marked in gray.}
\centering
\begin{subtable}[t]{.41\linewidth}
    \begin{center}
    \caption{Kinetics-400}
    \vspace{-0.3\baselineskip}
    \begin{scriptsize}
    \begin{tabular}{@{}lcc@{}}
    	\toprule
    	Methods & Top-1 Acc & Top-5 Acc \\
    	\midrule
CoCa-g~\cite{yu2022coca} & 66.4 & 87.1 \\
    	VideoCoCa-g~\cite{yan2022videococa} & 72.0 & 90.5 \\
    	Text4Vis-L~\cite{wu2023revisiting} & 61.0 & - \\
    	\fadecell{ImageBind-H~\cite{girdhar2023imagebind}} & \fadecell{50.0} & \fadecell{-} \\
    	\fadecell{LanguageBind-L~\cite{zhu2023languagebind}} & \fadecell{64.0} & \fadecell{-} \\
    	\fadecell{IMP-MoE-L~\cite{akbari2023alternating}} & \fadecell{77.0} & \fadecell{-} \\
    	\textbf{\OURS-B} & 71.3 \downcolor{0.7} & 91.7 \upcolor{1.2} \\
    	\textbf{\OURS-g} & \textbf{76.4} \upcolor{4.4} & \textbf{94.3} \upcolor{3.8} \\
    	\bottomrule
    \end{tabular}
    \end{scriptsize}
    \label{tbl:zs_cls:k400}
    \end{center}
\end{subtable}
\begin{subtable}[t]{.4\linewidth}
    \begin{center}
    \caption{Something-Something v2}
    \vspace{-0.3\baselineskip}
    \begin{scriptsize}
    \begin{tabular}{@{}lcc@{}}
    	\toprule
    	Methods & Temporal & Events \\
    	\midrule
    	VideoCLIP-B~\cite{xu2021videoclip} & 9.8 & 6.4 \\
    	CoCa-g~\cite{yu2022coca} & 13.4 & 10.4 \\
    	VideoCoCa-g~\cite{yan2022videococa} & 14.1 & 10.7 \\
VNLI-L~\cite{yarom2023you} & 14.6 & 10.4 \\
    	VideoCon-L~\cite{bansal2023videocon} & 15.2 & 11.4 \\
    	\fadecell{ImageBind-H~\cite{girdhar2023imagebind}} & \fadecell{10.5} & \fadecell{5.5} \\
    	\textbf{\OURS-B} & 16.1 \upcolor{0.9} & 11.9 \upcolor{0.5} \\
        \textbf{\OURS-g} & \textbf{18.6} \upcolor{3.4} & \textbf{15.7} \upcolor{4.3} \\
    	\bottomrule
    \end{tabular}
    \end{scriptsize}
    \label{tbl:zs_cls:ssv2}
    \end{center}
\end{subtable}
\vskip 0.1in
\begin{subtable}[t]{.3\linewidth}
    \begin{center}
    \caption{NExT-QA (ATP-Hard)}
    \vspace{-0.3\baselineskip}
    \setlength{\tabcolsep}{4pt}
    \begin{scriptsize}
    \begin{tabular}{@{}lc@{}}
        \toprule Methods & MC Acc \\ 
      	\midrule
      	CLIP-B~\cite{radford2021learning} & 23.8 \\
ATP-B~\cite{buch2022revisiting} & 20.2 \\
    	VideoCoCa-g~\cite{yan2022videococa} & 25.2 \\
    	TACT-B~\cite{bagad2023test} & 27.6 \\
    	\fadecell{ImageBind-H~\cite{girdhar2023imagebind}} & \fadecell{25.4} \\
      	\textbf{\OURS-B} & 31.3 \upcolor{3.7} \\
    	\textbf{\OURS-g} & \textbf{32.7} \upcolor{5.1} \\
      	\bottomrule
    \end{tabular}
    \end{scriptsize}
    \label{tbl:zs_cls:nextqa}
    \end{center}
\end{subtable}
\begin{subtable}[t]{.33\linewidth}
    \begin{center}
    \caption{Charades}
    \setlength{\tabcolsep}{4pt}
    \vspace{-0.3\baselineskip}
    \begin{scriptsize}
    \begin{tabular}{@{}lc@{}}
    	\toprule
    	Methods & mAP \\
    	\midrule
    	CLIP-B~\cite{radford2021learning} & 19.8 \\
    	CLIP-Hitchhiker-B~\cite{bain2022clip} & 21.1 \\
    	CoCa-g~\cite{yu2022coca} & 23.1 \\
    	VideoCoCa-g~\cite{yan2022videococa} & 25.8 \\
    	MAXI-B~\cite{lin2023match} & 23.8 \\
    	\textbf{\OURS-B} & 29.2 \upcolor{3.4} \\
    	\textbf{\OURS-g} & \textbf{32.4} \upcolor{6.6} \\
    	\bottomrule
    \end{tabular}
    \end{scriptsize}
    \label{tbl:zs_cls:ch}
    \end{center}
\end{subtable}
\begin{subtable}[t]{.28\linewidth}
    \begin{center}
    \caption{Charades-STA}
    \vspace{-0.3\baselineskip}
    \setlength{\tabcolsep}{4pt}
    \begin{scriptsize}
    \begin{tabular}{@{}lc@{}}
        \toprule Methods & MC Acc \\ 
      	\midrule
      	CoCa-g~\cite{yu2022coca} & 46.1 \\
    	VideoCoCa-g~\cite{yan2022videococa} & 47.2 \\
      	\textbf{\OURS-B} & 50.0 \upcolor{2.8} \\
    	\textbf{\OURS-g} & \textbf{50.4} \upcolor{3.2} \\
      	\bottomrule
    \end{tabular}
    \end{scriptsize}
    \label{tbl:zs_cls:chsta}
    \end{center}
\end{subtable}
\label{tbl:zs_cls}
\end{table*}

To enable zero-shot video-text retrieval and video classification capabilities of \OURS, we follow LiT~\cite{zhai2022lit} to learn a text encoder producing the text embeddings matched to their corresponding video embeddings out of \OURS. We choose the LiT text encoder to mirror the one in the first-stage training and attach an MAP head to the video encoder. The LiT tuning is over the same pretraining data from the first stage. More details are in~\cref{app:lit:implementation}.

\myparagraph{Datasets.} We evaluate \OURS's zero-shot video-text retrieval performance on three benchmarks: MSRVTT~\cite{xu2016msr,bain2021frozen}, VATEX~\cite{wang2019vatex}, and ActivityNet~\cite{krishna2017dense}. For zero-shot video classification tasks, we experiment with Kinetics-400~\cite{kay2017kinetics}, Charades~\cite{sigurdsson2016hollywood}, SSv2-Temporal and SSv2-Events~\cite{sevilla2021only,bagad2023test}, and the ATP-Hard subset of NExT-QA~\cite{buch2022revisiting}. 
SSv2 and NExT-QA (ATP-Hard) focus on motion and temporal reasoning, respectively. 
Moreover, we adapt Charades-STA~\cite{gao2017tall} to the zero-shot classification scenario by reformulating each of its samples in the test set into a multi-choice retrieval problem (see \cref{app:lit:charades_sta} for more details). 
We report results following the standard evaluation metric for each benchmark.

\myparagraph{Main results.} \cref{tbl:zs_retrieval,tbl:zs_cls} summarize the results of video-text retrieval and video classification, respectively. \OURS sets the new state of the art on most benchmarks, and the gains over the prior arts are exceptionally substantial on the challenging datasets (\eg, $9.5\%$ on ActivityNet, $4.4\%$ on SSv2-Events, and $6.6$ mAP on Charades). Most results from our base-scale \OURS-B are actually better than those of existing larger-scale models. Additionally, \OURS is on par with or better than the models pretrained with in-domain data and extra modalities (\eg, audios) in \cref{tbl:zs_cls}. These improvements in zero-shot retrieval and classification tasks present \OURS's strong generalization capabilities.

\subsection{Zero-shot video captioning and QA}\label{sec:exp:cap_vqa}

We further evaluate the inherent capabilities of \OURS on generative video-language tasks, \ie, captioning and QA, where we pair \OURS with a language decoder, PaLM-2~\cite{anil2023palm}. To connect the two models, we introduce and train several gluing layers while keeping both \OURS and the language decoder frozen. We then conduct evaluation under the zero-shot configuration on video captioning and QA benchmarks. Note that we do not tune our models separately for captioning and QA tasks. Please refer to \cref{app:videoulm} for implementation details.

\myparagraph{Datasets.} We evaluate the model in the zero-shot setting on the test splits of a suite of standard video captioning datasets including MSRVTT~\cite{xu2016msr}, VATEX~\cite{wang2019vatex}, and YouCook2~\cite{zhou2018towards}, and video QA benchmarks including MSRVTT-QA~\cite{xu2017video}, MSVD-QA~\cite{xu2017video}, and NExT-QA~\cite{xiao2021next}. For video QA, where it is imperative to match the length and style of the model's answers with groundtruths, we adopt the zero-shot approach of Flamingo~\cite{alayrac2022flamingo} and use two-shot text-only prompts from the training set of the downstream task. Additionally, for MSRVTT-QA and MSVD-QA, we experiment with the closed-vocabulary evaluation configuration~\cite{li2022blip,yang2022zero}. In this setting, we let the model score candidate answers according to their log-likelihoods and return the top one. 

\begin{table}[t]
\caption{\textbf{Comparison to state-of-the-art methods on zero-shot video captioning}. We report the CIDEr score for all benchmarks.}
\label{tbl:zs_cap}
\begin{center}
\begin{scriptsize}
\begin{tabular}{@{}l|ccc@{}}
    \toprule
    Methods & \textbf{MSRVTT} & \textbf{VATEX} & \textbf{YouCook2} \\
    \midrule
    \tabbar{4}{Captioning-only models} \\
    VideoCoCa-g~\cite{yan2022videococa} & 27.1 & 22.8 & 34.3 \\
    DeCap-B~\cite{li2023decap} & 18.6 & 18.7 & - \\
    \midrule
    \tabbar{4}{All-in-one models} \\
    Flamingo-3B~\cite{alayrac2022flamingo} & - & \textbf{40.1} & 55.8 \\
    Flamingo-9B~\cite{alayrac2022flamingo} & - & 39.5 & 55.0 \\
    \textbf{\OURS-B} w/ PaLM-2-1B & \textbf{40.3} \upcolor{13.} & 24.2 \downcolor{12.} & 52.3 \downcolor{3.5} \\
    \textbf{\OURS-B} w/ PaLM-2-8B & 38.5 \upcolor{11.} & 31.7 \downcolor{8.4} & \textbf{63.6} \upcolor{7.8} \\
    \bottomrule
\end{tabular}
\end{scriptsize}
\end{center}
\vskip -0.1in
\end{table}

\begin{table}[t]
\caption{\textbf{Comparison to state-of-the-art methods on zero-shot video QA}. We report the WUPS index~\cite{wu1994verb} for NExT-QA and Top-1 accuracy for the others. Methods that unfreeze their language models are marked in gray.}
\vskip -0.1in
\label{tbl:zs_vqa}
\begin{center}
\resizebox{\linewidth}{!}{
\begin{tabular}{@{}l|ccc@{}}
    \toprule
    Methods & \textbf{MSRVTT-QA} & \textbf{MSVD-QA} & \textbf{NExT-QA} \\
    \midrule
    \tabbar{4}{Question-answering-only models} \\
FrozenBiLM-L~\cite{yang2022zero} & 22.2 & 39.0 & - \\
    \midrule
    \tabbar{4}{All-in-one models} \\
    \fadecell{BLIP-B~\cite{li2022blip}} & \fadecell{19.2} & \fadecell{35.2} & \fadecell{-} \\
    \fadecell{HiTeA-B~\cite{ye2023hitea}} & \fadecell{21.7} & \fadecell{37.4} & \fadecell{-} \\
    \fadecell{mPLUG-2~\cite{xu2023mplug}} & \fadecell{43.8} & \fadecell{55.3} & \fadecell{-} \\
    Flamingo-3B~\cite{alayrac2022flamingo} & 11.0 & 27.5 & 21.3 \\
    Flamingo-9B~\cite{alayrac2022flamingo} & 13.7 & 30.2 & 23.0 \\
    \textbf{\OURS-B} w/ PaLM-2-1B & 28.5 \upcolor{6.3} & 39.5 \upcolor{0.5} & 23.8 \upcolor{0.8} \\
    \textbf{\OURS-B} w/ PaLM-2-8B & \textbf{32.0} \upcolor{9.8} & \textbf{47.1} \upcolor{8.1} & \textbf{27.4} \upcolor{4.4} \\
    \bottomrule
\end{tabular}}
\end{center}
\vskip -0.1in
\end{table}

\myparagraph{Main results.} \cref{tbl:zs_cap,tbl:zs_vqa} show the results of zero-shot video captioning and QA, respectively. Despite the simplicity of our model architecture and the small number of adapter parameters, our models are competitive and top the methods freezing both vision and language models except on VATEX. The results demonstrate that our \OURS encoder is able to generalize well to video-to-language generation tasks.

\subsection{CV for science tasks}
\label{sec:exp:ai4s}

While existing video analysis benchmarks commonly focus on human-centric data, we evaluate \OURS on a broad set of videos from scientific datasets to assess its generalizability and potential to be used in scientific applications.
These datasets include fields such as ethology~\cite{eyjolfsdottir2014detecting}, behavioral neuroscience~\cite{sun2021multi,burgos2012social}, cognitive science~\cite{ma2023chimpact}, and ecology~\cite{kholiavchenko2024kabr}. 
To the best of our knowledge, this work is the first to study the use of ViFMs on scientific datasets, highlighting their ability to match or surpass the performance of specialized models. We encourage the creation of more open-sourced datasets from real-world scientific experiments to unlock the potential of ViFMs to benefit various fields of science.

\myparagraph{Datasets.} 
We focus on large-scale video datasets annotated with domain expertise, captured in scientific experiments. These datasets consist of flies (Fly vs.\ Fly~\cite{eyjolfsdottir2014detecting}), mice  (CalMS21~\cite{sun2021multi}, CRIM13~\cite{burgos2012social}), chimpanzees (ChimpACT~\cite{ma2023chimpact}), and Kenyan animals (KABR~\cite{kholiavchenko2024kabr}). 
All the datasets are annotated for video classification of behavior, except for the ChimpACT dataset for spatiotemporal action localization.
We evaluate CRIM13 from cameras on the side perpendicular to the cage (``S''), as well as a top, overhead view (``T'').
We use standard data splits defined in previous works on these datasets, and all datasets are evaluated using the mAP metric, except KABR which uses macro-accuracy. Further implementation details are in \cref{app:cv4science}.

\myparagraph{Main results.} 
General ViFMs, using a shared frozen encoder across all evaluations, achieve performance comparable to (or exceeding) domain-specific models specialized for individual tasks (\cref{tbl:zs_ai4s}).  
In particular, \OURS generally performs the best and surpasses domain expert models with the base-scale model. Scaling to large-scale models further improves performance across all datasets.
These results demonstrate that ViFMs have the potential to significantly accelerate video analysis across diverse fields.

\begin{table*}[t]
\caption{\textbf{Comparison to state-of-the-art methods and domain experts on CV for Science benchmarks}. We report mean average precision (mAP) for all datasets, except for KABR which uses macro-accuracy.}
\label{tbl:zs_ai4s}
\begin{center}
\begin{scriptsize}
\begin{tabular}{@{}l|ccccc@{}}
    \toprule
    Methods & \textbf{Fly vs.\ Fly} & \textbf{CalMS21}  & \textbf{CRIM13 (S/T)} & \textbf{KABR} & \textbf{ChimpACT} \\
    \midrule
    \fadecell{Domain experts} & \fadecell{88.6} & \fadecell{88.9}  & \fadecell{-} & \fadecell{61.9} & \fadecell{24.4}\\
    \midrule
    \tabbar{6}{Base-scale models} \\
    CoCa-B~\cite{yu2022coca} & 80.1 & 89.2 & 58.2 / 58.4 & \textbf{62.0} &  12.6 \\
    InternVideo-B~\cite{wang2022internvideo} & 78.9 & 89.0 & 63.2 / 63.6 & 49.9 & 24.0 \\
    UMT-B~\cite{li2023unmasked} & 84.6 & 88.7 & 59.3 / 58.5 & 58.9 & 25.0\\
    \textbf{\OURS-B} & \textbf{89.1} \upcolor{4.5} & \textbf{91.1} \upcolor{0.9}   & \textbf{64.5} \upcolor{1.3} / \textbf{64.9} \upcolor{1.3} & 61.6 \downcolor{0.4} & \textbf{28.8} \upcolor{3.8}\\
    \midrule
    \tabbar{6}{Large-scale models} \\
InternVideo-L~\cite{wang2022internvideo} & 86.6 & \textbf{91.5}  & 65.7 / 65.2 & 51.4 & 25.7\\
    UMT-L~\cite{li2023unmasked} & 86.4 & 89.5  & 60.5 / 61.4 & 62.7 & 24.7 \\
    \textbf{\OURS-g} & \textbf{92.0} \upcolor{5.4} & \textbf{91.5} \upcolor{0.0}  & \textbf{65.9} \upcolor{0.2} / \textbf{66.8} \upcolor{1.6} & \textbf{63.3} \upcolor{0.6} & \textbf{31.5} \upcolor{5.8}\\ 
    \bottomrule
\end{tabular}
\end{scriptsize}
\end{center}
\vskip -0.1in
\end{table*}

\begin{figure}[t]
\begin{center}
\includegraphics[width=1\linewidth]{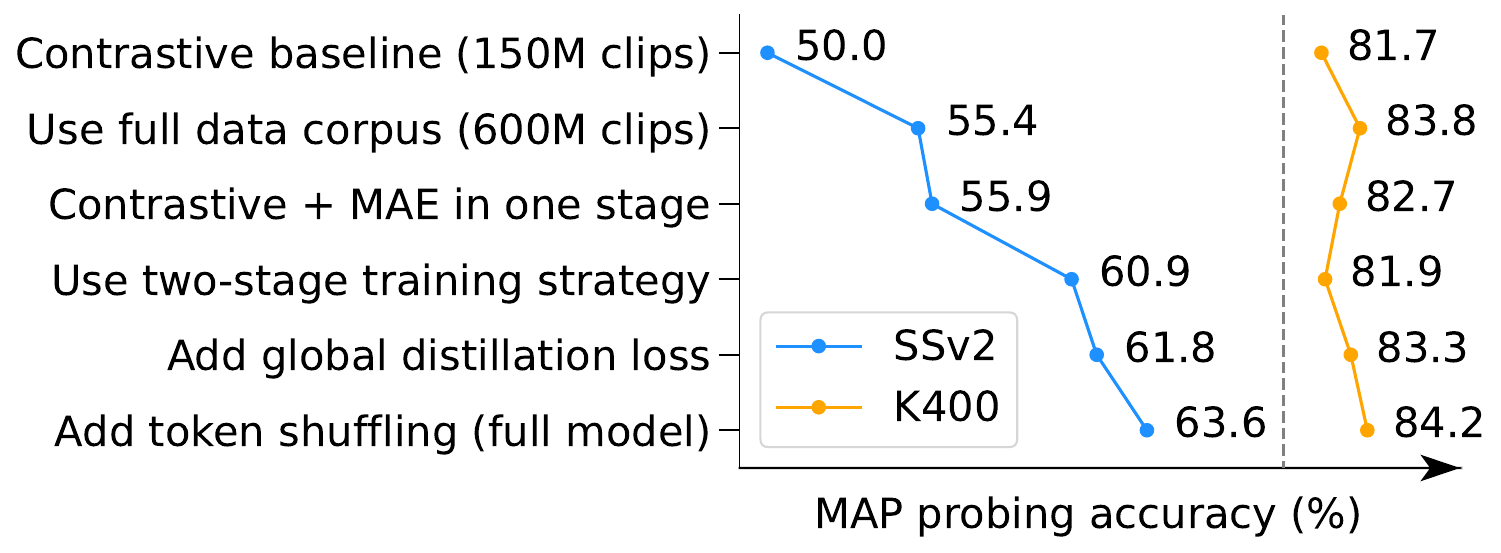} \\
\vskip -0.1in
\caption{\textbf{Ablation study.} From top to bottom: we begin by a video-text contrastive baseline and gradually add our major components to it. Each row is based on a modification of the immediately preceding row. We note that it is difficult to perform well on both K400 and SSv2 using only a single frozen encoder, but our final model with all improvements excels on both datasets.}
\label{fig:ablation_trajectory}
\end{center}
\vskip -0.1in
\end{figure}

\subsection{Ablation study} \label{sec:exp:ablate}
The main driving force behind \OURS includes the strategy and effort for collecting the pretraining data and the pretraining approach that improves upon masked autoencoding by the two-stage pretraining framework, the global distillation, and token shuffling. We run ablation studies to evaluate the effectiveness of these components. First, we train a video-text contrastive baseline as presented in~\cref{sec:app:train:teacher} over a smaller scale, publicly available corpus (150M video clips in total), including WTS-70M, YT-Temporal-180M, and InternVid. We then add our main components (larger pretraining data, two-stage training, losses, and token shuffling) to the baseline one at a time to see how the model performance evolves along the way. We also experiment with combining contrastive loss with masked autoencoding~\cite{feichtenhofer2022masked} in one stage to highlight the effectiveness of our two-stage training pipeline.

\cref{fig:ablation_trajectory} exhibits the ablation results, where we observe different performance evolving trajectories on motion-rich SSv2 and appearance-driven K400. Notably, the consistent improvements of \OURS on SSv2 suggest the effectiveness of our data curation and model designing efforts for facilitating motion understanding in videos. Although the contrastive baseline has already achieved competitive results on K400, the proposed global distillation and token shuffling further boost the accuracy. We provide more comprehensive ablation studies in \cref{app:ablation}.

\subsection{Limitations} 
One limitation of our approach is that we leverage a video corpus with noisy text as part of pretraining. This noisy text is potentially incomplete and biased, which could impact model performance. Moreover, long video understanding remains a challenge, since our current focus is on short video clips from which we sample $16$ frames as input to \OURS. Future work in this direction could leverage our encoder as part of a long video understanding system. Finally, while we advocate for the frozen-backbone evaluation, we acknowledge that there are scenarios that benefit more from end-to-end finetuning and parameter-efficient adaptation. Despite these limitations, the results demonstrate the potential impact of \OURS on a range of real-world video understanding tasks.
 \section{Related work}

\textbf{Foundation models (FMs)}~\cite{bommasani2021opportunities} have demonstrated tremendous promise with early work in LLMs~\cite{devlin2019bert,brown2020language}. Some ViFMs are built around LLMs~\cite{wang2022language,li2023blip,zhang2023video,chen2023videollm}, analyzing videos by feeding associated text to LLMs, such as ASR transcripts and machine-generated captions. In contrast, \OURS takes a video-centric view, and we aim to tackle a broader range of video understanding tasks.

\textbf{ViFMs.} Most recent FMs in CV focus on images~\cite{radford2021learning,yuan2021florence,jia2021scaling,yu2022coca,alayrac2022flamingo,yan2022videococa,wang2022omnivl,chen2023pali,xu2023mplug,girdhar2023imagebind,zhang2023cae,zhu2023languagebind}. Their pretraining data contains no or only a small portion of videos, and the model architectures and learning methods are for images by design. While these FMs can accept video frames as input, they fall short on motion and temporal modeling~\cite{yuan2023videoglue}. 
Our work directly addresses this gap by developing a video encoder designed for video-specific applications.

For videos, existing works mainly train FMs using self-supervised learning over the video-only modality~\cite{qian2021spatiotemporal,feichtenhofer2021large,recasens2021broaden,singh2021semi,wei2022masked,yuan2022contextualized,qian2022temporal,tong2022videomae,wang2023videomae} or video-language modeling of videos with noisy text~\cite{zellers2021merlot,fu2021violet,li2023lavender,wang2023all,cheng2023vindlu,piergiovanni2023mirasol3b,xiong2023spatiotemporally}. As \citet{wang2023paxion} point out, existing video-language models lack knowledge of actions, and yet self-supervised models from video-only data struggle with semantics. We instead bring the best of the two together. Related to our work, InternVideo~\cite{wang2022internvideo} glues a self-supervised VideoMAE model~\cite{wang2023videomae} and a video-language model together using cross-attention modules. Unlike \OURS, however, the two models have no mutual influence during pretraining and they redundantly process the same video from scratch simultaneously.

\textbf{Large-scale video datasets} are pivotal for ViFMs and have been a subject of interest. HowTo100M~\cite{miech2019howto100m}, YT-Temporal-1B~\cite{zellers2022merlot}, and HD-VILA-100M~\cite{xue2022advancing} associate speech transcriptions with videos. WebVid2M~\cite{bain2021frozen} and WTS70M~\cite{stroud2020learning} pair alt-text and other metadata with videos.  VideoCC3M~\cite{nagrani2022learning} retrieves videos that appear similar to images and transfer the image captions to corresponding videos.  VAST-27M~\cite{chen2023vast} and InternVid~\cite{ wang2023internvid} use  multi-modal and language models to caption videos. Still, these video-text datasets are significantly smaller than their counterparts for images, and many ViFMs adapt pretrained image-text models to the video space~\cite{fang2021clip2video,luo2022clip4clip,xue2022clipvip,liu2023revisiting,he2023vlab,wu2023building}. Our pretraining corpus has text associations from a hybrid mix of ASR transcripts, generated captions, and high-quality manually annotated captions.

\textbf{Pretraining strategy.} Our pretraining integrates vision-language contrastive learning~\cite{radford2021learning,xu2021videoclip,bain2022clip} and masked data modeling~\cite{devlin2019bert,he2022masked}. The former has led to strong late-fusion models like CLIP~\cite{radford2021learning}, ALIGN~\cite{jia2021scaling}, CoCa~\cite{yu2022coca}, and the latter is proven effective to learn from single-modality data like language~\cite{devlin2019bert,anil2023palm}, audio~\cite{borsos2023audiolm}, images~\cite{he2022masked,wang2023image,oquab2023dinov2}, and videos~\cite{tong2022videomae,wang2023videomae}. While EVA~\cite{fang2022eva,fang2023eva} and UMT~\cite{li2023unmasked} transfer indirect semantics from CLIP~\cite{radford2021learning} to masked modeling, we learn video-native semantics. We also introduce global distillation and token shuffling to the masked video modeling to orchestrate both appearance and motion cues.

 \section{Conclusion}

We present \OURS, a foundational video encoder that achieves state-of-the-art performance across a wide range of video understanding tasks.
Our design emphasizes both the data and modeling approach: we assemble the largest pretraining dataset of its kind, as well as develop a pretraining strategy that effectively learns appearance and motion information from it. In our comprehensive evaluation, \OURS achieves the best results on a majority of benchmarks. 
Notably, no other baseline models consistently achieve the second best, highlighting our unique generalizability.

\section*{Acknowledgements}

We sincerely thank David Hendon for their product management efforts and the help with open-sourcing, and Alex Siegman, Ramya Ganeshan, and Victor Gomes for their program and resource management efforts. We also thank Hassan Akbari, Sherry Ben, Yoni Ben-Meshulam, Chun-Te Chu, Sam Clearwater, Yin Cui, Sathish Thoppay Egambaram
, Ilya Figotin, Anja Hauth, Sergey Ioffe, Xuhui Jia, Lu Jiang, Zu Kim, Phoebe Kirk, Dan Kondratyuk, Yeqing Li, Bill Mark, Arsha Nagrani, Caroline Pantofaru, Sushant Prakash, Amanda Sadler, Rif A.\ Saurous, Cordelia Schmid, Bryan Seybold, Mojtaba Seyedhosseini, Rachel Stigler, Paul Voigtlaender, Pingmei Xu, Chaochao Yan, Xuan Yang, Rui Zhu, and Yukun Zhu for the discussions, support, and feedback that greatly contributed to this paper. Lastly, we thank Jay Yagnik, Rahul Sukthankar, and Tomas Izo for their enthusiastic support for this project.

\section*{Impact statement}

Advancements in video understanding have the potential to accelerate progress across various fields, including scientific research, education, robotics, healthcare, and content recommendation. 
These technologies could empower new scientific discoveries, enhance learning experiences, improve security and safety, and enable more responsive interactive systems.
However, it is crucial to address potential biases and misuses before one deploys related models to the real world.
This includes mitigating algorithmic biases, safeguarding privacy, and respecting rules and policies of responsible research. To ensure that the benefits of this technology are harnessed responsibly, we encourage continued open discussions in the community around the development of these new technologies.

\bibliography{flvid}

\begin{thebibliography}{148}
\providecommand{\natexlab}[1]{#1}
\providecommand{\url}[1]{\texttt{#1}}
\expandafter\ifx\csname urlstyle\endcsname\relax
  \providecommand{\doi}[1]{doi: #1}\else
  \providecommand{\doi}{doi: \begingroup \urlstyle{rm}\Url}\fi

\bibitem[Akbari et~al.(2021)Akbari, Yuan, Qian, Chuang, Chang, Cui, and
  Gong]{akbari2021vatt}
Akbari, H., Yuan, L., Qian, R., Chuang, W.-H., Chang, S.-F., Cui, Y., and Gong,
  B.
\newblock {VATT}: Transformers for multimodal self-supervised learning from raw
  video, audio and text.
\newblock In \emph{NeurIPS}, 2021.

\bibitem[Akbari et~al.(2023)Akbari, Kondratyuk, Cui, Hornung, Wang, and
  Adam]{akbari2023alternating}
Akbari, H., Kondratyuk, D., Cui, Y., Hornung, R., Wang, H., and Adam, H.
\newblock Alternating gradient descent and mixture-of-experts for integrated
  multimodal perception.
\newblock In \emph{NeurIPS}, 2023.

\bibitem[Alayrac et~al.(2022)Alayrac, Donahue, Luc, Miech, Barr, Hasson, Lenc,
  Mensch, Millican, Reynolds, et~al.]{alayrac2022flamingo}
Alayrac, J.-B., Donahue, J., Luc, P., Miech, A., Barr, I., Hasson, Y., Lenc,
  K., Mensch, A., Millican, K., Reynolds, M., et~al.
\newblock Flamingo: A visual language model for few-shot learning.
\newblock In \emph{NeurIPS}, 2022.

\bibitem[Anil et~al.(2023)Anil, Dai, Firat, Johnson, Lepikhin, Passos, Shakeri,
  Taropa, Bailey, Chen, et~al.]{anil2023palm}
Anil, R., Dai, A.~M., Firat, O., Johnson, M., Lepikhin, D., Passos, A.,
  Shakeri, S., Taropa, E., Bailey, P., Chen, Z., et~al.
\newblock {PaLM} 2 technical report.
\newblock \emph{arXiv preprint arXiv:2305.10403}, 2023.

\bibitem[Arnab et~al.(2021)Arnab, Dehghani, Heigold, Sun, Lu{\v{c}}i{\'c}, and
  Schmid]{arnab2021vivit}
Arnab, A., Dehghani, M., Heigold, G., Sun, C., Lu{\v{c}}i{\'c}, M., and Schmid,
  C.
\newblock {ViViT}: A video vision transformer.
\newblock In \emph{ICCV}, 2021.

\bibitem[Bagad et~al.(2023)Bagad, Tapaswi, and Snoek]{bagad2023test}
Bagad, P., Tapaswi, M., and Snoek, C.~G.
\newblock Test of time: Instilling video-language models with a sense of time.
\newblock In \emph{CVPR}, 2023.

\bibitem[Bain et~al.(2021)Bain, Nagrani, Varol, and Zisserman]{bain2021frozen}
Bain, M., Nagrani, A., Varol, G., and Zisserman, A.
\newblock Frozen in time: A joint video and image encoder for end-to-end
  retrieval.
\newblock In \emph{ICCV}, 2021.

\bibitem[Bain et~al.(2022)Bain, Nagrani, Varol, and Zisserman]{bain2022clip}
Bain, M., Nagrani, A., Varol, G., and Zisserman, A.
\newblock A {CLIP-Hitchhiker}'s guide to long video retrieval.
\newblock \emph{arXiv preprint arXiv:2205.08508}, 2022.

\bibitem[Bansal et~al.(2023)Bansal, Bitton, Szpektor, Chang, and
  Grover]{bansal2023videocon}
Bansal, H., Bitton, Y., Szpektor, I., Chang, K.-W., and Grover, A.
\newblock {VideoCon}: Robust video-language alignment via contrast captions.
\newblock \emph{arXiv preprint arXiv:2311.10111}, 2023.

\bibitem[Bao et~al.(2022)Bao, Dong, Piao, and Wei]{bao2022beit}
Bao, H., Dong, L., Piao, S., and Wei, F.
\newblock {BEiT}: {BERT} pre-training of image transformers.
\newblock In \emph{ICLR}, 2022.

\bibitem[Bommasani et~al.(2021)Bommasani, Hudson, Adeli, Altman, Arora, von
  Arx, Bernstein, Bohg, Bosselut, Brunskill,
  et~al.]{bommasani2021opportunities}
Bommasani, R., Hudson, D.~A., Adeli, E., Altman, R., Arora, S., von Arx, S.,
  Bernstein, M.~S., Bohg, J., Bosselut, A., Brunskill, E., et~al.
\newblock On the opportunities and risks of foundation models.
\newblock \emph{arXiv preprint arXiv:2108.07258}, 2021.

\bibitem[Borsos et~al.(2023)Borsos, Marinier, Vincent, Kharitonov, Pietquin,
  Sharifi, Roblek, Teboul, Grangier, Tagliasacchi, et~al.]{borsos2023audiolm}
Borsos, Z., Marinier, R., Vincent, D., Kharitonov, E., Pietquin, O., Sharifi,
  M., Roblek, D., Teboul, O., Grangier, D., Tagliasacchi, M., et~al.
\newblock {AudioLM}: A language modeling approach to audio generation.
\newblock \emph{IEEE/ACM Transactions on Audio, Speech, and Language
  Processing}, 31:\penalty0 2523--2533, 2023.

\bibitem[Brown et~al.(2020)Brown, Mann, Ryder, Subbiah, Kaplan, Dhariwal,
  Neelakantan, Shyam, Sastry, Askell, et~al.]{brown2020language}
Brown, T., Mann, B., Ryder, N., Subbiah, M., Kaplan, J.~D., Dhariwal, P.,
  Neelakantan, A., Shyam, P., Sastry, G., Askell, A., et~al.
\newblock Language models are few-shot learners.
\newblock In \emph{NeurIPS}, 2020.

\bibitem[Buch et~al.(2022)Buch, Eyzaguirre, Gaidon, Wu, Fei-Fei, and
  Niebles]{buch2022revisiting}
Buch, S., Eyzaguirre, C., Gaidon, A., Wu, J., Fei-Fei, L., and Niebles, J.~C.
\newblock Revisiting the ``video'' in video-language understanding.
\newblock In \emph{CVPR}, 2022.

\bibitem[Burgos-Artizzu et~al.(2012)Burgos-Artizzu, Doll{\'a}r, Lin, Anderson,
  and Perona]{burgos2012social}
Burgos-Artizzu, X.~P., Doll{\'a}r, P., Lin, D., Anderson, D.~J., and Perona, P.
\newblock Social behavior recognition in continuous video.
\newblock In \emph{CVPR}, 2012.

\bibitem[Caba~Heilbron et~al.(2015)Caba~Heilbron, Escorcia, Ghanem, and
  Carlos~Niebles]{caba2015activitynet}
Caba~Heilbron, F., Escorcia, V., Ghanem, B., and Carlos~Niebles, J.
\newblock {ActivityNet}: A large-scale video benchmark for human activity
  understanding.
\newblock In \emph{CVPR}, 2015.

\bibitem[Carreira et~al.(2018)Carreira, Noland, Banki-Horvath, Hillier, and
  Zisserman]{carreira2018short}
Carreira, J., Noland, E., Banki-Horvath, A., Hillier, C., and Zisserman, A.
\newblock A short note about {Kinetics-600}.
\newblock \emph{arXiv preprint arXiv:1808.01340}, 2018.

\bibitem[Chen et~al.(2023{\natexlab{a}})Chen, Zheng, Wang, Xu, Huang, Pan,
  Wang, Wang, Qiao, Lu, et~al.]{chen2023videollm}
Chen, G., Zheng, Y.-D., Wang, J., Xu, J., Huang, Y., Pan, J., Wang, Y., Wang,
  Y., Qiao, Y., Lu, T., et~al.
\newblock {VideoLLM}: Modeling video sequence with large language models.
\newblock \emph{arXiv preprint arXiv:2305.13292}, 2023{\natexlab{a}}.

\bibitem[Chen \& Huang(2021)Chen and Huang]{chen2021elaborative}
Chen, S. and Huang, D.
\newblock Elaborative rehearsal for zero-shot action recognition.
\newblock In \emph{ICCV}, 2021.

\bibitem[Chen et~al.(2023{\natexlab{b}})Chen, Li, Wang, Zhao, Sun, Zhu, and
  Liu]{chen2023vast}
Chen, S., Li, H., Wang, Q., Zhao, Z., Sun, M., Zhu, X., and Liu, J.
\newblock {VAST}: A vision-audio-subtitle-text omni-modality foundation model
  and dataset.
\newblock In \emph{NeurIPS}, 2023{\natexlab{b}}.

\bibitem[Chen et~al.(2023{\natexlab{c}})Chen, Wang, Changpinyo, Piergiovanni,
  Padlewski, Salz, Goodman, Grycner, Mustafa, Beyer, et~al.]{chen2023pali}
Chen, X., Wang, X., Changpinyo, S., Piergiovanni, A., Padlewski, P., Salz, D.,
  Goodman, S., Grycner, A., Mustafa, B., Beyer, L., et~al.
\newblock {PaLI}: A jointly-scaled multilingual language-image model.
\newblock In \emph{ICLR}, 2023{\natexlab{c}}.

\bibitem[Cheng et~al.(2023)Cheng, Wang, Lei, Crandall, Bansal, and
  Bertasius]{cheng2023vindlu}
Cheng, F., Wang, X., Lei, J., Crandall, D., Bansal, M., and Bertasius, G.
\newblock {VindLU}: A recipe for effective video-and-language pretraining.
\newblock In \emph{CVPR}, 2023.

\bibitem[Devlin et~al.(2019)Devlin, Chang, Lee, and Toutanova]{devlin2019bert}
Devlin, J., Chang, M.-W., Lee, K., and Toutanova, K.
\newblock {BERT}: Pre-training of deep bidirectional transformers for language
  understanding.
\newblock In \emph{NAACL-HLT}, 2019.

\bibitem[Dima et~al.(2022)Dima, Doughty, Farinella, Antonino, Evangelos, Ma,
  Davide, Munro, Toby, Price, et~al.]{dima2022rescaling}
Dima, D., Doughty, H., Farinella, G.~M., Antonino, F., Evangelos, K., Ma, J.,
  Davide, M., Munro, J., Toby, P., Price, W., et~al.
\newblock Rescaling egocentric vision: Collection, pipeline and challenges for
  {EPIC-KITCHENS-100}.
\newblock \emph{IJCV}, 130\penalty0 (1):\penalty0 33--55, 2022.

\bibitem[Dosovitskiy et~al.(2021)Dosovitskiy, Beyer, Kolesnikov, Weissenborn,
  Zhai, Unterthiner, Dehghani, Minderer, Heigold, Gelly,
  et~al.]{dosovitskiy2021image}
Dosovitskiy, A., Beyer, L., Kolesnikov, A., Weissenborn, D., Zhai, X.,
  Unterthiner, T., Dehghani, M., Minderer, M., Heigold, G., Gelly, S., et~al.
\newblock An image is worth 16x16 words: Transformers for image recognition at
  scale.
\newblock In \emph{ICLR}, 2021.

\bibitem[Eyjolfsdottir et~al.(2014)Eyjolfsdottir, Branson, Burgos-Artizzu,
  Hoopfer, Schor, Anderson, and Perona]{eyjolfsdottir2014detecting}
Eyjolfsdottir, E., Branson, S., Burgos-Artizzu, X.~P., Hoopfer, E.~D., Schor,
  J., Anderson, D.~J., and Perona, P.
\newblock Detecting social actions of fruit flies.
\newblock In \emph{ECCV}, 2014.

\bibitem[Fang et~al.(2021)Fang, Xiong, Xu, and Chen]{fang2021clip2video}
Fang, H., Xiong, P., Xu, L., and Chen, Y.
\newblock {CLIP2Video}: Mastering video-text retrieval via image {CLIP}.
\newblock \emph{arXiv preprint arXiv:2106.11097}, 2021.

\bibitem[Fang et~al.(2022)Fang, Wang, Xie, Sun, Wu, Wang, Huang, Wang, and
  Cao]{fang2022eva}
Fang, Y., Wang, W., Xie, B., Sun, Q.-S., Wu, L.~Y., Wang, X., Huang, T., Wang,
  X., and Cao, Y.
\newblock {EVA}: Exploring the limits of masked visual representation learning
  at scale.
\newblock In \emph{CVPR}, 2022.

\bibitem[Fang et~al.(2023)Fang, Sun, Wang, Huang, Wang, and Cao]{fang2023eva}
Fang, Y., Sun, Q., Wang, X., Huang, T., Wang, X., and Cao, Y.
\newblock {EVA-02}: A visual representation for neon genesis.
\newblock \emph{arXiv preprint arXiv:2303.11331}, 2023.

\bibitem[Feichtenhofer et~al.(2019)Feichtenhofer, Fan, Malik, and
  He]{feichtenhofer2018slowfast}
Feichtenhofer, C., Fan, H., Malik, J., and He, K.
\newblock {SlowFast} networks for video recognition.
\newblock In \emph{ICCV}, 2019.

\bibitem[Feichtenhofer et~al.(2021)Feichtenhofer, Fan, Xiong, Girshick, and
  He]{feichtenhofer2021large}
Feichtenhofer, C., Fan, H., Xiong, B., Girshick, R., and He, K.
\newblock A large-scale study on unsupervised spatiotemporal representation
  learning.
\newblock In \emph{CVPR}, 2021.

\bibitem[Feichtenhofer et~al.(2022)Feichtenhofer, Fan, Li, and
  He]{feichtenhofer2022masked}
Feichtenhofer, C., Fan, H., Li, Y., and He, K.
\newblock Masked autoencoders as spatiotemporal learners.
\newblock In \emph{NeurIPS}, 2022.

\bibitem[Fellbaum(2005)]{fellbaum2005wordnet}
Fellbaum, C.
\newblock {WordNet} and wordnets.
\newblock In Barber, A. (ed.), \emph{Encyclopedia of Language and Linguistics},
  pp.\  2--665. Elsevier, 2005.

\bibitem[Fu et~al.(2021)Fu, Li, Gan, Lin, Wang, Wang, and Liu]{fu2021violet}
Fu, T.-J., Li, L., Gan, Z., Lin, K., Wang, W.~Y., Wang, L., and Liu, Z.
\newblock {VIOLET}: End-to-end video-language transformers with masked
  visual-token modeling.
\newblock \emph{arXiv preprint arXiv:2111.12681}, 2021.

\bibitem[Gao et~al.(2017)Gao, Sun, Yang, and Nevatia]{gao2017tall}
Gao, J., Sun, C., Yang, Z., and Nevatia, R.
\newblock {TALL}: Temporal activity localization via language query.
\newblock In \emph{ICCV}, 2017.

\bibitem[Girdhar et~al.(2023)Girdhar, El-Nouby, Liu, Singh, Alwala, Joulin, and
  Misra]{girdhar2023imagebind}
Girdhar, R., El-Nouby, A., Liu, Z., Singh, M., Alwala, K.~V., Joulin, A., and
  Misra, I.
\newblock {ImageBind}: One embedding space to bind them all.
\newblock In \emph{CVPR}, 2023.

\bibitem[Goyal et~al.(2017{\natexlab{a}})Goyal, Ebrahimi~Kahou, Michalski,
  Materzynska, Westphal, Kim, Haenel, Fruend, Yianilos, Mueller-Freitag,
  et~al.]{goyal2017something}
Goyal, R., Ebrahimi~Kahou, S., Michalski, V., Materzynska, J., Westphal, S.,
  Kim, H., Haenel, V., Fruend, I., Yianilos, P., Mueller-Freitag, M., et~al.
\newblock The ``something something'' video database for learning and
  evaluating visual common sense.
\newblock In \emph{ICCV}, 2017{\natexlab{a}}.

\bibitem[Goyal et~al.(2017{\natexlab{b}})Goyal, Khot, Summers-Stay, Batra, and
  Parikh]{goyal2017making}
Goyal, Y., Khot, T., Summers-Stay, D., Batra, D., and Parikh, D.
\newblock Making the {V} in {VQA} matter: Elevating the role of image
  understanding in visual question answering.
\newblock In \emph{CVPR}, 2017{\natexlab{b}}.

\bibitem[Grauman et~al.(2022)Grauman, Westbury, Byrne, Chavis, Furnari,
  Girdhar, Hamburger, Jiang, Liu, Liu, et~al.]{grauman2022ego4d}
Grauman, K., Westbury, A., Byrne, E., Chavis, Z., Furnari, A., Girdhar, R.,
  Hamburger, J., Jiang, H., Liu, M., Liu, X., et~al.
\newblock {Ego4D}: Around the world in 3,000 hours of egocentric video.
\newblock In \emph{CVPR}, 2022.

\bibitem[Gu et~al.(2018)Gu, Sun, Ross, Vondrick, Pantofaru, Li,
  Vijayanarasimhan, Toderici, Ricco, Sukthankar, et~al.]{gu2018ava}
Gu, C., Sun, C., Ross, D.~A., Vondrick, C., Pantofaru, C., Li, Y.,
  Vijayanarasimhan, S., Toderici, G., Ricco, S., Sukthankar, R., et~al.
\newblock {AVA}: A video dataset of spatio-temporally localized atomic visual
  actions.
\newblock In \emph{CVPR}, 2018.

\bibitem[Gutmann \& Hyv{\"a}rinen(2010)Gutmann and
  Hyv{\"a}rinen]{gutmann2010noise}
Gutmann, M. and Hyv{\"a}rinen, A.
\newblock Noise-contrastive estimation: A new estimation principle for
  unnormalized statistical models.
\newblock In \emph{AISTATS}, 2010.

\bibitem[He et~al.(2022)He, Chen, Xie, Li, Doll{\'a}r, and
  Girshick]{he2022masked}
He, K., Chen, X., Xie, S., Li, Y., Doll{\'a}r, P., and Girshick, R.
\newblock Masked autoencoders are scalable vision learners.
\newblock In \emph{CVPR}, 2022.

\bibitem[He et~al.(2023)He, Chen, Ma, Huang, Jin, Liu, Fu, Yang, Liu, and
  Feng]{he2023vlab}
He, X., Chen, S., Ma, F., Huang, Z., Jin, X., Liu, Z., Fu, D., Yang, Y., Liu,
  J., and Feng, J.
\newblock {VLAB}: Enhancing video language pre-training by feature adapting and
  blending.
\newblock \emph{arXiv preprint arXiv:2305.13167}, 2023.

\bibitem[Hendricks \& Nematzadeh(2021)Hendricks and
  Nematzadeh]{hendricks2021probing}
Hendricks, L.~A. and Nematzadeh, A.
\newblock Probing image-language transformers for verb understanding.
\newblock In \emph{ACL}, 2021.

\bibitem[Hu et~al.(2022)Hu, Shen, Wallis, Allen-Zhu, Li, Wang, Wang, and
  Chen]{hu2021lora}
Hu, E.~J., Shen, Y., Wallis, P., Allen-Zhu, Z., Li, Y., Wang, S., Wang, L., and
  Chen, W.
\newblock {LoRA}: Low-rank adaptation of large language models.
\newblock In \emph{ICLR}, 2022.

\bibitem[Huang et~al.(2020)Huang, Pang, Zhu, Rivera, and
  Soricut]{huang2020multimodal}
Huang, G., Pang, B., Zhu, Z., Rivera, C., and Soricut, R.
\newblock Multimodal pretraining for dense video captioning.
\newblock \emph{arXiv preprint arXiv:2011.11760}, 2020.

\bibitem[Huang et~al.(2016)Huang, Ferraro, Mostafazadeh, Misra, Agrawal,
  Devlin, Girshick, He, Kohli, Batra, et~al.]{huang2016visual}
Huang, T.-H., Ferraro, F., Mostafazadeh, N., Misra, I., Agrawal, A., Devlin,
  J., Girshick, R., He, X., Kohli, P., Batra, D., et~al.
\newblock Visual storytelling.
\newblock In \emph{NAACL-HLT}, 2016.

\bibitem[Jain et~al.(2017)Jain, Kar, et~al.]{jain2017non}
Jain, P., Kar, P., et~al.
\newblock Non-convex optimization for machine learning.
\newblock \emph{Foundations and Trends{\textregistered} in Machine Learning},
  10\penalty0 (3-4):\penalty0 142--363, 2017.

\bibitem[Jia et~al.(2021)Jia, Yang, Xia, Chen, Parekh, Pham, Le, Sung, Li, and
  Duerig]{jia2021scaling}
Jia, C., Yang, Y., Xia, Y., Chen, Y.-T., Parekh, Z., Pham, H., Le, Q., Sung,
  Y.-H., Li, Z., and Duerig, T.
\newblock Scaling up visual and vision-language representation learning with
  noisy text supervision.
\newblock In \emph{ICML}, 2021.

\bibitem[Kay et~al.(2017)Kay, Carreira, Simonyan, Zhang, Hillier,
  Vijayanarasimhan, Viola, Green, Back, Natsev, et~al.]{kay2017kinetics}
Kay, W., Carreira, J., Simonyan, K., Zhang, B., Hillier, C., Vijayanarasimhan,
  S., Viola, F., Green, T., Back, T., Natsev, P., et~al.
\newblock The {Kinetics} human action video dataset.
\newblock \emph{arXiv preprint arXiv:1705.06950}, 2017.

\bibitem[Kholiavchenko et~al.(2024)Kholiavchenko, Kline, Ramirez, Stevens,
  Sheets, Babu, Banerji, Campolongo, Thompson, Van~Tiel,
  et~al.]{kholiavchenko2024kabr}
Kholiavchenko, M., Kline, J., Ramirez, M., Stevens, S., Sheets, A., Babu, R.,
  Banerji, N., Campolongo, E., Thompson, M., Van~Tiel, N., et~al.
\newblock {KABR}: In-situ dataset for kenyan animal behavior recognition from
  drone videos.
\newblock In \emph{WACV}, 2024.

\bibitem[Kingma \& Ba(2015)Kingma and Ba]{kingma2014adam}
Kingma, D.~P. and Ba, J.
\newblock Adam: A method for stochastic optimization.
\newblock In \emph{ICLR}, 2015.

\bibitem[Krishna et~al.(2017)Krishna, Hata, Ren, Fei-Fei, and
  Carlos~Niebles]{krishna2017dense}
Krishna, R., Hata, K., Ren, F., Fei-Fei, L., and Carlos~Niebles, J.
\newblock Dense-captioning events in videos.
\newblock In \emph{ICCV}, 2017.

\bibitem[Lee et~al.(2019)Lee, Lee, Kim, Kosiorek, Choi, and Teh]{lee2019set}
Lee, J., Lee, Y., Kim, J., Kosiorek, A., Choi, S., and Teh, Y.~W.
\newblock {Set Transformer}: A framework for attention-based
  permutation-invariant neural networks.
\newblock In \emph{ICML}, 2019.

\bibitem[Lei et~al.(2023)Lei, Berg, and Bansal]{lei2022revealing}
Lei, J., Berg, T.~L., and Bansal, M.
\newblock Revealing single frame bias for video-and-language learning.
\newblock In \emph{ACL}, 2023.

\bibitem[Li et~al.(2020)Li, Thotakuri, Ross, Carreira, Vostrikov, and
  Zisserman]{li2020ava}
Li, A., Thotakuri, M., Ross, D.~A., Carreira, J., Vostrikov, A., and Zisserman,
  A.
\newblock The {AVA-Kinetics} localized human actions video dataset.
\newblock \emph{arXiv preprint arXiv:2005.00214}, 2020.

\bibitem[Li et~al.(2022)Li, Li, Xiong, and Hoi]{li2022blip}
Li, J., Li, D., Xiong, C., and Hoi, S.
\newblock {BLIP}: Bootstrapping language-image pre-training for unified
  vision-language understanding and generation.
\newblock In \emph{ICML}, 2022.

\bibitem[Li et~al.(2023{\natexlab{a}})Li, Li, Savarese, and Hoi]{li2023blip}
Li, J., Li, D., Savarese, S., and Hoi, S.
\newblock {BLIP-2}: Bootstrapping language-image pre-training with frozen image
  encoders and large language models.
\newblock \emph{arXiv preprint arXiv:2301.12597}, 2023{\natexlab{a}}.

\bibitem[Li et~al.(2023{\natexlab{b}})Li, Wang, Li, Wang, He, Wang, and
  Qiao]{li2023unmasked}
Li, K., Wang, Y., Li, Y., Wang, Y., He, Y., Wang, L., and Qiao, Y.
\newblock Unmasked teacher: Towards training-efficient video foundation models.
\newblock In \emph{ICCV}, 2023{\natexlab{b}}.

\bibitem[Li et~al.(2023{\natexlab{c}})Li, Gan, Lin, Lin, Liu, Liu, and
  Wang]{li2023lavender}
Li, L., Gan, Z., Lin, K., Lin, C.-C., Liu, Z., Liu, C., and Wang, L.
\newblock {LAVENDER}: Unifying video-language understanding as masked language
  modeling.
\newblock In \emph{CVPR}, 2023{\natexlab{c}}.

\bibitem[Li et~al.(2023{\natexlab{d}})Li, Zhu, Wen, and Yang]{li2023decap}
Li, W., Zhu, L., Wen, L., and Yang, Y.
\newblock {DeCap}: Decoding {CLIP} latents for zero-shot captioning via
  text-only training.
\newblock In \emph{ICLR}, 2023{\natexlab{d}}.

\bibitem[Li et~al.(2018)Li, Li, and Vasconcelos]{li2018resound}
Li, Y., Li, Y., and Vasconcelos, N.
\newblock {RESOUND}: Towards action recognition without representation bias.
\newblock In \emph{ECCV}, 2018.

\bibitem[Li et~al.(2023{\natexlab{e}})Li, Fan, Hu, Feichtenhofer, and
  He]{li2023scaling}
Li, Y., Fan, H., Hu, R., Feichtenhofer, C., and He, K.
\newblock Scaling language-image pre-training via masking.
\newblock In \emph{CVPR}, 2023{\natexlab{e}}.

\bibitem[Li et~al.(2023{\natexlab{f}})Li, Wang, and Jia]{li2023llama}
Li, Y., Wang, C., and Jia, J.
\newblock {LLaMA-VID}: An image is worth 2 tokens in large language models.
\newblock \emph{arXiv preprint arXiv:2311.17043}, 2023{\natexlab{f}}.

\bibitem[Lin et~al.(2023{\natexlab{a}})Lin, Zhu, Ye, Ning, Jin, and
  Yuan]{lin2023video}
Lin, B., Zhu, B., Ye, Y., Ning, M., Jin, P., and Yuan, L.
\newblock {Video-LLaVA}: Learning united visual representation by alignment
  before projection.
\newblock \emph{arXiv preprint arXiv:2311.10122}, 2023{\natexlab{a}}.

\bibitem[Lin et~al.(2022)Lin, Wang, Soldan, Wray, Yan, XU, Gao, Tu, Zhao, Kong,
  et~al.]{lin2022egoclip}
Lin, K.~Q., Wang, J., Soldan, M., Wray, M., Yan, R., XU, E.~Z., Gao, D., Tu,
  R.-C., Zhao, W., Kong, W., et~al.
\newblock Egocentric video-language pretraining.
\newblock In \emph{NeurIPS}, 2022.

\bibitem[Lin et~al.(2023{\natexlab{b}})Lin, Karlinsky, Shvetsova, Possegger,
  Kozinski, Panda, Feris, Kuehne, and Bischof]{lin2023match}
Lin, W., Karlinsky, L., Shvetsova, N., Possegger, H., Kozinski, M., Panda, R.,
  Feris, R., Kuehne, H., and Bischof, H.
\newblock Match, expand and improve: Unsupervised finetuning for zero-shot
  action recognition with language knowledge.
\newblock In \emph{ICCV}, 2023{\natexlab{b}}.

\bibitem[Liu et~al.(2023)Liu, Huang, Li, Feng, Wu, and Li]{liu2023revisiting}
Liu, R., Huang, J., Li, G., Feng, J., Wu, X., and Li, T.~H.
\newblock Revisiting temporal modeling for clip-based image-to-video knowledge
  transferring.
\newblock In \emph{CVPR}, 2023.

\bibitem[Loshchilov \& Hutter(2019)Loshchilov and Hutter]{loshchilov2017adamw}
Loshchilov, I. and Hutter, F.
\newblock Decoupled weight decay regularization.
\newblock In \emph{ICLR}, 2019.

\bibitem[Luo et~al.(2022)Luo, Ji, Zhong, Chen, Lei, Duan, and
  Li]{luo2022clip4clip}
Luo, H., Ji, L., Zhong, M., Chen, Y., Lei, W., Duan, N., and Li, T.
\newblock {CLIP4Clip}: An empirical study of clip for end to end video clip
  retrieval and captioning.
\newblock \emph{Neurocomputing}, 508:\penalty0 293--304, 2022.

\bibitem[Ma et~al.(2023)Ma, Kaufhold, Su, Zhu, Terwilliger, Meza, Zhu, Rossano,
  and Wang]{ma2023chimpact}
Ma, X., Kaufhold, S.~P., Su, J., Zhu, W., Terwilliger, J., Meza, A., Zhu, Y.,
  Rossano, F., and Wang, Y.
\newblock {ChimpACT}: A longitudinal dataset for understanding chimpanzee
  behaviors.
\newblock \emph{arXiv preprint arXiv:2310.16447}, 2023.

\bibitem[Maaz et~al.(2023)Maaz, Rasheed, Khan, and Khan]{maaz2023video}
Maaz, M., Rasheed, H., Khan, S., and Khan, F.~S.
\newblock {Video-ChatGPT}: Towards detailed video understanding via large
  vision and language models.
\newblock \emph{arXiv preprint arXiv:2306.05424}, 2023.

\bibitem[McCloskey \& Cohen(1989)McCloskey and
  Cohen]{mccloskey1989catastrophic}
McCloskey, M. and Cohen, N.~J.
\newblock Catastrophic interference in connectionist networks: The sequential
  learning problem.
\newblock In \emph{Psychology of learning and motivation}, volume~24, pp.\
  109--165. Elsevier, 1989.

\bibitem[Miech et~al.(2019)Miech, Zhukov, Alayrac, Tapaswi, Laptev, and
  Sivic]{miech2019howto100m}
Miech, A., Zhukov, D., Alayrac, J.-B., Tapaswi, M., Laptev, I., and Sivic, J.
\newblock {HowTo100M}: Learning a text-video embedding by watching hundred
  million narrated video clips.
\newblock In \emph{ICCV}, 2019.

\bibitem[Momeni et~al.(2023)Momeni, Caron, Nagrani, Zisserman, and
  Schmid]{momeni2023verbs}
Momeni, L., Caron, M., Nagrani, A., Zisserman, A., and Schmid, C.
\newblock Verbs in action: Improving verb understanding in video-language
  models.
\newblock In \emph{ICCV}, 2023.

\bibitem[Monfort et~al.(2019)Monfort, Andonian, Zhou, Ramakrishnan, Bargal,
  Yan, Brown, Fan, Gutfreund, Vondrick, et~al.]{monfort2019moments}
Monfort, M., Andonian, A., Zhou, B., Ramakrishnan, K., Bargal, S.~A., Yan, T.,
  Brown, L., Fan, Q., Gutfreund, D., Vondrick, C., et~al.
\newblock {Moments in Time} dataset: one million videos for event
  understanding.
\newblock \emph{IEEE TPAMI}, 42\penalty0 (2):\penalty0 502--508, 2019.

\bibitem[Monfort et~al.(2021)Monfort, Jin, Liu, Harwath, Feris, Glass, and
  Oliva]{monfort2021spoken}
Monfort, M., Jin, S., Liu, A., Harwath, D., Feris, R., Glass, J., and Oliva, A.
\newblock {Spoken Moments}: Learning joint audio-visual representations from
  video descriptions.
\newblock In \emph{CVPR}, 2021.

\bibitem[Nagrani et~al.(2022)Nagrani, Seo, Seybold, Hauth, Manen, Sun, and
  Schmid]{nagrani2022learning}
Nagrani, A., Seo, P.~H., Seybold, B., Hauth, A., Manen, S., Sun, C., and
  Schmid, C.
\newblock Learning audio-video modalities from image captions.
\newblock In \emph{ECCV}, 2022.

\bibitem[Ni et~al.(2022)Ni, Peng, Chen, Zhang, Meng, Fu, Xiang, and
  Ling]{ni2022expanding}
Ni, B., Peng, H., Chen, M., Zhang, S., Meng, G., Fu, J., Xiang, S., and Ling,
  H.
\newblock Expanding language-image pretrained models for general video
  recognition.
\newblock In \emph{ECCV}, 2022.

\bibitem[Noroozi \& Favaro(2016)Noroozi and Favaro]{noroozi2016unsupervised}
Noroozi, M. and Favaro, P.
\newblock Unsupervised learning of visual representations by solving {Jigsaw}
  puzzles.
\newblock In \emph{ECCV}, 2016.

\bibitem[Oquab et~al.(2023)Oquab, Darcet, Moutakanni, Vo, Szafraniec, Khalidov,
  Fernandez, Haziza, Massa, El-Nouby, et~al.]{oquab2023dinov2}
Oquab, M., Darcet, T., Moutakanni, T., Vo, H., Szafraniec, M., Khalidov, V.,
  Fernandez, P., Haziza, D., Massa, F., El-Nouby, A., et~al.
\newblock {DINOv2}: Learning robust visual features without supervision.
\newblock \emph{arXiv preprint arXiv:2304.07193}, 2023.

\bibitem[Peng et~al.(2022)Peng, Dong, Bao, Ye, and Wei]{peng2022beit}
Peng, Z., Dong, L., Bao, H., Ye, Q., and Wei, F.
\newblock {BEiT} v2: Masked image modeling with vector-quantized visual
  tokenizers.
\newblock \emph{arXiv preprint arXiv:2208.06366}, 2022.

\bibitem[Piergiovanni et~al.(2023)Piergiovanni, Nobel, Kim, Ryoo, Gomes, and
  Angelova]{piergiovanni2023mirasol3b}
Piergiovanni, A., Nobel, I., Kim, D., Ryoo, M.~S., Gomes, V., and Angelova, A.
\newblock {Mirasol3B}: A multimodal autoregressive model for time-aligned and
  contextual modalities.
\newblock \emph{arXiv preprint arXiv:2311.05698}, 2023.

\bibitem[Pitcher-Cooper et~al.(2023)Pitcher-Cooper, Seth, Kao, Coughlan, and
  Yoon]{pitcher2023you}
Pitcher-Cooper, C., Seth, M., Kao, B., Coughlan, J.~M., and Yoon, I.
\newblock {You Described, We Archived}: A rich audio description dataset.
\newblock \emph{Journal on Technology and Persons with Disabilities}, 2023.

\bibitem[Qian et~al.(2021)Qian, Meng, Gong, Yang, Wang, Belongie, and
  Cui]{qian2021spatiotemporal}
Qian, R., Meng, T., Gong, B., Yang, M.-H., Wang, H., Belongie, S., and Cui, Y.
\newblock Spatiotemporal contrastive video representation learning.
\newblock In \emph{CVPR}, 2021.

\bibitem[Qian et~al.(2022)Qian, Li, Yuan, Gong, Liu, Brown, Yang, Adam, and
  Cui]{qian2022temporal}
Qian, R., Li, Y., Yuan, L., Gong, B., Liu, T., Brown, M., Yang, M.-H., Adam,
  H., and Cui, Y.
\newblock On temporal granularity in self-supervised video representation
  learning.
\newblock In \emph{BMVC}, 2022.

\bibitem[Radford et~al.(2021)Radford, Kim, Hallacy, Ramesh, Goh, Agarwal,
  Sastry, Askell, Mishkin, Clark, et~al.]{radford2021learning}
Radford, A., Kim, J.~W., Hallacy, C., Ramesh, A., Goh, G., Agarwal, S., Sastry,
  G., Askell, A., Mishkin, P., Clark, J., et~al.
\newblock Learning transferable visual models from natural language
  supervision.
\newblock In \emph{ICML}, 2021.

\bibitem[Recasens et~al.(2021)Recasens, Luc, Alayrac, Wang, Strub, Tallec,
  Malinowski, P{\u{a}}tr{\u{a}}ucean, Altch{\'e}, Valko,
  et~al.]{recasens2021broaden}
Recasens, A., Luc, P., Alayrac, J.-B., Wang, L., Strub, F., Tallec, C.,
  Malinowski, M., P{\u{a}}tr{\u{a}}ucean, V., Altch{\'e}, F., Valko, M., et~al.
\newblock Broaden your views for self-supervised video learning.
\newblock In \emph{ICCV}, 2021.

\bibitem[Regneri et~al.(2013)Regneri, Rohrbach, Wetzel, Thater, Schiele, and
  Pinkal]{regneri2013groundingtacos}
Regneri, M., Rohrbach, M., Wetzel, D., Thater, S., Schiele, B., and Pinkal, M.
\newblock Grounding action descriptions in videos.
\newblock \emph{Transactions of the Association for Computational Linguistics},
  1:\penalty0 25--36, 2013.

\bibitem[Ren et~al.(2015)Ren, He, Girshick, and Sun]{ren2015fasterrcnn}
Ren, S., He, K., Girshick, R., and Sun, J.
\newblock Faster {R-CNN}: Towards real-time object detection with region
  proposal networks.
\newblock In \emph{NeurIPS}, 2015.

\bibitem[Sevilla-Lara et~al.(2021)Sevilla-Lara, Zha, Yan, Goswami, Feiszli, and
  Torresani]{sevilla2021only}
Sevilla-Lara, L., Zha, S., Yan, Z., Goswami, V., Feiszli, M., and Torresani, L.
\newblock Only time can tell: Discovering temporal data for temporal modeling.
\newblock In \emph{WACV}, 2021.

\bibitem[Shazeer \& Stern(2018)Shazeer and Stern]{shazeer2018adafactor}
Shazeer, N. and Stern, M.
\newblock Adafactor: Adaptive learning rates with sublinear memory cost.
\newblock In \emph{ICML}, 2018.

\bibitem[Sigurdsson et~al.(2016)Sigurdsson, Varol, Wang, Farhadi, Laptev, and
  Gupta]{sigurdsson2016hollywood}
Sigurdsson, G.~A., Varol, G., Wang, X., Farhadi, A., Laptev, I., and Gupta, A.
\newblock {Hollywood in Homes}: Crowdsourcing data collection for activity
  understanding.
\newblock In \emph{ECCV}, 2016.

\bibitem[Singh et~al.(2021)Singh, Chakraborty, Varshney, Panda, Feris, Saenko,
  and Das]{singh2021semi}
Singh, A., Chakraborty, O., Varshney, A., Panda, R., Feris, R., Saenko, K., and
  Das, A.
\newblock Semi-supervised action recognition with temporal contrastive
  learning.
\newblock In \emph{CVPR}, 2021.

\bibitem[Singh et~al.(2022)Singh, Hu, Goswami, Couairon, Galuba, Rohrbach, and
  Kiela]{singh2022flava}
Singh, A., Hu, R., Goswami, V., Couairon, G., Galuba, W., Rohrbach, M., and
  Kiela, D.
\newblock {FLAVA}: A foundational language and vision alignment model.
\newblock In \emph{CVPR}, 2022.

\bibitem[Stroud et~al.(2020)Stroud, Lu, Sun, Deng, Sukthankar, Schmid, and
  Ross]{stroud2020learning}
Stroud, J.~C., Lu, Z., Sun, C., Deng, J., Sukthankar, R., Schmid, C., and Ross,
  D.~A.
\newblock Learning video representations from textual web supervision.
\newblock \emph{arXiv preprint arXiv:2007.14937}, 2020.

\bibitem[Sun et~al.(2021{\natexlab{a}})Sun, Karigo, Chakraborty, Mohanty, Wild,
  Sun, Chen, Anderson, Perona, Yue, et~al.]{sun2021multi}
Sun, J.~J., Karigo, T., Chakraborty, D., Mohanty, S.~P., Wild, B., Sun, Q.,
  Chen, C., Anderson, D.~J., Perona, P., Yue, Y., et~al.
\newblock The multi-agent behavior dataset: Mouse dyadic social interactions.
\newblock In \emph{NeurIPS D\&B}, 2021{\natexlab{a}}.

\bibitem[Sun et~al.(2021{\natexlab{b}})Sun, Kennedy, Zhan, Anderson, Yue, and
  Perona]{sun2021task}
Sun, J.~J., Kennedy, A., Zhan, E., Anderson, D.~J., Yue, Y., and Perona, P.
\newblock Task programming: Learning data efficient behavior representations.
\newblock In \emph{CVPR}, 2021{\natexlab{b}}.

\bibitem[Tan et~al.(2020)Tan, Wang, Li, Li, Ouyang, Yin, and
  Yan]{tan2020equalization}
Tan, J., Wang, C., Li, B., Li, Q., Ouyang, W., Yin, C., and Yan, J.
\newblock Equalization loss for long-tailed object recognition.
\newblock In \emph{CVPR}, 2020.

\bibitem[Tang et~al.(2019)Tang, Ding, Rao, Zheng, Zhang, Zhao, Lu, and
  Zhou]{tang2019coin}
Tang, Y., Ding, D., Rao, Y., Zheng, Y., Zhang, D., Zhao, L., Lu, J., and Zhou,
  J.
\newblock {COIN}: A large-scale dataset for comprehensive instructional video
  analysis.
\newblock In \emph{CVPR}, 2019.

\bibitem[Tang et~al.(2023)Tang, Bi, Xu, Song, Liang, Wang, Zhang, An, Lin, Zhu,
  et~al.]{tang2023video}
Tang, Y., Bi, J., Xu, S., Song, L., Liang, S., Wang, T., Zhang, D., An, J.,
  Lin, J., Zhu, R., et~al.
\newblock Video understanding with large language models: A survey.
\newblock \emph{arXiv preprint arXiv:2312.17432}, 2023.

\bibitem[Tong et~al.(2022)Tong, Song, Wang, and Wang]{tong2022videomae}
Tong, Z., Song, Y., Wang, J., and Wang, L.
\newblock {VideoMAE}: Masked autoencoders are data-efficient learners for
  self-supervised video pre-training.
\newblock In \emph{NeurIPS}, 2022.

\bibitem[Vaswani et~al.(2017)Vaswani, Shazeer, Parmar, Uszkoreit, Jones, Gomez,
  Kaiser, and Polosukhin]{vaswani2017attention}
Vaswani, A., Shazeer, N., Parmar, N., Uszkoreit, J., Jones, L., Gomez, A.~N.,
  Kaiser, {\L}., and Polosukhin, I.
\newblock Attention is all you need.
\newblock In \emph{NeurIPS}, 2017.

\bibitem[Voigtlaender et~al.(2023)Voigtlaender, Changpinyo, Pont-Tuset,
  Soricut, and Ferrari]{voigtlaender2023connecting}
Voigtlaender, P., Changpinyo, S., Pont-Tuset, J., Soricut, R., and Ferrari, V.
\newblock Connecting vision and language with video localized narratives.
\newblock In \emph{CVPR}, 2023.

\bibitem[Wang et~al.(2022{\natexlab{a}})Wang, Chen, Wu, Luo, Zhou, Zhao, Xie,
  Liu, Jiang, and Yuan]{wang2022omnivl}
Wang, J., Chen, D., Wu, Z., Luo, C., Zhou, L., Zhao, Y., Xie, Y., Liu, C.,
  Jiang, Y.-G., and Yuan, L.
\newblock {OmniVL}: One foundation model for image-language and video-language
  tasks.
\newblock In \emph{NeurIPS}, 2022{\natexlab{a}}.

\bibitem[Wang et~al.(2023{\natexlab{a}})Wang, Ge, Yan, Ge, Lin, Tsutsui, Lin,
  Cai, Wu, Shan, et~al.]{wang2023all}
Wang, J., Ge, Y., Yan, R., Ge, Y., Lin, K.~Q., Tsutsui, S., Lin, X., Cai, G.,
  Wu, J., Shan, Y., et~al.
\newblock All in one: Exploring unified video-language pre-training.
\newblock In \emph{CVPR}, 2023{\natexlab{a}}.

\bibitem[Wang et~al.(2023{\natexlab{b}})Wang, Huang, Zhao, Tong, He, Wang,
  Wang, and Qiao]{wang2023videomae}
Wang, L., Huang, B., Zhao, Z., Tong, Z., He, Y., Wang, Y., Wang, Y., and Qiao,
  Y.
\newblock {VideoMAE} v2: Scaling video masked autoencoders with dual masking.
\newblock In \emph{CVPR}, 2023{\natexlab{b}}.

\bibitem[Wang et~al.(2022{\natexlab{b}})Wang, Chen, Wu, Chen, Dai, Liu, Jiang,
  Zhou, and Yuan]{wang2022bevt}
Wang, R., Chen, D., Wu, Z., Chen, Y., Dai, X., Liu, M., Jiang, Y.-G., Zhou, L.,
  and Yuan, L.
\newblock {BEVT}: {BERT} pretraining of video transformers.
\newblock In \emph{CVPR}, 2022{\natexlab{b}}.

\bibitem[Wang et~al.(2023{\natexlab{c}})Wang, Chen, Wu, Chen, Dai, Liu, Yuan,
  and Jiang]{wang2023masked}
Wang, R., Chen, D., Wu, Z., Chen, Y., Dai, X., Liu, M., Yuan, L., and Jiang,
  Y.-G.
\newblock Masked video distillation: Rethinking masked feature modeling for
  self-supervised video representation learning.
\newblock In \emph{CVPR}, 2023{\natexlab{c}}.

\bibitem[Wang et~al.(2023{\natexlab{d}})Wang, Bao, Dong, Bjorck, Peng, Liu,
  Aggarwal, Mohammed, Singhal, Som, et~al.]{wang2023image}
Wang, W., Bao, H., Dong, L., Bjorck, J., Peng, Z., Liu, Q., Aggarwal, K.,
  Mohammed, O.~K., Singhal, S., Som, S., et~al.
\newblock Image as a foreign language: {BEiT} pretraining for vision and
  vision-language tasks.
\newblock In \emph{CVPR}, 2023{\natexlab{d}}.

\bibitem[Wang et~al.(2019)Wang, Wu, Chen, Li, Wang, and Wang]{wang2019vatex}
Wang, X., Wu, J., Chen, J., Li, L., Wang, Y.-F., and Wang, W.~Y.
\newblock {VATEX}: A large-scale, high-quality multilingual dataset for
  video-and-language research.
\newblock In \emph{ICCV}, 2019.

\bibitem[Wang et~al.(2022{\natexlab{c}})Wang, Li, Li, He, Huang, Zhao, Zhang,
  Xu, Liu, Wang, et~al.]{wang2022internvideo}
Wang, Y., Li, K., Li, Y., He, Y., Huang, B., Zhao, Z., Zhang, H., Xu, J., Liu,
  Y., Wang, Z., et~al.
\newblock {InternVideo}: General video foundation models via generative and
  discriminative learning.
\newblock \emph{arXiv preprint arXiv:2212.03191}, 2022{\natexlab{c}}.

\bibitem[Wang et~al.(2023{\natexlab{e}})Wang, He, Li, Li, Yu, Ma, Chen, Wang,
  Luo, Liu, et~al.]{wang2023internvid}
Wang, Y., He, Y., Li, Y., Li, K., Yu, J., Ma, X., Chen, X., Wang, Y., Luo, P.,
  Liu, Z., et~al.
\newblock {InternVid}: A large-scale video-text dataset for multimodal
  understanding and generation.
\newblock \emph{arXiv preprint arXiv:2307.06942}, 2023{\natexlab{e}}.

\bibitem[Wang et~al.(2022{\natexlab{d}})Wang, Li, Xu, Zhou, Lei, Lin, Wang,
  Yang, Zhu, Hoiem, et~al.]{wang2022language}
Wang, Z., Li, M., Xu, R., Zhou, L., Lei, J., Lin, X., Wang, S., Yang, Z., Zhu,
  C., Hoiem, D., et~al.
\newblock Language models with image descriptors are strong few-shot
  video-language learners.
\newblock In \emph{NeurIPS}, 2022{\natexlab{d}}.

\bibitem[Wang et~al.(2023{\natexlab{f}})Wang, Blume, Li, Liu, Cho, Tang,
  Bansal, and Ji]{wang2023paxion}
Wang, Z., Blume, A., Li, S., Liu, G., Cho, J., Tang, Z., Bansal, M., and Ji, H.
\newblock Paxion: Patching action knowledge in video-language foundation
  models.
\newblock In \emph{NeurIPS}, 2023{\natexlab{f}}.

\bibitem[Wei et~al.(2022)Wei, Fan, Xie, Wu, Yuille, and
  Feichtenhofer]{wei2022masked}
Wei, C., Fan, H., Xie, S., Wu, C.-Y., Yuille, A., and Feichtenhofer, C.
\newblock Masked feature prediction for self-supervised visual pre-training.
\newblock In \emph{CVPR}, 2022.

\bibitem[Wu et~al.(2021)Wu, Huang, Zhang, Li, Ji, Yang, Sapiro, and
  Duan]{wu2021godiva}
Wu, C., Huang, L., Zhang, Q., Li, B., Ji, L., Yang, F., Sapiro, G., and Duan,
  N.
\newblock Godiva: Generating open-domain videos from natural descriptions.
\newblock \emph{arXiv preprint arXiv:2104.14806}, 2021.

\bibitem[Wu et~al.(2023)Wu, Sun, and Ouyang]{wu2023revisiting}
Wu, W., Sun, Z., and Ouyang, W.
\newblock Revisiting classifier: Transferring vision-language models for video
  recognition.
\newblock In \emph{AAAI}, 2023.

\bibitem[Wu \& Palmer(1994)Wu and Palmer]{wu1994verb}
Wu, Z. and Palmer, M.
\newblock Verb semantics and lexical selection.
\newblock In \emph{ACL}, 1994.

\bibitem[Wu et~al.(2024)Wu, Weng, Peng, Yang, Li, Davis, and
  Jiang]{wu2023building}
Wu, Z., Weng, Z., Peng, W., Yang, X., Li, A., Davis, L.~S., and Jiang, Y.-G.
\newblock Building an open-vocabulary video {CLIP} model with better
  architectures, optimization and data.
\newblock \emph{IEEE TPAMI}, 2024.

\bibitem[Xiao et~al.(2021)Xiao, Shang, Yao, and Chua]{xiao2021next}
Xiao, J., Shang, X., Yao, A., and Chua, T.-S.
\newblock {NExT-QA}: Next phase of question-answering to explaining temporal
  actions.
\newblock In \emph{CVPR}, 2021.

\bibitem[Xiong et~al.(2023)Xiong, Zhao, Gong, Yang, Schroff, Liu, Hsieh, and
  Yuan]{xiong2023spatiotemporally}
Xiong, Y., Zhao, L., Gong, B., Yang, M.-H., Schroff, F., Liu, T., Hsieh, C.-J.,
  and Yuan, L.
\newblock Spatiotemporally discriminative video-language pre-training with text
  grounding.
\newblock \emph{arXiv preprint arXiv:2303.16341}, 2023.

\bibitem[Xu et~al.(2017)Xu, Zhao, Xiao, Wu, Zhang, He, and Zhuang]{xu2017video}
Xu, D., Zhao, Z., Xiao, J., Wu, F., Zhang, H., He, X., and Zhuang, Y.
\newblock Video question answering via gradually refined attention over
  appearance and motion.
\newblock In \emph{ACM MM}, 2017.

\bibitem[Xu et~al.(2021)Xu, Ghosh, Huang, Okhonko, Aghajanyan, Metze,
  Zettlemoyer, and Feichtenhofer]{xu2021videoclip}
Xu, H., Ghosh, G., Huang, P.-Y., Okhonko, D., Aghajanyan, A., Metze, F.,
  Zettlemoyer, L., and Feichtenhofer, C.
\newblock {VideoCLIP}: Contrastive pre-training for zero-shot video-text
  understanding.
\newblock In \emph{EMNLP}, 2021.

\bibitem[Xu et~al.(2023)Xu, Ye, Yan, Shi, Ye, Xu, Li, Bi, Qian, Wang,
  et~al.]{xu2023mplug}
Xu, H., Ye, Q., Yan, M., Shi, Y., Ye, J., Xu, Y., Li, C., Bi, B., Qian, Q.,
  Wang, W., et~al.
\newblock {mPLUG-2}: A modularized multi-modal foundation model across text,
  image and video.
\newblock In \emph{ICML}, 2023.

\bibitem[Xu et~al.(2016)Xu, Mei, Yao, and Rui]{xu2016msr}
Xu, J., Mei, T., Yao, T., and Rui, Y.
\newblock {MSR-VTT}: A large video description dataset for bridging video and
  language.
\newblock In \emph{CVPR}, 2016.

\bibitem[Xu et~al.(2020)Xu, Zhao, Rojas, Thabet, and Ghanem]{xu2019gtad}
Xu, M., Zhao, C., Rojas, D.~S., Thabet, A., and Ghanem, B.
\newblock {G-TAD}: Sub-graph localization for temporal action detection.
\newblock In \emph{CVPR}, 2020.

\bibitem[Xue et~al.(2022)Xue, Hang, Zeng, Sun, Liu, Yang, Fu, and
  Guo]{xue2022advancing}
Xue, H., Hang, T., Zeng, Y., Sun, Y., Liu, B., Yang, H., Fu, J., and Guo, B.
\newblock Advancing high-resolution video-language representation with
  large-scale video transcriptions.
\newblock In \emph{CVPR}, 2022.

\bibitem[Xue et~al.(2023)Xue, Sun, Liu, Fu, Song, Li, and Luo]{xue2022clipvip}
Xue, H., Sun, Y., Liu, B., Fu, J., Song, R., Li, H., and Luo, J.
\newblock {CLIP-ViP}: Adapting pre-trained image-text model to video-language
  representation alignment.
\newblock In \emph{ICLR}, 2023.

\bibitem[Yan et~al.(2022)Yan, Zhu, Wang, Cao, Zhang, Ghosh, Wu, and
  Yu]{yan2022videococa}
Yan, S., Zhu, T., Wang, Z., Cao, Y., Zhang, M., Ghosh, S., Wu, Y., and Yu, J.
\newblock {VideoCoCa}: Video-text modeling with zero-shot transfer from
  contrastive captioners.
\newblock \emph{arXiv preprint arXiv:2212.04979}, 2022.

\bibitem[Yang et~al.(2022)Yang, Miech, Sivic, Laptev, and Schmid]{yang2022zero}
Yang, A., Miech, A., Sivic, J., Laptev, I., and Schmid, C.
\newblock Zero-shot video question answering via frozen bidirectional language
  models.
\newblock In \emph{NeurIPS}, 2022.

\bibitem[Yarom et~al.(2023)Yarom, Bitton, Changpinyo, Aharoni, Herzig, Lang,
  Ofek, and Szpektor]{yarom2023you}
Yarom, M., Bitton, Y., Changpinyo, S., Aharoni, R., Herzig, J., Lang, O., Ofek,
  E., and Szpektor, I.
\newblock What you see is what you read? improving text-image alignment
  evaluation.
\newblock In \emph{NeurIPS}, 2023.

\bibitem[Ye et~al.(2023)Ye, Xu, Yan, Xu, Qian, Zhang, and Huang]{ye2023hitea}
Ye, Q., Xu, G., Yan, M., Xu, H., Qian, Q., Zhang, J., and Huang, F.
\newblock {HiTeA}: Hierarchical temporal-aware video-language pre-training.
\newblock In \emph{ICCV}, 2023.

\bibitem[Yu et~al.(2022)Yu, Wang, Vasudevan, Yeung, Seyedhosseini, and
  Wu]{yu2022coca}
Yu, J., Wang, Z., Vasudevan, V., Yeung, L., Seyedhosseini, M., and Wu, Y.
\newblock {CoCa}: Contrastive captioners are image-text foundation models.
\newblock \emph{TMLR}, 2022.
\newblock ISSN 2835-8856.
\newblock URL \url{https://openreview.net/forum?id=Ee277P3AYC}.

\bibitem[Yuan et~al.(2021)Yuan, Chen, Chen, Codella, Dai, Gao, Hu, Huang, Li,
  Li, et~al.]{yuan2021florence}
Yuan, L., Chen, D., Chen, Y.-L., Codella, N., Dai, X., Gao, J., Hu, H., Huang,
  X., Li, B., Li, C., et~al.
\newblock Florence: A new foundation model for computer vision.
\newblock \emph{arXiv preprint arXiv:2111.11432}, 2021.

\bibitem[Yuan et~al.(2022)Yuan, Qian, Cui, Gong, Schroff, Yang, Adam, and
  Liu]{yuan2022contextualized}
Yuan, L., Qian, R., Cui, Y., Gong, B., Schroff, F., Yang, M.-H., Adam, H., and
  Liu, T.
\newblock Contextualized spatio-temporal contrastive learning with
  self-supervision.
\newblock In \emph{CVPR}, 2022.

\bibitem[Yuan et~al.(2023)Yuan, Gundavarapu, Zhao, Zhou, Cui, Jiang, Yang, Jia,
  Weyand, Friedman, et~al.]{yuan2023videoglue}
Yuan, L., Gundavarapu, N.~B., Zhao, L., Zhou, H., Cui, Y., Jiang, L., Yang, X.,
  Jia, M., Weyand, T., Friedman, L., et~al.
\newblock {VideoGLUE}: Video general understanding evaluation of foundation
  models.
\newblock \emph{arXiv preprint arXiv:2307.03166}, 2023.

\bibitem[Zellers et~al.(2021)Zellers, Lu, Hessel, Yu, Park, Cao, Farhadi, and
  Choi]{zellers2021merlot}
Zellers, R., Lu, X., Hessel, J., Yu, Y., Park, J.~S., Cao, J., Farhadi, A., and
  Choi, Y.
\newblock {MERLOT}: Multimodal neural script knowledge models.
\newblock In \emph{NeurIPS}, 2021.

\bibitem[Zellers et~al.(2022)Zellers, Lu, Lu, Yu, Zhao, Salehi, Kusupati,
  Hessel, Farhadi, and Choi]{zellers2022merlot}
Zellers, R., Lu, J., Lu, X., Yu, Y., Zhao, Y., Salehi, M., Kusupati, A.,
  Hessel, J., Farhadi, A., and Choi, Y.
\newblock {MERLOT Reserve}: Neural script knowledge through vision and language
  and sound.
\newblock In \emph{CVPR}, 2022.

\bibitem[Zeng et~al.(2022)Zeng, Attarian, Ichter, Choromanski, Wong, Welker,
  Tombari, Purohit, Ryoo, Sindhwani, et~al.]{zeng2022socratic}
Zeng, A., Attarian, M., Ichter, B., Choromanski, K., Wong, A., Welker, S.,
  Tombari, F., Purohit, A., Ryoo, M., Sindhwani, V., et~al.
\newblock Socratic models: Composing zero-shot multimodal reasoning with
  language.
\newblock \emph{arXiv preprint arXiv:2204.00598}, 2022.

\bibitem[Zhai et~al.(2022{\natexlab{a}})Zhai, Kolesnikov, Houlsby, and
  Beyer]{zhai2022scaling}
Zhai, X., Kolesnikov, A., Houlsby, N., and Beyer, L.
\newblock Scaling vision transformers.
\newblock In \emph{CVPR}, 2022{\natexlab{a}}.

\bibitem[Zhai et~al.(2022{\natexlab{b}})Zhai, Wang, Mustafa, Steiner, Keysers,
  Kolesnikov, and Beyer]{zhai2022lit}
Zhai, X., Wang, X., Mustafa, B., Steiner, A., Keysers, D., Kolesnikov, A., and
  Beyer, L.
\newblock {LiT}: Zero-shot transfer with locked-image text tuning.
\newblock In \emph{CVPR}, 2022{\natexlab{b}}.

\bibitem[Zhang et~al.(2023{\natexlab{a}})Zhang, Li, and Bing]{zhang2023video}
Zhang, H., Li, X., and Bing, L.
\newblock {Video-LLaMA}: An instruction-tuned audio-visual language model for
  video understanding.
\newblock \emph{arXiv preprint arXiv:2306.02858}, 2023{\natexlab{a}}.

\bibitem[Zhang et~al.(2023{\natexlab{b}})Zhang, Chen, Yuan, Chen, Wang, Wang,
  Han, Chen, Pi, Yao, Han, Ding, and Wang]{zhang2023cae}
Zhang, X., Chen, J., Yuan, J., Chen, Q., Wang, J., Wang, X., Han, S., Chen, X.,
  Pi, J., Yao, K., Han, J., Ding, E., and Wang, J.
\newblock {CAE} v2: Context autoencoder with {CLIP} latent alignment.
\newblock \emph{TMLR}, 2023{\natexlab{b}}.
\newblock ISSN 2835-8856.
\newblock URL \url{https://openreview.net/forum?id=f36LaK7M0F}.

\bibitem[Zhao et~al.(2024)Zhao, Zhao, Zhou, Wu, Chu, Miao, Schroff, Adam, Liu,
  Gong, et~al.]{zhao2024distilling}
Zhao, Y., Zhao, L., Zhou, X., Wu, J., Chu, C.-T., Miao, H., Schroff, F., Adam,
  H., Liu, T., Gong, B., et~al.
\newblock Distilling vision-language models on millions of videos.
\newblock In \emph{CVPR}, 2024.

\bibitem[Zhou et~al.(2022)Zhou, Wei, Wang, Shen, Xie, Yuille, and
  Kong]{zhou2022ibot}
Zhou, J., Wei, C., Wang, H., Shen, W., Xie, C., Yuille, A., and Kong, T.
\newblock {iBOT}: Image {BERT} pre-training with online tokenizer.
\newblock In \emph{ICLR}, 2022.

\bibitem[Zhou et~al.(2018)Zhou, Xu, and Corso]{zhou2018towards}
Zhou, L., Xu, C., and Corso, J.
\newblock Towards automatic learning of procedures from web instructional
  videos.
\newblock In \emph{AAAI}, 2018.

\bibitem[Zhu et~al.(2024)Zhu, Lin, Ning, Yan, Cui, Wang, Pang, Jiang, Zhang,
  Li, et~al.]{zhu2023languagebind}
Zhu, B., Lin, B., Ning, M., Yan, Y., Cui, J., Wang, H., Pang, Y., Jiang, W.,
  Zhang, J., Li, Z., et~al.
\newblock {LanguageBind}: Extending video-language pretraining to {N}-modality
  by language-based semantic alignment.
\newblock In \emph{ICLR}, 2024.

\end{thebibliography}
\bibliographystyle{icml2024}

\clearpage
\appendix
\section{Pretraining data}
\label{app:data}

\subsection{Data curation}
\label{app:data:curation}
\cref{sec:app:data} and \cref{tbl:pretraining_data} have described the pretraining corpus of videos, and the following provides more details about the three in-house datasets.

\textbf{Anonymous-Corpus \#1} consists of about 36M commercially licensed stock video-caption pairs, where the videos and text are manually uploaded by professional contributors. Hence, the quality of the videos and captions is high in this corpus compared with the rest. Note that we do not do any filtering on this set.

\textbf{Anonymous-Corpus \#2} contains 170M (video, ASR transcript) pairs from 44.6M YouTube videos. Its construction process is similar to HowTo100M~\cite{miech2019howto100m}, but the whole corpus is larger and more diverse. Furthermore, view counts and video lengths are filtered using simple metadata. ASR sentence boundaries define the clip boundaries. The clip-text pairs are filtered based on a groundedness score similar to CLIP's similarity score~\cite{wu2021godiva}.

\textbf{Anonymous-Corpus \#3} includes 71.5M (clip, machine-generated caption) pairs from 36.7M YouTube videos. The clips are captioned using vision-language models~\cite{chen2023pali} and further summarized using an LLM~\cite{anil2023palm}. The corpus is similar to InternVid~\cite{wang2023internvid} in terms of construction but a magnitude larger in size and diversity. The initial video selection of this dataset process is similar to Anonymous-Corpus \#2, but additional filters are applied to exclude videos composed of primarily talking heads using a face detection model. Also, static clips are eliminated by ensuring semantic feature embeddings from the frames are not static. Hence, the video content is more diverse than Anonymous-Corpus \#2.

\subsection{Corpus analysis}

\begin{figure*}[t]
\begin{center}
\begin{subfigure}{0.33\linewidth}
\includegraphics[width=\linewidth]{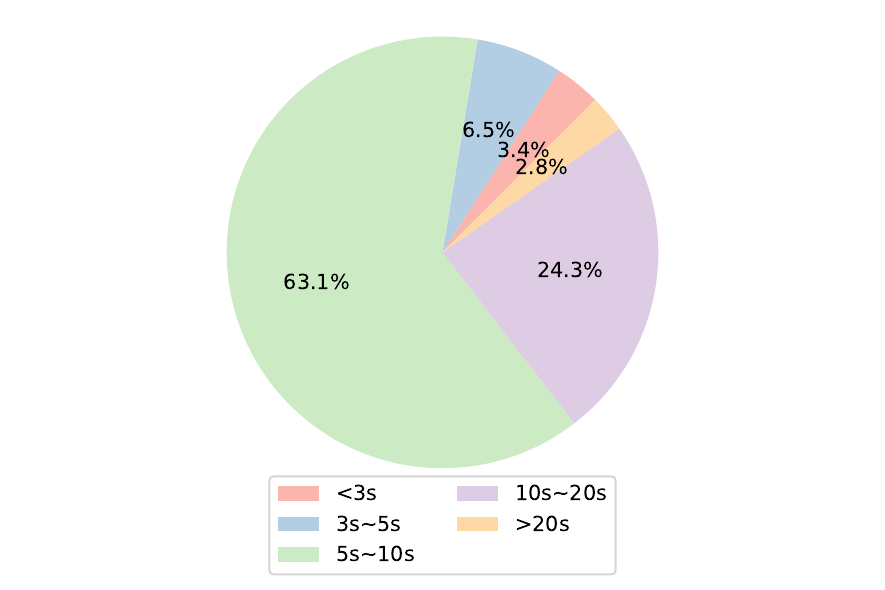}
\caption{Duration (seconds)}
\label{fig:corpus_durations}
\end{subfigure}
\begin{subfigure}{0.33\linewidth}
\includegraphics[width=\linewidth]{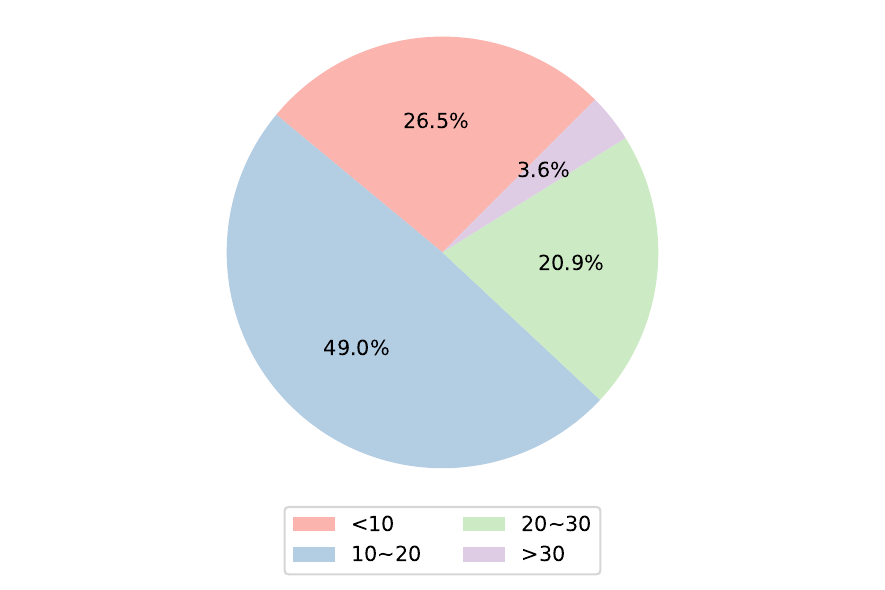}
\caption{Caption length (\# of words)}
\label{fig:corpus_caption_length}
\end{subfigure}
\begin{subfigure}{0.33\linewidth}
\includegraphics[width=\linewidth]{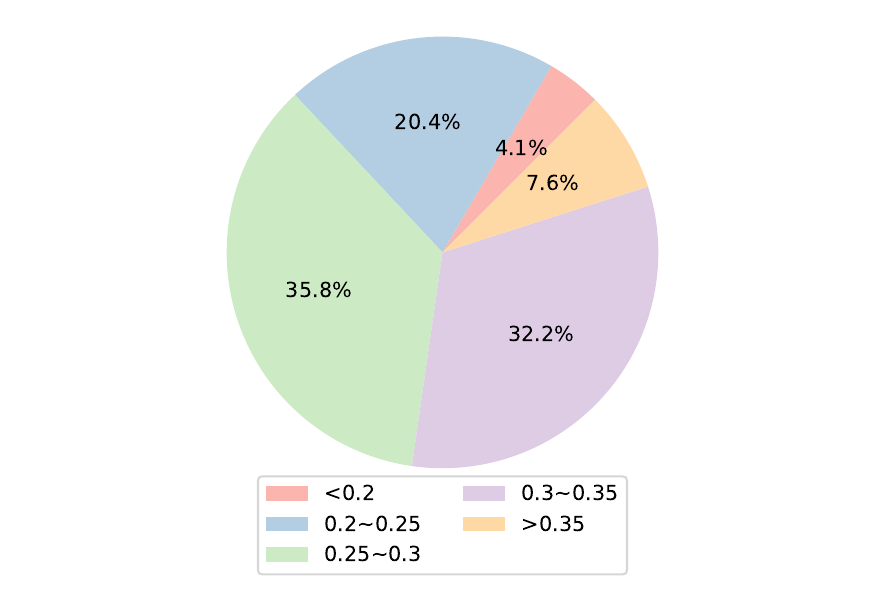}
\caption{CLIP similarity scores}
\label{fig:corpus_clip_scores}
\end{subfigure}
\caption{\textbf{Analysis on the video-text pretraining corpus.}}
\label{fig:corpus_analysis}
\end{center}
\vskip -0.1in
\end{figure*}

We randomly sample 100K videos from our video-text pretraining data and show the breakdown analysis in \cref{fig:corpus_analysis}.
We notice that most of our clips are between 5 to 10 seconds in length and contain 10 to 20 words in the parallel text.
In addition, a considerable proportion of clips has duration longer than 10 seconds or captions longer than 20 words.
We further show the the CLIP similarity score~\cite{wu2021godiva} of our corpus in \cref{fig:corpus_clip_scores}.
The large variations of the CLIP similarity scores demonstrate the diverse caption quality of our training data, which we believe is a byproduct of the various ways used to harvest the text.

Furthermore, we provide an in-depth analyses on each Anonymous dataset and contrast them with the other datasets we used in \cref{tbl:corpus}. We find that the datasets with captions generated by VLMs and LLMs (\eg, InternVid and Anonymous-Corpus \#3) perform the best. In addition, retrieval performance is correlated with dataset size, groundingness (CLIP score), ``dynamic degree'' (optical flow), and the presence of humans. Finally, we compute the top-50 object categories represented by our dataset obtained by running an open-source Tensorflow object detection API on the center-frame of 100K clips from our pretraining data. We find \emph{person} to be the top category, followed by \emph{car}, \emph{chair}, \emph{TV}, \emph{bottle}, \emph{book}, \emph{potted plant}, and \emph{bowl}. We hope this gives more insights into the composition of our datasets.

\begin{table*}
    \caption{
    \textbf{More details about different characteristics of each dataset.}
    The noun/verb ratio and the presence of humans in the video are inferred from the captions using WordNet synsets~\cite{fellbaum2005wordnet} and the NTLK library. Optical flow is computed at $8$ FPS and $256 \times 256$ resolution, and the average optical flow across videos is presented. The last column, in addition to the data analyses, is the average of zero-shot R@1 text-video retrieval results on MSRVTT (full-split) and VATEX.
    }
    \label{tbl:corpus}
    \begin{center}
    \begin{scriptsize}
\begin{tabular}{l|ccccccc}
    \toprule
    \multirow{2}{*}{Datasets} & \multirow{2}{*}{Caption source} & \multirow{2}{*}{\# of clips} & Noun / Verb & \% of videos & Optical flow & Grounding & Average \\
    & & & ratio & with humans & magnitude (px) & score & ZS R@1 \\
    \midrule
    InternVid~\cite{wang2023internvid} & Generated & 7.0M & 3.4 & 79.9 & 2.47 & 0.32 & 45.7 \\
    YT-Temporal-180M~\cite{zellers2021merlot} & ASR & 87.8M & 1.6 & 81.8 & 1.73 & 0.24 & 35.8 \\
    VideoCC~\cite{nagrani2022learning} & Retrievals & 133.5M & 8.0 & 69.0 & 3.21 & 0.27 & 32.6 \\
    WTS-70M~\cite{stroud2020learning} & Metadata & 55.1M & 1.3 & 18.5 & 3.67 & 0.26 & 19.4 \\
    \midrule
    Anonymous-Corpus \#1 & Manual & 36.1M & 5.9 & 76.9 & 1.55 & 0.30 & 37.0 \\
    Anonymous-Corpus \#2 & ASR & 170.3M & 2.8 & 62.2 & 2.15 & 0.26 & 38.6 \\
    Anonymous-Corpus \#3 & Generated & 71.5M & 3.2 & 92.0 & 3.17 & 0.30 & 49.4 \\
    \bottomrule
    \end{tabular}
\end{scriptsize}
    \end{center}
    \vskip -0.1in
\end{table*}
 \section{Model architecture}
\label{app:architecture}
\cref{tab:arch_encoder} shows the \OURS model architecture.
As mentioned in \cref{sec:app:arch}, the architecture follows the factorized design of ViViT~\cite{arnab2021vivit}. It consists of two separate Transformer modules: a spatial module and a temporal module. After an input video is partitioned into several non-overlapping patches (\ie, tokens), the spatial module first models interactions between tokens from the same temporal index. Then the output sequence of token embeddings are forwarded through the temporal module to model interactions between tokens from different temporal indices. The temporal module shares the same setup of the spatial counterpart, except that its number of layers is fixed to four because no performance improvements are observed with more layers added to our largest \OURS model. The positional embeddings of our models are learnable~\cite{devlin2019bert} and decoupled in spatial and temporal dimensions. They are utilized to encode the position information of the input tokens in space and time, respectively. When we add image-text data to the first-stage pretraining, the images are treated as one-frame videos, and we crop the temporal positional embeddings when handling the image input. Following CoCa~\cite{yu2022coca}, we pretrain the model with spatial resolution of $288 \times 288$ and patch size $18 \times 18$. We uniformly sample $8$ frames from each video for pretraining and $16$ frames for evaluation by interpolating the temporal positional embedding of our video encoder. 

\begin{table}[t]
    \caption{\textbf{Encoder architecture of \OURS-g}. When describing the output shape, we use \{temporal, spatial, and channel\} as the order of dimensions when applicable, and we omit the batch size for simplicity. We highlight the dimension that a step applies to by \underline{underline}. Note that the drop token or masking ratio $\rho$ is set to $0.5$ in Stage 1 and $0.65$ in Stage 2.}
	\label{tab:arch_encoder}
\begin{center}
\begin{scriptsize}
    \begin{tabular}	{@{}l|ll@{}}
        \toprule
        Step & Block & Output shape \\
        \midrule
        Data & - & $8 \times 288 \times 288 \times 3$ \\
        Preprocess & Patchify $[1, 18, 18]$ & $8 \times \underline{256} \times 1408 $\\
        Drop token / Mask & Tube / BEVT & $\underline{[8 \times (1-\rho)] \times 256} \times 1408$ \\
        Spatial encoder & MSA ($6144$) $\times 40$ & $[8 \times (1-\rho)] \times \underline{256} \times 1408 $ \\
Normalization & LayerNorm & $[8 \times (1-\rho)] \times \underline{256} \times 1408$ \\
        Transpose & Switch dimension & $\underline{256} \times \underline{[8\times(1-\rho)]} \times 1408$ \\
        Temporal encoder & MSA ($6144$) $\times 4$ & $256 \times \underline{[8\times(1-\rho)]} \times 1408$ \\
        Normalization & Layer Norm &  $256 \times \underline{[8\times(1-\rho)]} \times 1408$ \\
        Transpose & Switch dimension  &  $\underline{[8\times(1-\rho)]} \times \underline{256} \times 1408$ \\
        Reshape & Merge dimension & $\underline{[2048\times(1-\rho)]} \times 1408$ \\
    	\bottomrule
		\end{tabular}
\end{scriptsize}
	\end{center}
	\vskip -0.1in
\end{table}

\section{Implementation details}
\label{app:implement}

In this section, we describe the implementation details and training setups of \OURS. We summarize the pretraining configurations in \cref{tab:pretrain_config}.

\begin{table}[t]
    \caption{\textbf{Summary of our pretraining configurations.}}
	\label{tab:pretrain_config}
	\vskip 0.1in
	\begin{center}
\begin{scriptsize}
    \begin{tabular}	{@{}l|ll@{}}
        \toprule
        Configuration & Stage 1 & Stage 2 \\
        \midrule
Optimizer & AdaFactor & AdaFactor \\
        Base learning rate & $5\times 10^{-4}$ & $5\times 10^{-4}$\\
        Learning rate schedule & linear decay & cosine decay \\
        Warmup iterations & $2\times 10^4$ & $2.5 \times10^{4} $ \\
        Training iterations & $2\times 10^5$ & $3\times 10^5$ \\
        Weight decay & $1\times 10^{-4}$ & $0.05$ \\
        Batch size & $4096$ & $4096$ \\
        Drop token or Mask & $0.5$ (Tube mask) & $0.65$ (BEVT mask)\\
        \bottomrule
		\end{tabular}
		\end{scriptsize}
\end{center}
	\vskip -0.1in
\end{table}

\subsection{Stage 1}
\label{sec:appendix:impl:stage1}

\paragraph{Model design.} The text encoder of the first-stage model is a standard Transformer~\cite{vaswani2017attention}. Together with the spatial module in our encoder, it is initialized from the unimodal text decoder of CoCa~\cite{yu2022coca}. We attach a MAP layer~\cite{lee2019set,yu2022coca} to the end of the video encoder to extract the global embedding from the encoder output. For the text encoder, we append a learnable class token at the end of the input sentence and use its corresponding output as the text embedding. 

\paragraph{Training.} In contrast to existing methods that use batch mixing, we adopt alternating gradient descent (AGD)~\cite{jain2017non} to contrastively train our first-stage model with multiple datasets. It alternates samples from different datasets as mini-batches during training, shown effective in a multi-task and multi-dataset scenario~\cite{akbari2023alternating}. This is particularly useful for our model to avoid easy negatives within a batch, since samples from the same dataset usually follow the same distribution and are harder to distinguish. Furthermore, we observe that the AGD approach scales well as we add more datasets or increase the size of the corpus.

The training of the first-stage model follows the conventional setup of vision-language contrastive learning~\cite{radford2021learning}. To reduce the memory cost during pretraining, we drop $50\%$ of video tokens as in~\citet{li2023scaling} and the tube masking strategy~\cite{feichtenhofer2022masked} is employed for dropping tokens. The teacher model is optimized using Adafactor~\cite{shazeer2018adafactor} with the batch size of $4096$. We set the learning rate to $1 \times 10^{-4}$ for our base model and $5 \times 10^{-5}$ for the giant model. We train the first-stage model for $2 \times 10^5$ steps with $2 \times 10^4$ warm up steps and linear learning rate decay. A symmetric cross-entropy loss \cite{gutmann2010noise,radford2021learning,jia2021scaling,cheng2023vindlu} is used in the first-stage training.

\subsection{Stage 2}

\begin{algorithm}[t]
\caption{Token shuffling pseudo-implementation.}
\newcommand{\hlbox}[1]{\fboxsep=1.2pt\hspace*{-\fboxsep}\colorbox{black!10}{\detokenize{#1}}}
\lstset{style=simple}
\begin{lstlisting}[language=Python]
# token_emb    : visible token embedding [b, m, dim]
# pos_emb      : positional embedding [n, dim]
# mask_emb     : mask embedding [dim]
# b            : mini-batch size
# m            : sequence length of visible tokens
# n            : full sequence length of input video

z = expand_dims(mask_emb, axis=[0, 1]) # [1, 1, dim]
z = tile(z, reps=[b, n - m, 1]) # [b, n - m, dim]
out_emb = concat([token_emb, z], axis=1)
out_emb = shuffle(out_emb, axis=1)

x = expand_dims(pos_emb, axis=0) # [1, n, dim]
out_emb = out_emb + x # feed out_emb to decoder
\end{lstlisting}
\label{alg:shuffle}
\end{algorithm}

\paragraph{Token-wise distillation.} As discussed in \cref{sec:app:train:student}, after training the first-stage model with contrastive learning, we train the \OURS video encoder with masked modeling to reconstruct the spatiotemporal embeddings from the first-stage model.
As shown in \cref{fig:pipeline}, the training pipeline of the second stage is similar to MVD~\cite{wang2023masked}.
After patchifying the input video sequence to a set of tokens, we apply BEVT masking~\cite{wang2022bevt} with a masking ratio of $0.65$ to randomly remove some of the tokens.
The second-stage video encoder, which is initialized from the first-stage encoder, takes the remaining visible tokens as input and predict their embeddings.
A learnable \textsc{mask} token is then used to fill in the position of the masked tokens to form a full sequence together with these visible embeddings.
The full sequence of embeddings is then randomly shuffled and added with positional embedding before being fed into a shallow decoder which is a four-layer Transformer.
A linear layer is then used to align the output of the  decoder with the embeddings of the first-stage video encoder by minimizing their cosine distance.
\cref{alg:shuffle} presents a pseudocode implementation of the proposed token shuffling for masked video modeling.

\begin{table}[t]
    \caption{\textbf{Decoder architectures of \OURS-g}. We highlight the dimension that a step applies to by \underline{underline}. Note that the masking ratio $\rho$ is set to $0.65$.}
	\label{tab:arch_decoder}
\begin{center}
\begin{scriptsize}
    \begin{tabular}	{@{}l|lll@{}}
        \toprule
        \multirow{2}{*}{Step} & \multirow{2}{*}{Block} & \multicolumn{2}{c}{Decoder output shape} \\
        & & Local & Global \\
        \midrule
        Data & - & $ 2048 \times 1408 $ & $[2048\times(1-\rho)] \times 1408$ \\
        Projector & MLP & $ \underline{2048} \times 512 $ & $\underline{[2048\times(1-\rho)]} \times 512 $ \\
        Decoder & MSA ($2048$) $\times 4$ & $\underline{2048} \times 512 $ & $\underline{[2048\times(1-\rho)]} \times 512 $  \\
        Projector & MLP & $\underline{2048} \times 1408$ & $\underline{[2048\times(1-\rho)]} \times 1408$ \\
    	\bottomrule
		\end{tabular}
	\end{scriptsize}
\end{center}
	\vskip -0.1in
\end{table}

\paragraph{Global distillation.} To distill the global visual embedding from the first-stage model, we employ a four-layer Transformer decoder followed by a MAP layer to take the visible embeddings from the second-stage video encoder as input and output a global embedding.
We do not apply token shuffling or add positional embedding for this decoder.
We then align this second-stage global embedding to the global visual embedding from the first-stage model using a cosine distance loss.
Please note that the global visual embedding from the first-stage model is predicted by the same MAP head of contrastive training in the first stage.
\cref{tab:arch_decoder} shows the decoder architectures in this stage.

\paragraph{Training.} We train the second-stage video encoder using the same video clips for the first-stage  model, excluding WebLI~\cite{chen2023pali}, the image-based dataset.
We use Adafactor~\cite{shazeer2018adafactor} for optimization.
The second-stage video encoder is trained with batch size $4096$ and a starting learning rate of $5 \times 10^{-4}$.
The learning rate is decayed to $1 \times 10^{-5}$ with a cosine annealing schedule.
$2.5 \times 10^4$ warm up steps are also used to linearly increase the learning rate from $0$ to $5 \times 10^{-4}$ at the beginning.
The second-stage video encoder is trained for $3 \times 10^5$ steps.
We apply the same weight for token-wise distillation loss and global distillation loss in the second-stage training. \section{Evaluation data}
\label{app:evaluation_data}

\cref{tab:eval_datasets} summarizes all the datasets and their corresponding metrics utilized for evaluation in this paper.
The evaluation datasets are categorized into four parts: general video-only understanding (VideoGLUE~\cite{yuan2023videoglue}), video-text retrieval, captioning \& QA, and CV for science.
Within each category, we select representative datasets and report the standard metric on each of them.

We compare the performance of \OURS to the previous best-performing foundation models in \cref{fig:flvid_all_stats}.
For each dataset and task, we compute the performance gain ($\Delta$Score) with respect to the best reported number achieved by an image or video foundation model.
We collect all of them and plot in descending order.

\begin{table*}
    \caption{
    \textbf{Summary of evaluation datasets.}
    We report the corresponding standard metric for each dataset, including Top-1/5 Accuracy (Acc.) for classification and question answering, mean Average Precision (mAP) for multi-label classification, Recall@1/5 for retrieval, multi-choice retrieval accuracy (MC Acc.) for multi-choice retrieval, CIDEr score for captioning, Wu-Palmer Similarity (WUPS) index for question answering, and macro-accuracy (Macro Acc.) for the KABR dataset.
    }
    \label{tab:eval_datasets}
    \begin{center}
    \begin{scriptsize}
\begin{tabular}{l|l|c|c|l}
    \toprule
    Datasets & Tasks & Zero-shot & Abbr. & Metrics \\
    \midrule
    \tabrow Kinetics-400~\cite{kay2017kinetics} & Video Classification & \xmark & VC & Top-1 Acc. \\
    MiT~\cite{monfort2019moments} & Video Classification & \xmark  & VC & Top-1 Acc. \\
    \tabrow SSv2~\cite{goyal2017something} & Video Classification & \xmark & VC & Top-1 Acc. \\
    Diving48~\cite{li2018resound} & Video Classification & \xmark & VC & Top-1 Acc. \\
    \tabrow Charades~\cite{sigurdsson2016hollywood} & Video Classification & \xmark & VC & mAP \\
    ActivityNet~\cite{caba2015activitynet} & Temporal Action Localization & \xmark & TAL & mAP \\
    \tabrow AVA~\cite{gu2018ava} & Spatiotemporal Action Localization & \xmark & STAL & mAP \\
    AVA-Kinetics~\cite{li2020ava} & Spatiotemporal Action Localization & \xmark & STAL & mAP \\
    \midrule
    \tabrow MSRVTT~\cite{xu2016msr} & Text-to-Video Retrieval & \cmark & ZST2V & Recall@1, Recall@5 \\
    MSRVTT~\cite{xu2016msr} & Video-to-Text Retrieval & \cmark & ZSV2T & Recall@1, Recall@5 \\
    \tabrow VATEX~\cite{wang2019vatex} & Text-to-Video Retrieval & \cmark & ZST2V & Recall@1, Recall@5 \\
    VATEX~\cite{wang2019vatex} & Video-to-Text Retrieval & \cmark & ZSV2T & Recall@1, Recall@5 \\
    \tabrow ActivityNet~\cite{caba2015activitynet} & Text-to-Video Retrieval & \cmark & ZST2V & Recall@1, Recall@5 \\
    ActivityNet~\cite{caba2015activitynet} & Video-to-Text Retrieval & \cmark & ZSV2T & Recall@1, Recall@5 \\
    \tabrow Kinetics-400~\cite{kay2017kinetics} & Video Classification & \cmark & ZSC & Top-1 \& Top-5 Acc. \\
    Kinetics-600~\cite{carreira2018short} & Video Classification & \cmark & ZSC & Top-1 \& Top-5 Acc. \\
    \tabrow SSv2-Temporal~\cite{sevilla2021only} & Video Classification & \cmark & ZSC & Top-1 Acc. \\
    SSv2-Events~\cite{bagad2023test} & Video Classification & \cmark & ZSC & Top-1 Acc. \\
    \tabrow NExT-QA (ATP-Hard)~\cite{xiao2021next} & Video Classification &  \cmark & ZSC & MC Acc. \\
    Charades~\cite{sigurdsson2016hollywood} & Video Classification & \cmark & ZSC & mAP \\
    \tabrow Charades-STA~\cite{gao2017tall} & Video Classification & \cmark & ZSC & MC Acc. \\
    \midrule
    MSRVTT~\cite{xu2016msr} & Video Captioning & \cmark & ZSCap & CIDEr \\
    \tabrow VATEX~\cite{wang2019vatex} & Video Captioning & \cmark & ZSCap & CIDEr \\
    YouCook2~\cite{zhou2018towards} & Video Captioning & \cmark & ZSCap & CIDEr \\
    \tabrow MSRVTT-QA~\cite{xu2017video} & Video Question Answering & \cmark & ZSQA & Top-1 Acc. \\
    MSVD-QA~\cite{xu2017video} & Video Question Answering & \cmark & ZSQA & Top-1 Acc. \\
    \tabrow NExT-QA~\cite{xiao2021next} & Video Question Answering & \cmark & ZSQA & WUPS \\
    \midrule
    Fly vs. Fly~\cite{eyjolfsdottir2014detecting} & Video Classification & \xmark & VC & mAP \\
    \tabrow CalMS21~\cite{sun2021multi} & Video Classification & \xmark & VC & mAP \\
    CRIM13 (Side view)~\cite{burgos2012social} & Video Classification & \xmark & VC & mAP \\
    \tabrow CRIM13 (Top view)~\cite{burgos2012social} & Video Classification & \xmark & VC & mAP \\
    KABR~\cite{kholiavchenko2024kabr} & Video Classification & \xmark & VC & Macro Acc. \\
    \tabrow ChimpACT~\cite{ma2023chimpact} & Spatiotemporal Action Localization & \xmark & STAL & mAP \\
    \bottomrule
    \end{tabular}
\end{scriptsize}
    \end{center}
    \vskip -0.1in
\end{table*} \begin{table*}
\caption{
\textbf{Results of FM adaptation using frozen features on video understanding tasks.}
The model backbones are frozen and only weights in the task heads are updated using the downstream tasks' training sets.
$^{\ast}$ indicates the model is evaluated under the setting with trainable FLOPs alignment.
}
\label{tbl:videoglue_fz}
\begin{center}
\begin{scriptsize}
\begin{tabular}{@{}l|cc|cc|c|c|cc|c@{}} 
\toprule
\multirow{2}{*}{Methods} & \multicolumn{2}{c|}{VC (A)} & \multicolumn{2}{c|}{VC (M)} & VC (ML) & TAL & \multicolumn{2}{c|}{STAL} & Trainable \\
& \textbf{K400} & \textbf{MiT} & \textbf{SSv2} & \textbf{D48} & \textbf{Charades} & \textbf{ActivityNet} & \textbf{AVA} & \textbf{AVA-K} & FLOPs (B) \\
\midrule
CLIP-B~\cite{radford2021learning} & 75.2 & 32.6 & 41.0 & 44.1 & 11.2 & 32.7 & 21.1 & 25.9 & 3.72 \\
VATT-B~\cite{akbari2021vatt} & 75.1 & 32.1 & 57.8 & 49.7 & 33.3 & 35.3 & 20.3 & 22.2 & 3.72 \\
CoCa-B~\cite{yu2022coca} & 73.1 & 32.0 & 41.5 & 34.1 & 8.8 & 33.0 & 23.3 & 24.7 & 3.72 \\
FLAVA-B~\cite{singh2022flava} & 71.3 & 29.7 & 40.6 & 45.9 & 12.6 & 32.2 & 18.8 & 21.5 & 3.72 \\
VideoMAE-B~\cite{tong2022videomae} & 65.1 & 23.0 & 53.9 & 59.5 & 11.3 & 33.0 & 16.0 & 19.9 & 3.72 \\
InternVideo-B~\cite{wang2022internvideo} & 69.3 & 26.3 & 58.2 & 55.6 & 13.0 & 33.3 & 13.4 & 15.7 & 3.72 \\
UMT-B~\cite{li2023unmasked} & 77.1 & 34.0 & 47.7 & 47.8 & 30.1 & 35.8 & 20.7 & 21.1 & 3.72 \\
\textbf{\OURS-B}$^{\ast}$ & 82.8 & 40.0 & 61.8 & 59.5 & 38.7 & 36.6 & 29.9 & 32.0 & 3.72 \\
\textbf{\OURS-B} & 84.2 & 40.8 & 63.6 & 67.4 & 40.4 & 36.6 & 30.6 & 31.8 & 9.71 \\
\bottomrule
\end{tabular}
\end{scriptsize}
\end{center}
\vskip -0.1in
\end{table*} \begin{table*}
\caption{
\textbf{Results of FM adaptation using frozen backbones with MLAP heads on video understanding tasks.}
MLAP takes multiple frozen features from an FM as inputs and map them hierarchically
for the final task prediction. Only the MLAP layer weights are updated using the downstream tasks’ training sets.
$^{\ast}$ indicates the model is evaluated under the setting with trainable FLOPs alignment.
}
\label{tbl:videoglue_mlap}
\begin{center}
\begin{scriptsize}
\begin{tabular}{@{}l|cc|cc|c|c|cc|c@{}} 
\toprule
\multirow{2}{*}{Methods} & \multicolumn{2}{c|}{VC (A)} & \multicolumn{2}{c|}{VC (M)} & VC (ML) & TAL & \multicolumn{2}{c|}{STAL} & Trainable \\
& \textbf{K400} & \textbf{MiT} & \textbf{SSv2} & \textbf{D48} & \textbf{Charades} & \textbf{ActivityNet} & \textbf{AVA} & \textbf{AVA-K} & FLOPs (B) \\
\midrule
CLIP-B~\cite{radford2021learning} & 77.1 & 39.0 & 50.1 & 55.8 & 41.5 & 33.9 & 27.7 & 29.6 & 14.9 \\
VATT-B~\cite{akbari2021vatt} & 75.1 & 35.6 & 58.7 & 60.1 & 58.2 & 35.0 & 22.9 & 24.1 & 14.9 \\
CoCa-B~\cite{yu2022coca} & 74.2 & 37.2 & 45.9 & 48.4 & 19.6 & 33.3 & 24.4 & 27.0 & 14.9 \\
FLAVA-B~\cite{singh2022flava} & 71.5 & 34.5 & 43.1 & 58.5 & 38.2 & 32.4 & 21.3 & 23.2 & 14.9 \\
VideoMAE-B~\cite{tong2022videomae} & 71.7 & 32.2 & 57.4 & 69.6 & 35.9 & 33.4 & 19.6 & 22.1 & 14.9 \\
InternVideo-B~\cite{wang2022internvideo} & 73.7 & 34.7 & 60.3 & 71.9 & 40.5 & 33.6 & 15.9 & 17.7 & 14.9 \\
UMT-B~\cite{li2023unmasked} & 77.5 & 38.0 & 51.2 & 55.5 & 55.8 & 36.0 & 24.6 & 25.8 & 14.9 \\
\textbf{\OURS-B}$^{\ast}$ & 83.7 & 43.9 & 64.6 & 70.7 & 56.6 & 37.2 & 31.5 & 33.1 & 14.9 \\
\textbf{\OURS-B} & 84.5 & 43.8 & 66.3 & 73.6 & 58.6 & 37.2 & 31.4 & 33.0 & 38.8 \\
\bottomrule
\end{tabular}
\end{scriptsize}
\end{center}
\vskip -0.1in
\end{table*} \begin{table*}
\caption{
\textbf{Results of FM adaptation using frozen backbones with low-rank adapters and task heads.}
Only the weights of the low-rank adapters and task heads are updated using downstream tasks' training sets.
$^{\ast}$ indicates the model is evaluated under the setting with trainable FLOPs alignment.
}
\label{tbl:videoglue_adpt}
\begin{center}
\begin{scriptsize}
\begin{tabular}{@{}l|cc|cc|c|c|cc|c@{}} 
\toprule
\multirow{2}{*}{Methods} & \multicolumn{2}{c|}{VC (A)} & \multicolumn{2}{c|}{VC (M)} & VC (ML) & TAL & \multicolumn{2}{c|}{STAL} & Trainable \\
& \textbf{K400} & \textbf{MiT} & \textbf{SSv2} & \textbf{D48} & \textbf{Charades} & \textbf{ActivityNet} & \textbf{AVA} & \textbf{AVA-K} & FLOPs (B) \\
\midrule
CLIP-B~\cite{radford2021learning} & 80.2 & 39.7 & 56.0 & 77.2 & 44.2 & - & 24.5 & 28.0 & 6.44 \\
VATT-B~\cite{akbari2021vatt} & 75.0 & 36.5 & 63.5 & 68.9 & 53.5 & - & 22.3 & 25.8 & 6.44 \\
CoCa-B~\cite{yu2022coca} & 80.9 & 41.4 & 56.1 & 67.1 & 45.8 & - & 26.6 & 28.7 & 6.44 \\
FLAVA-B~\cite{singh2022flava} & 74.7 & 34.1 & 52.1 & 68.4 & 40.8 & - & 17.9 & 23.8 & 6.44 \\
VideoMAE-B~\cite{tong2022videomae} & 73.6 & 30.6 & 61.4 & 76.0 & 43.0 & - & 16.6 & 23.3 & 6.44 \\
InternVideo-B~\cite{wang2022internvideo} & 75.5 & 31.3 & 63.9 & 73.6 & 46.2 & - & 19.2 & 25.5 & 6.44 \\
UMT-B~\cite{li2023unmasked} & 81.5 & 40.4 & 61.8 & 78.5 & 50.0 & - & 27.8 & 29.4 & 6.44 \\
\textbf{\OURS-B}$^{\ast}$ & 84.5 & 44.0 & 66.3 & 83.0 & 57.8 & - & 33.6 & 35.7 & 8.71 \\
\textbf{\OURS-B} & 85.7 & 43.9 & 68.8 & 85.1 & 60.6 & - & 34.1 & 35.8 & 22.8 \\
\bottomrule
\end{tabular}
\end{scriptsize}
\end{center}
\vskip -0.1in
\end{table*} \begin{table*}
\caption{
\textbf{Results of FM adaptation by end-to-end fine-tuning.}
All the model weights are updated using the downstream tasks' training sets.
$^{\ast}$ indicates the model is evaluated under the setting with trainable FLOPs alignment.
}
\vskip 0.1in
\label{tbl:videoglue_e2e}
\begin{center}
\begin{scriptsize}
\begin{tabular}{@{}l|cc|cc|c|c|cc|c@{}} 
\toprule
\multirow{2}{*}{Methods} & \multicolumn{2}{c|}{VC (A)} & \multicolumn{2}{c|}{VC (M)} & VC (ML) & TAL & \multicolumn{2}{c|}{STAL} & Trainable \\
& \textbf{K400} & \textbf{MiT} & \textbf{SSv2} & \textbf{D48} & \textbf{Charades} & \textbf{ActivityNet} & \textbf{AVA} & \textbf{AVA-K} & FLOPs (B) \\
\midrule
CLIP-B~\cite{radford2021learning} & 81.0 & 39.0 & 46.6 & 75.7 & 54.3 & - & 27.1 & 28.9 & 367 \\
VATT-B~\cite{akbari2021vatt} & 77.1 & 34.8 & 65.1 & 77.6 & 55.7 & - & 27.0 & 28.4 & 371 \\
CoCa-B~\cite{yu2022coca} & 82.6 & 43.6 & 66.8 & 79.6 & 55.0 & - & 27.7 & 31.0 & 367 \\
FLAVA-B~\cite{singh2022flava} & 79.1 & 38.3 & 61.1 & 72.0 & 48.6 & - & 22.0 & 25.6 & 367 \\
VideoMAE-B~\cite{tong2022videomae} & 78.7 & 36.1 & 65.5 & 75.5 & 51.4 & - & 23.5 & 26.2 & 367 \\
InternVideo-B~\cite{wang2022internvideo} & 80.1 & 35.9 & 67.0 & 75.8 & 52.2 & - & 27.2 & 29.8 & 367 \\
UMT-B~\cite{li2023unmasked} & 83.3 & 38.7 & 67.0 & 79.2 & 57.1 & - & 28.8 & 30.9 & 367 \\
\textbf{\OURS-B}$^{\ast}$ & 84.4 & 43.9 & 68.2 & 82.3 & 58.1 & - & 33.3 & 35.3 & 374 \\
\textbf{\OURS-B} & 85.7 & 44.2 & 70.0 & 84.9 & 60.1 & - & 33.4 & 35.9 & 977 \\
\bottomrule
\end{tabular}
\end{scriptsize}
\end{center}
\vskip -0.1in
\end{table*} \begin{table*}
\caption{\textbf{Stratified average scores under four adaptation methods and the final VGS.} $^{\ast}$ indicates the model is evaluated under the setting with trainable FLOPs alignment.}
\vskip 0.1in
\label{tbl:videoglue_score}
\begin{center}
\begin{scriptsize}
\begin{tabular}{@{}l|c|c|c|c|c@{}} 
\toprule
Methods & Frozen & MLAP & Adapter & E2E & VGS \\
\midrule
CLIP-B~\cite{radford2021learning} & 32.8 & 43.3 & 49.3 & 52.8 & 41.5 \\
VATT-B~\cite{akbari2021vatt} & 39.4 & 46.3 & 49.9 & 52.7 & 45.1 \\
CoCa-B~\cite{yu2022coca} &  31.2 & 36.3 & 49.0 & 55.2 & 39.7 \\
FLAVA-B~\cite{singh2022flava} &  31.7 & 39.3 & 44.1 & 49.4 & 38.5 \\
VideoMAE-B~\cite{tong2022videomae} & 32.6 & 40.9 & 45.9 & 51.0 & 39.9 \\
InternVideo-B~\cite{wang2022internvideo} & 33.1 & 42.2 & 47.7 & 52.5 & 41.0 \\
UMT-B~\cite{li2023unmasked} & 38.0 & 45.6 &  52.4 & 55.3 & 45.3 \\
\textbf{\OURS-B}$^\ast$ & 45.6 & 51.5 & 57.8 & 57.9 & 51.3  \\
\bottomrule
\end{tabular}
\end{scriptsize}
\end{center}
\vskip -0.1in
\end{table*} 
\section{VideoGLUE}
\label{app:videoglue}

\subsection{Tasks and task heads for \OURS}
\label{app:videoglue:tasks}

We follow the VideoGLUE~\cite{yuan2023videoglue} setup for video-only evaluations.  
Given a video clip of shape $T\times H\times W\times 3$, \OURS produces a set of visual tokens of shape $T\times\frac{H}{18}\times\frac{W}{18}\times D$, where $T$, $H$, and $W$ are the number of frames, image height, and image width, respectively, and $D$ is the feature length.

In all our video classification tasks, we employ a multi-head attention pooling (MAP) layer as our task head, which consists of Transformer layers with $12$ heads and hidden size $768$. 
A class token is prepended to cross-attend to all visual tokens from \OURS for final classifications.
We use batch size $256$ when training the task heads.
We apply the same data augmentation strategies and training recipes for each individual dataset as described in VideoGLUE and perform multi-view evaluations.

Spatiotemporal action localization requires to localize person instances in an input video and recognize their actions.
In our experiments, instance-level features are first RoIPooled~\cite{ren2015fasterrcnn} from visual tokens by using corresponding instance boxes. 
These features are then used to query all other visual tokens through cross-attention layers.
We use a Transformer layer with $12$ heads and hidden size $768$ as the task heads.
Final query tokens are classified via a linear classifier.
We use the groundtruth instance boxes with their associated action labels for training.
At test time, we use the same pretrained person detector as in~\citet{feichtenhofer2018slowfast} for person detection on AVA. 
On AVA-Kinetics, we use the detector described in~\citet{li2020ava}. 
We train the models with batch size $256$.

For temporal action localization, we only apply \OURS under frozen and multi-layer attention pooler (MLAP) settings, since the long video samples do not allow end-to-end tuning.
In the MLAP setting, we pool features and input them to a G-TAD head~\cite{xu2019gtad}.
We use batch size $32$ and train G-TAD on ActivityNet v1.3 for $10$ epochs.

We employ AdamW~\cite{loshchilov2017adamw} optimizer and cosine learning rate decay in all video-only experiments. 
For more details on the experiment setups, we refer readers to the VideoGLUE paper~\cite{yuan2023videoglue}.

We experiment \OURS under two configurations regarding different input video sizes. In the first configuration (marked by asterisk ``$^{\ast}$'' in \cref{tbl:videoglue_fz,tbl:videoglue_e2e,tbl:videoglue_adpt,tbl:videoglue_mlap}), we use $8$ frames and $252\times252$ image resolution for feature extraction when training video classification task heads, which results in a sequence of $8\times14\times14$ tokens. On AVA and AVA-Kinetics, video clips of shape $8\times288\times288$ (\ie, token length $8\times16\times16$) are used for both training and evaluation. This configuration aligns the trainable FLOPs of \OURS with the other baseline models reported in~\citet{yuan2023videoglue}. In the second configuration, we use video clips of shape $16\times288\times288$ as input for all the experiments. This configuration aligns with the pretraining setup with higher trainable FLOPs and accounts for the results in \cref{sec:exp:videoglue}.

\subsection{Adaptations}
We follow~\citet{yuan2023videoglue} to report model performances under four adaptation settings, namely frozen model backbone with simple MAP heads, with MLAP heads, and with low-rank adapters, and finally end-to-end finetuning. 

Frozen features with simple MAP heads update a one-layer task head only, which pools over the visual tokens from the backbone.
The MLAP head upgrades from the one-layer MAP pooler by taking a stack of visual tokens as input and mapping them hierarchically for final task prediction. 
In our experiments, the attention pooler in MLAP has four cross-attention layers, following~\citet{yuan2023videoglue}.
The low-rank adaptation inserts low-rank adapter modules~\cite{hu2021lora} with trainable weights into the pretrained video encoders, and uses a one-layer MAP task head. 
During training, both the adapter weights and task heads are updated, and the other weights from video backbones are kept frozen.
We set the inner dimension of the adapter layers to be $64$ for all our experiments.
Finally, end-to-end finetuning is done with a one-layer MAP task head while we update weights in both the backbones and task heads.

\subsection{Results}
\label{app:videoglue:benchmark}

In~\cref{tbl:videoglue_fz,tbl:videoglue_mlap,tbl:videoglue_adpt,tbl:videoglue_e2e}, we report the detailed benchmark results using the aforementioned four adaptation settings\footnote{The UMT-B/16-25M checkpoint is obtained from~\url{https://github.com/OpenGVLab/unmasked_teacher/blob/main/multi_modality/MODEL_ZOO.md}.}.
In Table~\ref{tbl:videoglue_score}, the stratified average scores for each FM under four adaptations are reported. We also report the final VideoGLUE score (VGS) for each FM, which weights their absolute performances with the respective adaptation costs.

Notably, from~\cref{tbl:videoglue_fz,tbl:videoglue_mlap,tbl:videoglue_adpt,tbl:videoglue_e2e}, we can see that when evaluated under the same trainable FLOPs, \OURS consistently and significantly outperforms other FMs across different benchmarks and tasks.
Aligning with the pretraining setup further improves the performance.
The strong results under both configurations indicate the efficacy of the learned representations by \OURS.

From the overall benchmark results, we note that \OURS achieves the best across the board, surpassing existing FMs by a large margin. 
\OURS-B performs strongly on both appearance-based video understanding datasets and motion-aware recognition tasks, thanks to our two-stage pre-training design.
More interestingly, \OURS-B improves upon baseline FMs more significantly on the regime of low adaptation costs according to Table~\ref{tbl:videoglue_score}. 
 \begin{table}[t]
\caption{\textbf{Zero-shot video-text retrieval on MSRVTT}. We follow the full split produced by~\citet{xu2016msr} which contains $2,990$ videos for testing.}
\label{tbl:msrvtt_zs}
\begin{center}
\begin{scriptsize}
\begin{tabular}{l|cccc}
    \toprule
    \multirow{3}{*}{Methods} & \multicolumn{4}{c}{\textbf{MSRVTT}} \\
    & \multicolumn{2}{c}{Text $\rightarrow$ Video} & \multicolumn{2}{c}{Video $\rightarrow$ Text} \\
    & R@1 & R@5 & R@1 & R@5 \\
    \midrule
    CLIP-B~\cite{radford2021learning} & 23.3 & 44.2 & 43.3 & 73.3 \\
    SM-B~\cite{zeng2022socratic} & - & - & 46.9 & 73.5 \\
    CoCa-g~\cite{yu2022coca} & 30.0 & 52.4 & 49.9 & 73.4 \\
    VideoCoCa-g~\cite{yan2022videococa} & 34.3 & 57.8 & 64.7 & 85.2 \\
    \midrule
    \multirow{2}{*}{\textbf{\OURS-B}} & 37.0 & 61.5 & 67.7 & 87.5 \\
    & \upcolor{2.7} & \upcolor{3.7} & \upcolor{3.0} & \upcolor{2.3} \\
    \hline
    \multirow{2}{*}{\textbf{\OURS-g}} & \textbf{39.7} & \textbf{63.7} & \textbf{71.0} & \textbf{90.0} \\
    & \upcolor{5.4} & \upcolor{5.9} & \upcolor{6.3} & \upcolor{4.8} \\
    \bottomrule
\end{tabular}
\end{scriptsize}
\end{center}
\vskip -0.1in
\end{table}

\begin{table}[t]
\caption{\textbf{Zero-shot video classification on Kinetics-600.} Models pretrained with extra modalities in addition to vision and language (\eg, audio) are marked in gray.}
\label{tbl:k600_zs}
\vskip 0.1in
\begin{center}
\resizebox{.96\linewidth}{!}{
\begin{tabular}{lcc}
\toprule
Methods & Top-1 Acc & Top-5 Acc \\
\midrule
ER-ZSAR-B~\cite{chen2021elaborative} & 42.1 & 73.1 \\
CLIP-B~\cite{radford2021learning} & 63.5 & 86.8 \\
CoCa-g~\cite{yu2022coca} & 65.1 & 87.1 \\
VideoCoCa-g~\cite{yan2022videococa} & 70.1 & 88.9 \\
X-CLIP-B~\cite{ni2022expanding} & 65.2 & 86.1 \\
X-Florence-B~\cite{ni2022expanding} & 68.8 & 88.4 \\
Text4Vis-L~\cite{wu2023revisiting} & 68.9 & - \\
MAXI-B~\cite{lin2023match} & 71.5 & 92.5 \\
\fadecell{LanguageBind-L~\cite{zhu2023languagebind}} & \fadecell{61.9} & \fadecell{-} \\
\fadecell{IMP-MoE-L~\cite{akbari2023alternating}} & \fadecell{76.8} & \fadecell{-} \\
\textbf{\OURS-B} & 69.7 \downcolor{1.8} & 90.6 \downcolor{1.9} \\
\textbf{\OURS-g} & \textbf{75.6} \upcolor{4.1} & \textbf{93.2} \upcolor{0.7} \\
\bottomrule
\end{tabular}}
\end{center}
\vskip -0.1in
\end{table}

\section{Zero-shot video-text retrieval}\label{app:lit}

\subsection{Implementation details}\label{app:lit:implementation}

In general, LiT~\cite{zhai2022lit} can be viewed as an efficient way to equip any pretrained vision encoder with zero-shot classification and retrieval capabilities. Here, we follow LiT to pair \OURS with a text encoder to assess its zero-shot performance on discriminative video-language tasks: video-text retrieval and video classification as text retrieval. We let the text encoder mirror the corresponding text encoder from the first-stage model and attach a MAP head to \OURS. Both the text encoder and MAP head are initialized from the teacher model pretrained in Stage~1. Note that as LiT, the video encoder is locked (frozen) during training.

We use exactly the same pretraining data and configurations of the first-stage model to tune the model in this stage. To further boost the model performance, we tune our model only with Anonymous-Corpus \#3 in the last training epoch, whose captions are produced following~\citet{zhao2024distilling}. When evaluating the tuned models on zero-shot video classification tasks, we turn the groundtruth class labels into text descriptions with the text prompts introduced in CLIP~\cite{radford2021learning}.

\subsection{Zero-shot classification on Charades-STA}\label{app:lit:charades_sta}

As mentioned in \cref{sec:exp:vt_retrieval}, in order to evaluate the fine-grained temporal reasoning capability of \OURS, we adapt Charades-STA~\cite{gao2017tall} to the zero-shot video classification task. Charades-STA is originally proposed for temporal grounding where multiple sequential descriptions are annotated with their start and end timestamps for a video.
We repurpose Charades-STA for multi-choice video-to-text retrieval by trimming the video into multiple clips using the annotated timestamps. 
The multi-choice video-to-text retrieval then is performed by retrieving the correct description for a video clip from all sequential descriptions of this video.

\subsection{Additional results on MSRVTT}\label{app:lit:msrvtt}

In~\cref{tbl:msrvtt_zs}, we report zero-shot video-text retrieval results on the full split of MSRVTT produced by~\citet{xu2016msr}, which contains $2,990$ videos for testing. We observe that \OURS outperforms previous methods by a large margin. More importantly, our base-scale model is also better than existing larger-scale models (\eg, CoCa-g and VideoCoCa-g). These findings are consistent with the results in~\cref{tbl:zs_retrieval}, which confirms the strong capability of \OURS on zero-shot video-text retrieval tasks. \emph{Note that, in the appendix, all video-text retrieval results on MSRVTT are calculated using this full split, unless otherwise stated.}

\subsection{Additional results on Kinetics-600}

In addition to \cref{tbl:zs_cls} in the main paper, we provide zero-shot video classification results on Kinetics-600 (K600)~\cite{carreira2018short} in this section. As shown in~\cref{tbl:k600_zs}, we can find that \OURS achieves the best results compared with state-of-the-art FMs that are pretrained with vision and language modalities. Although LanguageBind~\cite{zhu2023languagebind} and IMP~\cite{akbari2023alternating} use additional modalities (\eg, audio) during pretrainig, our results are still comparable to them. More importantly, our base-scale model is able to outperform a majority of methods with even larger scales. These observations are consistent with the ones we draw from \cref{tbl:zs_cls} in the main text. \begin{table}[t]
\caption{\textbf{Datasets included in Academic-Corpus.}}
\label{tbl:academic_corpus}
\vskip 0.1in
\begin{center}
\resizebox{.98\linewidth}{!}{
\begin{tabular}{lc}
\toprule
Datasets & \# of clips \\
\midrule
Video Story Telling~\cite{huang2016visual} & 3K \\
TACoS~\cite{regneri2013groundingtacos} & 4K \\
YouDescribe~\cite{pitcher2023you} & 19K\\
Charades~\cite{sigurdsson2016hollywood, gao2017tall} & 20k \\
COIN~\cite{tang2019coin} & 24K \\
VITT~\cite{huang2020multimodal} & 35K \\
VLN ~\cite{voigtlaender2023connecting}  & 37K \\
EPIC-Kitchens-100~\cite{dima2022rescaling} & 67K \\ 
Spoken Moments in Time~\cite{monfort2021spoken} & 481K \\
Ego4D~\cite{grauman2022ego4d, lin2022egoclip} & 3.8M \\
\bottomrule
\end{tabular}
}
\end{center}
\vskip -0.1in
\end{table}

\begin{table*}[t]
\caption{\textbf{More detailed comparison to state-of-the-art methods on zero-shot video question answering.}  We include additional results under the two-shot prompting and closed-vocabulary settings. We report Top-1 accuracy for both MSRVTT-QA and MSVD-QA. Methods that unfreeze their language models during training are marked in gray.}
\vskip 0.1in
\label{tbl:zs_vqa_more}
\begin{center}
\begin{scriptsize}
\begin{tabular}{@{}l|cc|cc@{}}
    \toprule
    Methods & Two-shot prompting & Closed-vocab. & \textbf{MSRVTT-QA} & \textbf{MSVD-QA} \\
    \midrule
    \tabbar{5}{Question-answering-only models} \\
FrozenBiLM-L~\cite{yang2022zero} & \xmark & \cmark & 22.2 & 39.0 \\
    \midrule
    \tabbar{5}{All-in-one models} \\
    \fadecell{BLIP-B~\cite{li2022blip}} & \fadecell{\xmark} & \fadecell{\cmark} & \fadecell{19.2} & \fadecell{35.2} \\
    \fadecell{HiTeA-B~\cite{ye2023hitea}} & \fadecell{\xmark} & \fadecell{\cmark} & \fadecell{21.7} & \fadecell{37.4} \\
    \fadecell{mPLUG-2~\cite{xu2023mplug}} & \fadecell{\xmark} & \fadecell{\cmark} & \fadecell{43.8} & \fadecell{55.3} \\
    Flamingo-3B~\cite{alayrac2022flamingo} & \cmark & \xmark & 11.0 & 27.5 \\
    Flamingo-9B~\cite{alayrac2022flamingo} & \cmark & \xmark & 13.7 & 30.2 \\
    \midrule
    \textbf{\OURS-B} w/ PaLM-2-1B & \cmark & \xmark & 19.5 & 36.7 \\
    \textbf{\OURS-B} w/ PaLM-2-1B & \xmark & \cmark & 23.1 & 43.2 \\
    \textbf{\OURS-B} w/ PaLM-2-1B & \cmark & \cmark & 28.5 & 39.5 \\
    \midrule
    \textbf{\OURS-B} w/ PaLM-2-8B & \cmark & \xmark & 24.8 & 42.7 \\
    \textbf{\OURS-B} w/ PaLM-2-8B & \xmark & \cmark & 23.4 & 43.4 \\
    \textbf{\OURS-B} w/ PaLM-2-8B & \cmark & \cmark & \textbf{32.0} & \textbf{47.1} \\
    \bottomrule
\end{tabular}
\end{scriptsize}
\end{center}
\vskip -0.1in
\end{table*}

\section{Gluing \OURS with PaLM-2}
\label{app:videoulm}
 
In \cref{sec:exp:cap_vqa}, we provided evidence of the strength and generalizability of \OURS by showing that we can easily fuse it with a pretrained LLM decoder in a further training stage for good performance on tasks that are generative in language such as video captioning and video QA. We provide details about model training and our evaluation protocols in what follows.   

\paragraph{Implementation.} We pass the features of our video encoder through a one-layer Perceiver Resampler~\cite{alayrac2022flamingo} that outputs a fixed number of continuous tokens representing the input video. It is always set to be $256$ in our experiments. These tokens are then prepended to the embedded text prompt and fed into a LLM decoder, \ie, PaLM-2~\cite{anil2023palm}. The resampled features are then added with the original query features via skip connection. Note that there are two differences from the original implementation~\cite{alayrac2022flamingo}. First, a separate LayerNorm is used for query and key features as we find it works better than the shared LayerNorm. Second, we do not concatenate the key features with the query features before the cross attention, since the feature dimensions from \OURS is different from the pretrained PaLM-2. Otherwise, the feature dimensions would need to be projected via a linear projection layer before the concatenation, and we find it leads to unstable training. We ablate with different number of Resampler layers (\ie, $1$, $3$, and $6$) and find that the one-layer Resampler works the best in our experiments.

\paragraph{Model training.}
We train this multimodal model on a combination of video-text captioning data from the pretraining stage, an aggregated Academic-Corpus, and VQAv2~\cite{goyal2017making} (an image QA dataset) using a standard autoregressive language modeling loss.  \cref{tbl:academic_corpus} lists all the datasets in Academic-Corpus, which includes Ego4D~\cite{grauman2022ego4d}, EPIC-Kitchens~\cite{dima2022rescaling}, Spoken Moments In Time~\cite{monfort2021spoken}, \etc, totalling 4.4M video clips. 

Both the video encoder and the LLM are kept entirely frozen during training, only the one-layer Resampler is optimized. We train \OURS-B with PaLM-2-1B and PaLM-2-8B separately. We set batch size to be $256$ for PaLM-2-1B and $64$ for PaLM-2-8B and trained for $2 \times 10^5$ steps. We use Adam optimizer~\cite{kingma2014adam} with weight decay $1 \times 10^{-4}$ and the learning rate is set to be peaked at $5 \times 10^{-4}$ with warmup steps $1 \times 10^4$ and then linearly decreased. Beta1 is set to be $0.9$ and Beta2 is set to be $0.999$. We do not set EMA decay, L2 regularizer weight decay, and gradient clipping in the training.  Each frame is center-cropped to $346$ before being randomly cropped to $288$ during the training and center-cropped to $288$ during the evaluation. We set the maximum decoding steps to be $32$ since the datasets in this work have relatively short answers. We use greedy decoding for all our experiments.

\paragraph{Model evaluation.}
We report both open-vocabulary and closed-vocabulary evaluation results for MSRVTT-QA and MSVD-QA in \cref{tbl:zs_vqa_more}. For the open-vocabulary configuration, we adopt the zero-shot approach of Flamingo~\cite{alayrac2022flamingo} and use two-shot text-only prompts from the training set on each downstream dataset. The use of two-shot text-only prompts is to guide the output style of the answers. We use the following process to select the two-shot prompts for each dataset. We first choose the two most common answers from the training set of the dataset, and then for each of them, a question is randomly drawn from ones in the training set with the corresponding answer.\footnote{The final text-only two-shot prompts we employed are \emph{``question: who is talking to his family? answer: man.''} and \emph{``question: what is a woman doing? answer: talk.''} on MSRVTT-QA, and \emph{``question: who is using a wrench on a pipe fitting? answer: man.''} and \emph{``question: who breaks an egg into a bowl? answer: woman.''} on MSVD-QA.} Compared to Flamingo-9B, \OURS-B with PaLM-2-8B shows an absolute $11.1\%$ and $12.5\%$ gain on MSRVTT-QA and MSVD-QA, respectively.

Additionally, for MSRVTT-QA and MSVD-QA, we experiment with the closed-vocabulary evaluation configuration, following~\citet{li2022blip,yang2022zero}. In this case, instead of directly outputting an answer via the language decoder, we score candidate answers using the log-likelihood of the decoder and choose the answer with the top score. The candidate answers are picked by taking the top-$K$ most frequently appearing one-token answers from the training and validation sets of the dataset, where $K$ is optimized over the validation set by ablating over the values $\{100, 250, 500, 1000, 2000\}$.  For both MSRVTT-QA and MSVD-QA, we find $K = 250$ to work best. Any example where the groundtruth answer is not one of the candidate answers is automatically marked as incorrect. This method additionally steers the model towards answers that fit the exact style of the particular dataset and boosts performance further. In the closed-vocabulary evaluation configuration, \OURS-B with PaLM-2-8B outperforms FrozenBiLM-L by an absolute margin of $1.2\%$ and $4.4\%$ on MSRVTT-QA and MSVD-QA, respectively.

Recently, a number of works \cite{maaz2023video,lin2023video,li2023llama} have begun evaluating captioning and VideoQA tasks using an LLM-in-the-loop protocol, where an LLM such as ChatGPT\footnote{\url{https://chat.openai.com}} is used to compare predictions to ground-truth answers along a number of different dimensions (\eg, correctness of information, temporal understanding, consistency). This can help mitigate the issue of metrics like exact-match and BLEU score being overly reliant on superficial token matching. We leave it to future work to compare against these models using these new protocols.

 \section{CV for Science}\label{app:cv4science}

We evaluate the CV for Science datasets using frozen features with the same feature extraction setup (MAP probing) as the VideoGLUE tasks in Section~\ref{app:videoglue:tasks}. 
The datasets are: Fly vs.\ Fly~\cite{eyjolfsdottir2014detecting} for fly video classification, CalMS21~\cite{sun2021multi} for mouse video classification from top view, CRIM13~\cite{burgos2012social} for mouse video classification with top and side views, ChimpACT~\cite{ma2023chimpact} for chimp spatiotemporal action localization, and KABR~\cite{kholiavchenko2024kabr}  for video classification with Kenyan animals.
The domain expert models reported in the main paper are trained on the training split of each dataset, and reported originally in task programming~\cite{sun2021task} for Fly vs.\ Fly, CalMS21 1D ConvNet with extra unlabelled data~\cite{sun2021multi} for CalMS21, KABR X3D~\cite{kholiavchenko2024kabr} for KABR, and ChimpACT SlowFast~\cite{ma2023chimpact} for ChimpACT.
For each dataset, we use the train and test splits defined by existing work, with the same metrics (mAP for all works, except KABR, which uses macro-accuracy averaged across classes). For Fly vs.\ Fly, we use the data split defined in~\citet{sun2021task}, which includes all behaviors with more than $1000$ frames of annotations in the training set. We note that following previous work~\cite{sun2021multi,sun2021task}, in datasets where there are background classes, the metric is only averaged across behaviors-of-interest (not including background classes).

We extract all frames from the video at the original FPS of each dataset. We use $16$ frames as input in Fly vs.\ Fly, CalMS21, and CRIM13, $64$ frames for ChimpACT, and $16$ frames with a stride of $5$ for KABR, following baselines. Note that for ChimpACT, the benchmark uses groundtruth bounding boxes during training and testing, which we follow. 

The training setup and implementation details are similar to the VideoGLUE frozen-backbone setting (MAP probing) in \cref{app:videoglue:tasks}.
We use the AdamW~\cite{loshchilov2017adamw} optimizer and cosine learning rate decay in the CV for science experiments.
For data augmentation, we use the same ones as other video classification datasets in VideoGLUE (\eg, Charades, Diving48, and MiT) for our video classification datasets. For ChimpACT (spatiotemporal action localization), we use the AVA data augmentation. We use a learning rate of $5 \times 10^{-5}$ for video classification and spatiotemporal action localization, except for KABR, where the base-scale model uses $5 \times 10^{-6}$ and large-scale model uses $1 \times 10^{-6}$. Following the baseline, KABR is also trained with the EQL loss~\cite{tan2020equalization} to account for class imbalance. Finally, all models are trained with $0.5$ dropout rate. 

 \section{Ablation studies}\label{app:ablation}

\subsection{Data}
We study how to combine datasets with different caption qualities, quantities, and distributions when training a video-text contrastive model.
In \cref{tab:data_ablation_teacher}, three different combination methods are considered:
(1) simply mixing different datasets (denoted with ``+'' and ``\xmark'' for AGD);
(2) training with one dataset and then continue training with another dataset (denoted with ``$\rightarrow$'' and ``\xmark'' for AGD);
(3) combining different datasets with AGD (denoted with ``+'' and ``\cmark'' for AGD).
We choose two representative datasets, namely InternVid \cite{wang2023internvid} and YTT180M \cite{zellers2021merlot} and report Recall@1 (R@1) for zero-shot video-to-text (ZSV2T) and text-to-video (ZST2V) retrieval for this study.
We notice that simply mixing InternVid and YTT180M results in a large performance drop on VATEX when compared with only using InternVid.
On the other hand, training on one dataset then continue training on the other dataset is highly affected by the order of datasets.
For instance, compared with only using InternVid, the performance of InternVid $\rightarrow$ YTT180M drops by a large margin on both MSRVTT and VATEX, while YTT180M $\rightarrow$ InternVid improves on three out of four metrics.
Hence, this approach is not scalable with the number of datasets.
Alternatively, AGD consistently improves the performance on MSRVTT and ZSV2T of VATEX compared with YTT180M $\rightarrow$ InternVid and InternVid with only a slightly drop in ZST2V of VATEX.
As a result, AGD is chosen for combining different training datasets when we train the video-text contrastive model.

We further report the performance of AGD with all our pretraining corpus in the last row of \cref{tab:data_ablation_teacher}.
We observe a large improvement across all metrics, demonstrating that AGD scales well with the number of datasets.

\begin{table}[t]
    \caption{\textbf{Data ablation for the first-stage model.} We report both zero-shot text-to-video (ZST2V) and video-to-text (ZSV2T) retrieval results of \OURS-B for each configuration under the evaluation metric of Recall@1.}
	\label{tab:data_ablation_teacher}
	\vskip 0.1in
	\begin{center}
\resizebox{0.98\linewidth}{!}{
		\begin{tabular}	{lccccc}
			\toprule
			\multirow{2}{*}{Data} & \multirow{2}{*}{AGD} & \multicolumn{2}{c}{\textbf{MSRVTT}}  & \multicolumn{2}{c}{\textbf{VATEX}} \\
			  &  & ZST2V & ZSV2T  & ZST2V & ZSV2T \\
			\midrule
			
	InternVid & \xmark & 28.9 & 55.8 & \underline{40.8} & 57.3 \\
	YTT180M & \xmark & 19.6 & 47.5 & 26.3 & 49.7  \\
			\midrule
	InternVid + YT180M & \xmark & 29.5 & 56.7 & 29.9 & 41.0\\
	InternVid $\rightarrow$ YT180M & \xmark & 21.7 & 54.7 & 29.8 & 54.8 \\
	YTT180M $\rightarrow$ InternVid & \xmark & 29.3 & 56.4 & 39.9 & 57.5\\
	InternVid + YTT180M & \cmark & \underline{29.8} & \underline{57.4} & 39.4 & \underline{58.2} \\
	\midrule
Full pretraining corpus & \cmark & \textbf{34.3} & \textbf{64.4} & \textbf{52.0} & \textbf{69.7}\\
	\bottomrule
	\end{tabular}
	}
\end{center}
    \vskip -0.1in
\end{table}

To better understand how our models perform when only public datasets are included for pretraining, we conduct experiments under two setups where only (1) InternVid~(7M) and (2) all public datasets~(150M, including InternVid, YT-Temporal-180M, and WTS-70M) are used, respectively. As shown in \cref{tab:public_data}, we find our first-stage model has already achieved overall favorable results, while our second-stage model yields better results, especially on SSv2. As with the case with any foundation model, the pretraining data is one of the factors to improve performance, but it is not the only factor. The proposed two-stage pretraining strategy, token-wise distillation with token shuffling, and global distillation all contribute to the performance improvements.

\begin{table}[t]
    \caption{\textbf{Model performance when only public datasets are used for pretraining.} Results using MAP probing with frozen backbone for \OURS-B are reported. We report Top-1 accuracy on K400 and SSv2.}
	\vskip 0.1in
	\begin{center}
	\begin{scriptsize}
\begin{tabular}	{lccc}
			\toprule
			Methods & Corpus size & \textbf{K400} & \textbf{SSv2} \\
			\midrule
    UMT-B~\cite{li2023unmasked} & 25M & 77.1 & 47.7 \\
    VideoMAE-B~\cite{tong2022videomae} & 1M & 65.1 & 53.9 \\
    \midrule
    \OURS-B (Stage 1) & 7M & 81.0 & 47.6 \\
    \OURS-B (Stage 1) & 150M & 81.7 & 50.0 \\
    \OURS-B (Stage 2) & 7M & 81.5 & 60.3 \\
    \OURS-B (Stage 2) & 150M & \textbf{82.7} & \textbf{60.5} \\
		\bottomrule
		\end{tabular}
\end{scriptsize}
	\end{center}
	\vskip -0.1in
\label{tab:public_data}
\end{table}

\subsection{Model design}

\begin{table}[t]
    \caption{\textbf{Joint \textit{vs.}\ factorized attention.} Results of the first-stage \OURS-B model are reported. We report Top-1 accuracy on K400 and SSv2 using MAP probing with frozen backbone.}
	\label{tab:factorized}
	\vskip 0.1in
	\begin{center}
\begin{scriptsize}
		\begin{tabular}	{lcccc}
			\toprule
			\multirow{2}{*}{Encoder architectures} & \multicolumn{2}{c}{\textbf{MSRVTT}} & \textbf{K400} & \textbf{SSv2} \\
			& ZST2V & ZSV2T & VC & VC \\
			\midrule
    Joint attention & \textbf{34.2} & 64.0 & \textbf{84.5} & 52.6 \\
    Factorized attention & \textbf{34.2} & \textbf{64.6} & 82.9 & \textbf{55.5} \\
		\bottomrule
		\end{tabular}
	\end{scriptsize}
	\end{center}
	\vskip -0.1in
\end{table}

\paragraph{Factorized encoder.} We favored the factorized attention over joint attention in the encoder because it balances the cost (\eg, memory, efficiency) and performance well. The factorized attention is especially appealing given that contrastive learning (our Stage 1) demands a large batch size. We note that the original ViViT paper~\cite{arnab2021vivit} also recommended the factorized-attention architecture over the other variants. In \cref{tab:factorized}, we can see that the two attention schemes lead to similar performance on text-video retrieval tasks, while the joint attention gives rise to slightly better accuracy on K400 and yet worse on SSv2. The overall performances of these two models are similar, reinforcing that the factorized attention is probably a better choice given its smaller memory footprint.

\begin{table}[t]
    \caption{\textbf{Effects of model initialization.} Results of the first-stage \OURS-B model are reported. We report Top-1 accuracy on K400 and SSv2 using MAP probing with frozen backbone.}
	\label{tab:initialization}
	\vskip 0.1in
	\begin{center}
\begin{scriptsize}
		\begin{tabular}	{lcccc}
			\toprule
			\multirow{2}{*}{Configurations} & \multicolumn{2}{c}{\textbf{MSRVTT}} & \textbf{K400} & \textbf{SSv2} \\
			& ZST2V & ZSV2T & VC & VC \\
			\midrule
	CLIP~\cite{radford2021learning} (frozen) & 27.6 & 52.4 & 73.1 & 39.8 \\
	CLIP~\cite{radford2021learning} & 27.4 & 53.3 & 79.6 & \textbf{51.1} \\
    CoCa~\cite{yu2022coca} & \textbf{29.9} & \textbf{57.3} & \textbf{81.7} & 50.0 \\
		\bottomrule
		\end{tabular}
	\end{scriptsize}
	\end{center}
	\vskip -0.1in
\end{table}

\paragraph{Initialization.} To understand how the end results vary with respect to the initialization from different image models, we replace CoCa~\cite{yu2022coca} with CLIP~\cite{radford2021learning} as the spatial encoder in our model. We then experiment with two variants: (1) freezing CLIP while training the temporal layers and (2) unfreezing all weights in training. Note that we use CLIP-B/14 and all the public datasets in our pretraining corpus for model training in this experiment. Results are shown in \cref{tab:initialization}. We can see that the results using CLIP for initialization are still comparable with those with the CoCa initialization, and unfreezing the CLIP weights benefits all evaluation benchmarks in the table.

\begin{figure}[t]
\vskip 0.1in
\begin{center}
\includegraphics[width=1\linewidth]{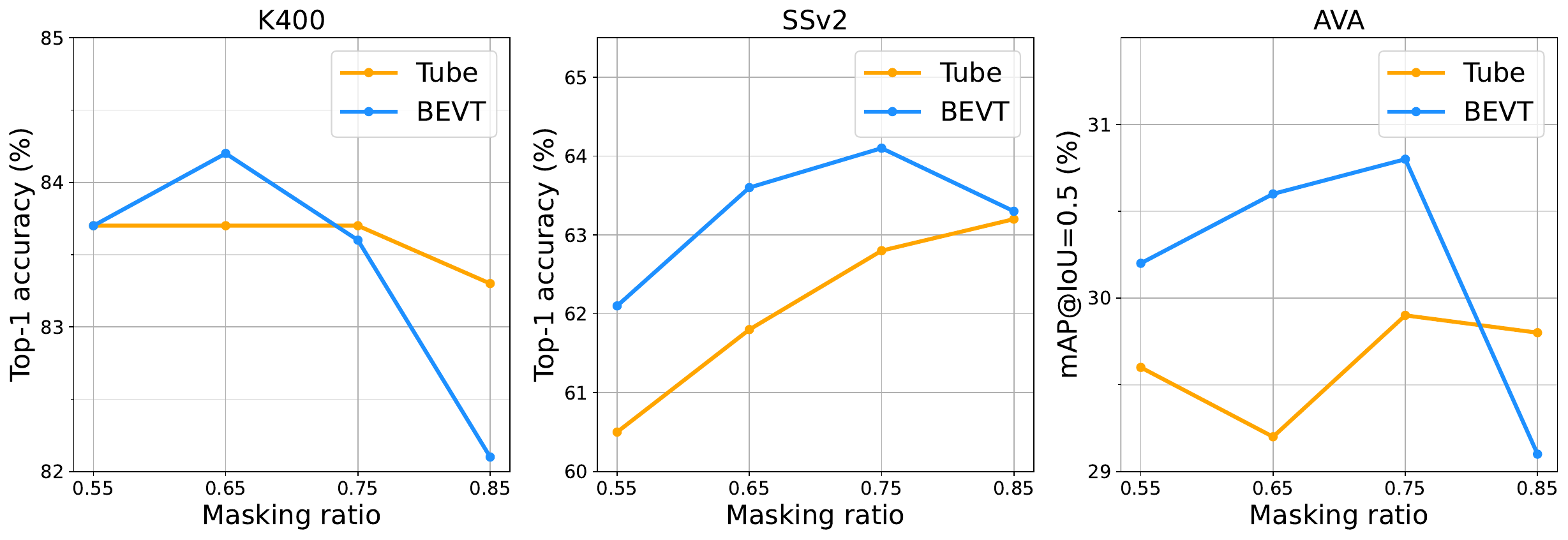} \\
\caption{\textbf{Ablation study for second-stage masking strategy and masking ratio.} Results using MAP probing with frozen backbone for \OURS-B are reported.}
\label{fig:ablation_masking}
\end{center}
\vskip -0.1in
\end{figure}

\paragraph{Masking method.} We then study the impact of masking method and masking ratio on the second-stage model using \OURS-B as an example.
In Figure~\ref{fig:ablation_masking}, we compare the performance of the second-stage model under tube masking~\cite{tong2022videomae} and BEVT masking~\cite{wang2022bevt} with various masking ratios on different video focused tasks.
We notice that BEVT masking outperforms tube masking in most cases. 
When comes to the masking ratio, BEVT masking with masking ratio $0.65$ and $0.75$ have similar performance and outperform the other masking ratios.
As a result, if not otherwise specified, all the second-stage models are trained with the BEVT masking with $0.65$ masking ratio.

\begin{figure}[t]
\vskip 0.1in
\begin{center}
\includegraphics[width=0.86\linewidth]{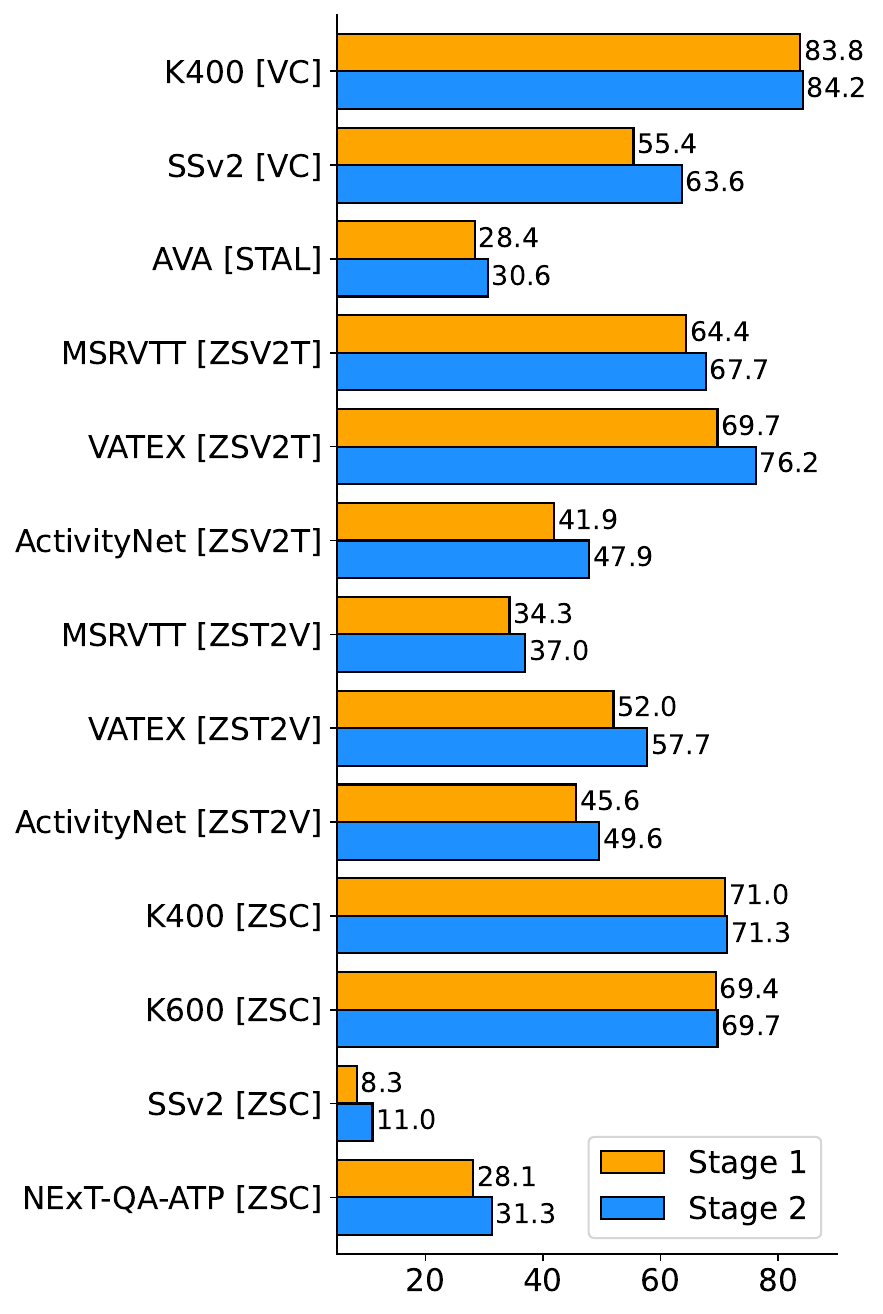}
\caption{\textbf{Comparison between the first-stage and second-stage models of \OURS-B on video understanding tasks.} For video-only tasks, results are from using MAP probing with a frozen backbone.}
\label{fig:ablation_student_teacher}
\end{center}
\vskip -0.1in
\end{figure}

\begin{table}[t]
    \caption{\textbf{Ablation study for the second-stage model training strategy.} Results using MAP probing with frozen backbone for \OURS-B are reported. We report Top-1 accuracy on K400 and SSv2, and mean average precision (mAP) on AVA.}
	\vskip 0.1in
	\begin{center}
	\begin{scriptsize}
\begin{tabular}	{lccc}
			\toprule
			Models & \textbf{K400} & \textbf{SSv2} & \textbf{AVA} \\
			\midrule
    Full configuration & \textbf{84.2} & 63.6 & \textbf{30.6} \\
    w/o token shuffling & 83.6 \downcolor{0.6} & 61.8  \downcolor{1.8} & 29.4 \downcolor{1.2} \\
    w/o global distillation & 83.4 \downcolor{0.8} & \textbf{64.2} \upcolor{0.6} & 29.0 \downcolor{1.6} \\
\bottomrule
		\end{tabular}
\end{scriptsize}
	\end{center}
	\vskip -0.1in
\label{tab:stage2_ablation}
\end{table}

\begin{figure}[t]
\vskip 0.1in
\begin{center}
\begin{subfigure}{\linewidth}
\centerline{\includegraphics[width=\linewidth]{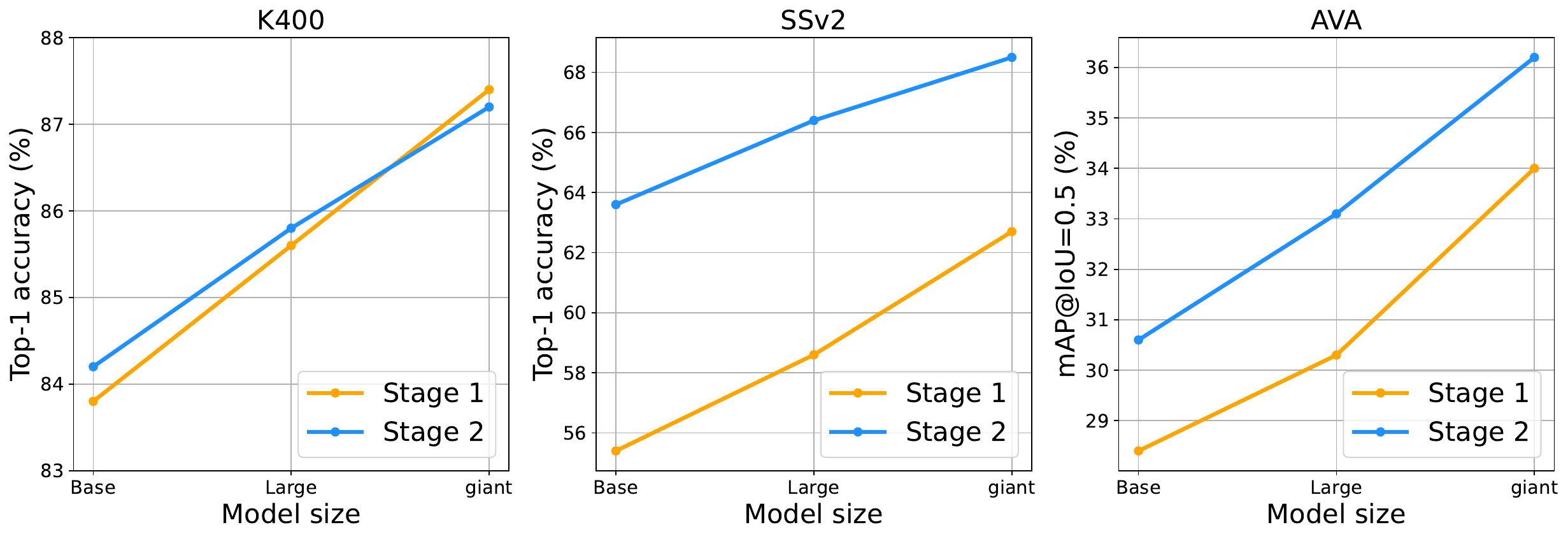}}
\caption{Scaling student and teacher model size.}
\label{fig:model_scaling_variable_teacher}
\end{subfigure}
\vskip 0.1in
\begin{subfigure}{\linewidth}
\centerline{\includegraphics[width=\linewidth]{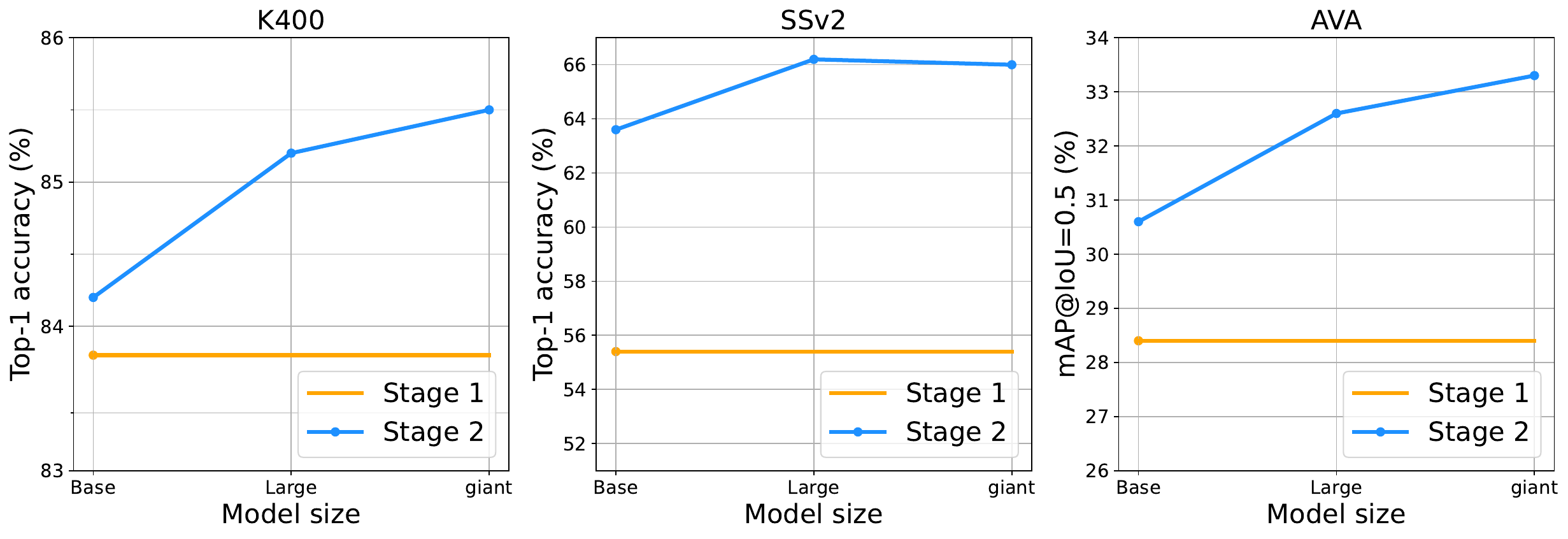}}
\caption{Scaling student model size with a fixed \emph{Base} teacher model.}
\label{fig:model_scaling_fixed_teacher}
\end{subfigure}
\caption{\textbf{Preliminary studies on model scaling.} Results using MAP probing with frozen backbone are reported. We report Top-1 accuracy on K400 and SSv2, and mean average precision (mAP) on AVA.}
\label{fig:model_scaling}
\end{center}
\vskip -0.1in
\end{figure}

\paragraph{Token shuffling and global distillation.} We then study the performance of token shuffling and global distillation which are the two new techniques introduced in our masking distillation method.
We show the results of the second-stage model (\OURS-B) without token shuffling or global distillation and compare the results with the full second-stage model on video classification (K400 and SSv2) and spatiotemporal action location (AVA) tasks in Table~\ref{tab:stage2_ablation}.
From this table, we notice that both token shuffling and global distillation help improving the performance of the second-stage model by a large margin.
Especially, token shuffling improves the performance of the second-stage model on motion-focused video classification dataset SSv2 by $1.8\%$.
We believe that token shuffling introduces a harder learning objective to the second-stage model that is akin to Jigsaw puzzle~\cite{noroozi2016unsupervised}, forcing the second-stage model to better understand the motion in the video.
Global distillation, on the other hand, can further boost the performance of the second-stage model on appearance-based video tasks ($0.8\%$ on K400 and $1.6\%$ on AVA) and plays an important role in training a good second-stage model.

\paragraph{The second-stage training.} Figure~\ref{fig:ablation_student_teacher} compares the second-stage model with the first-stage model on different video datasets across a variety of tasks using \OURS-B.
Specifically, for video-only tasks such as video classification (K400 [VC] and SSv2 [VC]) and spatiotemporal action localization (AVA [STAL]), we apply the frozen feature and only update the weights in the task head.
We report top-1 accuracy for video classification and mAP@IoU=$0.5$ for spatiotemporal action localization, respectively.
For zero-shot video-to-text retrieval (MSRVTT [ZSV2T], VATEX [ZSV2T], and ActivityNet [ZSV2T]), zero-shot text-to-video retrieval (MSRVTT [ZST2V], VATEX [ZST2V], and ActivityNet [ZST2V]), and zero-short video classification (K400 [ZSC], K600 [ZSC], SSv2 [ZSC], and NExT-QA-ATP [ZSC]) tasks, we apply LiT~\cite{zhai2022lit} to learn a text encoder paired up with the second-stage model as described in \cref{sec:exp:vt_retrieval} and report Recall@1 (R@1).
We notice that the second-stage training significantly improves the performance of the video encoder compared with the first-stage model across all video tasks on different datasets, strongly demonstrating the effectiveness of the proposed two-stage training.

\begin{table}[t]
\caption{\textbf{Studies on data scaling for the second-stage model.} We report results of the \textit{Large} model using MAP probing with frozen backbone and only the full pretraining corpus is used to train the first-stage model.}
	\label{tab:data_ablation_student}
	\vskip 0.1in
	\begin{center}
	\begin{scriptsize}
\begin{tabular}	{lcccc}
			\toprule
			Data & \# of clips &  \textbf{K400} & \textbf{SSv2} & \textbf{AVA} \\
			\midrule
	Full pretraining corpus & 618M & 85.8 & 66.4 & 33.1\\
	+ additional video-only  & 898M & 86.1 & 66.7 & 33.7\\
		\bottomrule
		\end{tabular}
\end{scriptsize}
	\end{center}
	\vskip -0.1in
\end{table}

\subsection{Scaling properties}

In \cref{fig:model_scaling_variable_teacher}, we study the scaling behavior of our models by keeping the data fixed. 
We find that both our first-stage model and second-stage model scale well with the model size.
Interestingly, the second-stage model shows consistent improvements over the first-stage model of around $8\%$ on SSv2  and $2.2\%$ on AVA, across the model sizes. 
In \cref{fig:model_scaling_fixed_teacher}, we scale the second-stage model by fixing the first-stage model to be of Base size.
For \textit{Large} and \textit{giant} second-stage models, as they are incompatible with the first-stage model of Base size, we initialize them with the corresponding image model of CoCa \cite{yu2022coca}.
We observe that even with a fixed first-stage model, our second-stage models still show a reasonable scaling capability.

In Table~\ref{tab:data_ablation_teacher}, we demonstrate strong data scaling capability of the first-stage model where the model trained on the pre-training corpus outperforms the one trained on InternVid \cite{wang2023internvid} by 5.4\% on MSRVTT ZST2V retrieval and 11.2\% on VATEX ZST2V retrieval.
This motivates us to also study the data scaling ability of our second-stage model.
An interesting aspect of our second-stage training is that it works with video-only data without annotations.
This equips us to economically increase the corpus size during the second-stage training.
To test the benefit of data scaling, we mine additional 280M video clips without annotations from YouTube and add them to our second-stage training.
We use model with \textit{Large} size as an example and train the first-stage model using pretraining corpus.
We then compare the second-stage model trained only on the pretraining corpus with that trained with both the pretraining corpus and the additional clips in \cref{tab:data_ablation_student}.
We can see that our second-stage model scales well with data.
Note that prior work on masked modeling either does not demonstrate good data-scaling properties or shows marginal improvements with data scaling~\cite{feichtenhofer2022masked, tong2022videomae}.

\end{document}